%% file: paper.tex
\documentclass{article}

\usepackage{arxiv}

\usepackage[utf8]{inputenc} 
\usepackage[T1]{fontenc}    
\usepackage{graphicx,xcolor}
\usepackage{hyperref}
\usepackage{subfigure}
\usepackage{booktabs} 
\usepackage{algorithm}
\usepackage{algorithmic}
\usepackage{amsmath}
\usepackage{amssymb}
\usepackage{mathtools}
\usepackage{amsthm}
\usepackage{multirow}
\usepackage{bm}
\usepackage{bbm}
\usepackage{natbib}
\usepackage[capitalize,noabbrev]{cleveref}
\crefname{section}{Sec.}{Sec.}
\crefname{equation}{Eq.}{Eq.}
\theoremstyle{plain}
\newtheorem{theorem}{Theorem}[section]

\newtheorem{lemma}[theorem]{Lemma}

\theoremstyle{definition}

\theoremstyle{remark}

\usepackage[textsize=tiny]{todonotes}
\allowdisplaybreaks[1]

\title{Learning from Label Proportions \\ with Instance-wise Consistency}

\author{Ryoma Kobayashi \\
	The University of Tokyo \\
	\texttt{kobayashi@mi.t.u-tokyo.ac.jp} \\
	\And
  Yusuke Mukuta \\
	The University of Tokyo / RIKEN \\
	\texttt{mukuta@mi.t.u-tokyo.ac.jp} \\
	\And
  Tatsuya Harada \\
	The University of Tokyo / RIKEN \\
	\texttt{harada@mi.t.u-tokyo.ac.jp} \\
}

\renewcommand{\shorttitle}{Learning from Label Proportions with Instance-wise Consistency}

\begin{document}
\date{}
\maketitle

\begin{abstract}
Learning from Label Proportions (LLP) is a weakly supervised learning method that aims to perform instance classification
from training data consisting of pairs of bags containing multiple instances and the class label proportions within the bags.
Previous studies on multiclass LLP can be divided into two categories according to the learning task:
per-instance label classification and per-bag label proportion estimation.
However, these methods often results in high variance estimates of the risk when applied to complex models,
or lack statistical learning theory arguments.
To address this issue, we propose new learning methods
based on statistical learning theory for both per-instance and per-bag policies.
We demonstrate that the proposed methods are respectively risk-consistent and classifier-consistent in an instance-wise manner,
and analyze the estimation error bounds.
Additionally, we present a heuristic approximation method that utilizes an existing method for regressing label proportions
to reduce the computational complexity of the proposed methods.
Through benchmark experiments, we demonstrated the effectiveness of the proposed methods.
\footnote{Our code is available at \url{https://github.com/ryoma-k/LLPIWC}.}
\end{abstract}

\section{Introduction}
\label{sec:intro}
With the development of machine learning---particularly deep learning---there has been a recent surge of interest in techniques for learning from incomplete observational data.
This field, which is known as weakly supervised learning, encompasses various approaches for dealing with incomplete or noisy data, such as semi-supervised learning \cite{books/mit/06/CSZ2006,6813505,sakai2017semi},
label noise learning \cite{NIPS2013_3871bd64,NEURIPS2018_a19744e2},
multiple instance learning \cite{AMORES201381,pmlr-v80-ilse18a},
partial label learning \cite{NEURIPS2020_7bd28f15,pmlr-v119-lv20a},
complementary label learning \cite{NIPS2017_1dba5eed},
positive unlabeled learning \cite{NIPS2014_35051070},
positive confidence learning \cite{NEURIPS2018_bd135462},
similar unlabeled learning \cite{pmlr-v80-bao18a},
and similar dissimilar learning \cite{10.1162/neco_a_01373}.
Each of these properties was clarified according to statistical learning theory \cite{books/mit/9780262047074}.

Learning from Label Proportions (LLP)---a weakly supervised learning method that is the focus of this study---involves learning instance classification problems using pairs of bags consisting of instances and their class-label proportions.
LLP has been applied in various domains, including image and video analysis \cite{10.1145/2647868.2654935,6909685,Ding2017,li2015alter},
physics \cite{dery2017weakly,4470249},
medicine \cite{doi:10.1177/0962280216651098,bortsova2018deep},
and activity analysis \cite{poyiadzi2018label}.
In this study, we focus on multiclass LLP (MCLLP), which can be divided into two categories according to the learning task:
per-instance label classification \cite{zhang2022learning,abs-1905-12909,ijcai2021-377}
and per-bag label proportion estimation \cite{ArdehalyC17a,NEURIPS2019_4fc84805, pmlr-v129-tsai20a,DBLP:conf/acml/YangZL21,9897895}.
The existing method for the former \cite{zhang2022learning} is based on statistical learning theory and exhibits classification performance consistency.
It estimates the overall expected losses and, as we show later, tends to misestimate the risk---particularly when applied to complex models such as deep learning models.
For the latter, the existing methods rely on empirical techniques,
such as regressing class proportions by averaging instance outputs \cite{ArdehalyC17a}.
However, these studies lacked a clear statistical foundation.
To address these issues, we propose two new MCLLP methods: one for per-instance label classification and one for per-bag label proportion classification.
The first method employs a risk-consistent approach utilizing a loss function equivalent to supervised learning
for \textit{each individual instance} along with their expectations.
In contrast, the second method employs a classifier-consistent approach, producing an instance classifier equivalent to those obtained through supervised learning; the proof is again discussed for \textit{each individual instance}.
In addition, we present a heuristic approximation method to reduce the computational complexity of these methods,
which utilizes the average operation commonly employed in existing per-bag learning approaches.
The contributions of this study can be summarized as follows.
\begin{itemize}
  \item We introduce two MCLLP methods for per-instance label and per-bag label proportion classification and prove their
        consistency in an instance-wise manner.
        We also derive bounds for their estimation errors.
  \item We analyze the method using the mean output operation \cite{ArdehalyC17a} and demonstrate that it can be perceived as a maximization of the likelihood of label proportions.
        Then, we propose a heuristic approximation method using this operation to reduce the computational complexity of our methods.
\end{itemize}

\section{Formulations and Related Studies}
\label{sec:related}
In this section, we introduce standard multiclass classification, LLP, and partial label learning (PLL)
and subsequently review related studies.

\subsection{Standard Multiclass Classification}
In $C$-class classification, let $\mathcal{X}$ be the input space and $\mathcal{Y}=\{1,... ,C\}$ be the space of the labels.
Typically, we assume that data $(x,y) \in (\mathcal{X}, \mathcal{Y})$ are independently sampled from the probability distribution $p(x,y)$.
Let $\ell\colon \mathbb{R}^{k}\times \mathcal{Y} \rightarrow \mathbb{R}_{\geq 0}$ denotes the multiclass loss function,
which measures the difference between the label and predicted output.
The goal of multiclass classification is to minimize the subsequent predictive loss
for the hypothesis $f\colon \mathcal{X} \rightarrow \mathbb{R}^{C}$.
\begin{align}
  \label{eq:risk_supervised}
  R(f) = \underset{(x,y)\sim p(x,y)} {\mathbb{E}} [\ell(f(x),y)].
\end{align}
In practice, $p(x,y)$ is unobserved, and we use training data $\tilde D=\{(x_{i},y_{i})\}_{i=1}^{n}$
to minimize the empirical loss \cite{Vapnik1998}:
\begin{align*}
  \hat{R}(f) = \frac{1}{n} \sum_{i=1}^{n}\ell(f(x_{i}),y_{i}).
\end{align*}
A method is called risk-consistent if it has the same risk as \cref{eq:risk_supervised},
whereas a method is called classifier-consistent if it yields the same optimal classifier as the supervised classifier.

\subsection{LLP}
In LLP, pairs of bags consisting of multiple instances and the proportion of labels they contain are given.
In this study, we formulate LLP using unordered pairs that allow overlap, that is, multisets, instead of the proportions of class labels.
Hereinafter, the multiset is denoted by $\{|\cdot|\}$, for example, $\{|1, 1, 2|\}$.
Let $\mathcal{X}^{K}$ and $\mathcal{Y}^{K}$ be the direct product spaces of bags and labels, respectively, with K instances,
and let $\mathcal{S}^{K}$ be the space of label multisets.
With training data $\tilde D = \{(X_{i}, S_{i}) \in (\mathcal{X}^{K}, \mathcal{S}^{K}) \}_{i=1}^{n}$,
we represent each instance and label as $X_{i} = (X_{i}^{(1)},... ,X_{i}^{(K)})$ and $Y_{i} = (Y_{i}^{(1)},... ,Y_{i}^{(K)})$, where the top and bottom subscripts associate them with a particular bag.
Furthermore, we define $\mathcal{Y}_{\sigma(S)}^{K}$ as the set of all possible label assignments from the label multiset $S$:
\begin{align*}
  \mathcal{Y}_{\sigma(S)}^{K} := \left\{Y \in \mathcal{Y}^{K} \middle| \{|Y^{(1)},...,Y^{(K)}|\} = S\right\}.
\end{align*}
The LLP then assumes that the following holds:
\begin{align}
  \label{eq:PSXY}
  P(S_{i}|X_{i},Y_{i}) = P(S_{i}|Y_{i}) =
  \begin{cases}
    1 & \text{if~~~} Y_{i} \in \mathcal{Y}_{\sigma(S_{i})}^{K}, \\
    0 & \text{otherwise.}
  \end{cases}
\end{align}

Various techniques have been applied to address LLP,
including linear models \cite{WANG2013273,8257973,10.1007/978-3-319-44636-3_8},
support vector machines \cite{icml-2010-Ruping,pmlr-v28-yu13a,7549044,7836756,pinball,CHEN2017126,7397422,LU2019413},
a clustering algorithm \cite{10.5555/2034161.2034185},
Bayesian networks \cite{HERNANDEZGONZALEZ20133425,doi:10.1177/0962280216651098},
random forest \cite{DBLP:journals/nn/ShiLQW18},
and graph-based approaches \cite{poyiadzi2018label}.
Here, we focus on the learning tasks used in existing research and divide them into two categories:
per-instance label classification and per-bag label proportion estimation.
The learning task of per-instance label classification can be further divided into two policies.
The initial policy estimates the parameters or losses by utilizing an equation that holds only for expectations.
This policy has been the subject of longstanding investigations in the field of LLP and has been analyzed
using statistical learning theory \cite{JMLR:v10:quadrianto09a,NIPS2014_a8baa565,lu2018on,NEURIPS2020_fcde1491}.
As an example of a study related to MCLLP, \citet{zhang2022learning} recently proposed an extension of the binary classification LLP method
using label noise forward correction \cite{NEURIPS2020_fcde1491} for MCLLP and achieved classifier-consistency.
However, it is important to note that the hypothesis that minimizes the loss of label noise is equivalent to the hypothesis
that minimizes the loss of supervised learning on expectation but not necessarily on the loss for each instance.
Therefore, using complex models such as deep learning may lead to high variance estimation of the risk, and
as discussed in \cref{sec:exp}, its classification performance may not be as high as that of the other methods.
\citet{2207.01555} proposed a technique utilizing label noise backward correction with the constraint of per-bag loss.
Statistical analysis of this constraint has yet to be conducted.
The other policy for taking loss per instance is to take classification loss by creating pseudo-labels.
\citet{abs-1905-12909} and  \citet{ijcai2021-377} proposed creating pseudo-labels using entropy-constrained
optimal transport and taking per-instance losses.
However, these methods lack a statistical background, and their properties are unclear.
To estimate per-bag label proportions,
\citet{yu2015learning} introduced Empirical Proportion Risk Minimization (EPRM),
which involves empirical risk minimization for the bag proportions in the binary setting.
Under the assumptions of the bag generation process and distribution of the model outputs,
they showed that the instance classification error can be bounded by the classification error of the label proportion.
Building on this approach, \citet{ArdehalyC17a} proposed Deep LLP (DLLP) as an extension of EPRM for handling multiple classes.
DLLP aims to minimize the Kullback–Leibler divergence between the predicted mean probabilities for each class, which is
represented by $\bar p (y=c | X_{i}) := \frac{1}{K} \sum_{k} p(y = c | X_{i}^{(k)})$,
and the proportion of each class within a bag, which is represented by $p (y=c|S_{i}) = \frac{1}{K} \sum_{y \in S} \mathbbm{1}\{y = c\}$,
as follows:
\begin{align}
  \label{eq:risk_dllp}
  L_{prop} = - \sum_{c} p(y=c | S_{i}) \log(\bar p (y=c| X_{i})).
\end{align}
In subsequent studies, DLLP was applied to self-supervised \cite{pmlr-v129-tsai20a},
contrastive learning \cite{DBLP:conf/acml/YangZL21},
and semi-supervised learning using generative adversarial networks \cite{NEURIPS2019_4fc84805},
although the use of the mean operation for these methods is not fully understood,
and the consistency of instance classification is yet to be demonstrated in multiclass settings.
As a method that does not use the mean operation,
\citet{9897895} proposed a method that maximizes the likelihood of bag proportions using the EM algorithm.
However, its background based on statistical learning theory has not been shown.

\subsection{PLL}
PLL involves predicting the correct label when multiple candidate labels are provided.
MCLLP can be viewed as a PLL setting by treating all labels given for a particular bag as potential candidates.
A notable contribution to the field of PLL was PRODEN \cite{pmlr-v119-lv20a},
which progressively updates the labels of instances during the learning process and was shown to be risk-consistent \cite{NEURIPS2020_7bd28f15}.
A more generalized risk \cite{2112-12303} is expressed as follows:
\begin{align}
  \label{eq:risk_pllrc}
  R_{PLLrc}(f) = \underset{p(x,S)} {\mathbb{E}} [\sum_{y \in S} \frac{P(y|x)}{P(S|x)} \ell(f(x),y)].
\end{align}
\citet{NEURIPS2020_7bd28f15} showed that a classifier-consistent method can be realized
by classifying label candidates under the assumption of a distribution of partial labels.
If the probability that the random variable representing a candidate label is $S = T_{j}$ is estimated using
a multivalued function $q_{j}(x) = P(S = T_{j} | x)$, the risk is expressed as
\begin{align}
  \label{eq:risk_pllcc}
  R_{PLLcc}(f) = \underset{p(x,S)} {\mathbb{E}} [\ell(q(x),S)].
\end{align}
Inspired by PLL methods based on statistical learning theory,
we propose a learning method involving per-instance label classification and per-bag label proportion classification.

\section{Per-Instance Method}
\label{sec:rc}
In this section, we propose a method for MCLLP that involves per-instance label classification.
We demonstrate that our method is risk-consistent and present a learning procedure following the computation of its estimation error bounds.

\subsection{Risk Estimation}
We assume instance label independence, as follows:
\begin{align}
  \label{eq:PYX}
  P(Y | X) = \Pi_{i=1}^{K} P(Y^{(i)} | X^{(i)}).
\end{align}
Let $\mathit{set}(S)$ be the set of elements of multiset $S$ excluding duplicates, for example, $\mathit{set}(\{|1,1,2|\}) = \{1,2\}$.
The probability of a pair of bags and their labels $P(X, Y)$ can be transformed using the pair $(X, S)$ of a bag and its label proportion as follows:
\begin{align}
  \label{eq:PXY}
  P(X,Y)
  &= \sum_{S\in\mathcal{S}^{K}} P(X,Y,S) = \sum_{S\in\mathcal{S}^{K}} P(X,S) \frac{P(X,Y,S)}{P(X,S)} \nonumber \\
  &= \sum_{S\in\mathcal{S}^{K}} P(X,S) \frac{P(Y,S|X)}{P(S|X)}
\end{align}
\begin{align}
  \label{eq:PYSX}
  P(Y,S|X) = P(S|X,Y) P(Y|X)
           \overset{(\ref{eq:PSXY})} =
  \begin{cases}
    P(Y|X) & \text{if~~~} Y \in \mathcal{Y}_{\sigma(S)}^{K}, \\
    0 & \text{otherwise}
  \end{cases}
\end{align}
\begin{align}
  \label{eq:PSX}
  P(S|X) &= \sum_{Y\in \mathcal{Y}^{K}} P(Y,S|X)
  \overset{(\ref{eq:PYSX})} = \sum_{Y\in \mathcal{Y}_{\sigma(S)}^{K}} P(Y|X) \nonumber \\
     &\overset{(\ref{eq:PYX})} = \sum_{y \in \mathit{set}(S)} P(y | X^{(k)}) P(S \backslash y | X \backslash X^{(k)}).
\end{align}
The risk function $R(f)$ used in ordinary supervised learning can be transformed using these relationships, yielding the following expression:
\begin{align}
  \label{eq:risk_rc}
  &{} R(f) = \underset{p(x,y)} {\mathbb{E}} \left[\ell(f(x),y)\right]
  = \frac{1}{K} \underset{p(X,Y)} {\mathbb{E}}
     \left[ \sum_{k} \ell(f(X^{(k)}), Y^{(k)}) \right] \nonumber \\
  &= \frac{1}{K} \int_{\mathcal{X}^{K}} \sum_{Y \in \mathcal{Y}^{K}} P(X,Y) \sum_{k} \ell(f(X^{(k)}), Y^{(k)}) dX \nonumber \\
  &\overset{(\ref{eq:PXY})}= \frac{1}{K} \int_{\mathcal{X}^{K}} \sum_{Y \in \mathcal{Y}^{K}}
  \sum_{S\in\mathcal{S}^{K}}
  P(X,S) \frac{P(Y,S|X)}{P(S|X)} \sum_{k} \ell(f(X^{(k)}), Y^{(k)}) dX \nonumber \\
  &\overset{(\ref{eq:PYSX})}= \frac{1}{K} \underset{p(X,S)} {\mathbb{E}} \left[
  \sum_{Y \in \mathcal{Y}_{\sigma(S)}^{K}} \frac{P(Y|X)}{P(S|X)}
  \sum_{k} \ell(f(X^{(k)}), Y^{(k)}) \right] \nonumber \\
  &\overset{(\ref{eq:PSX})} = \frac{1}{K} \underset{p(X,S)} {\mathbb{E}} \left[
  \sum_{k} \sum_{y \in \mathit{set}(S)}
  \frac{P(y | X^{(k)}) P(S \backslash y | X \backslash X^{(k)})}
{\sum_{y^{\prime} \in \mathit{set}(S)} P(y^{\prime} | X^{(k)}) P(S \backslash y^{\prime} | X \backslash X^{(k)})} \ell(f(X^{(k)}), y) \right] \nonumber \\
  &:= R_{rc}(f).
\end{align}
We define $R_{rc}$ as the risk estimated from a pair $(X, S)$ of instance sets and label proportions in \cref{eq:risk_rc}.
From this equation, it is possible to learn the possible label candidates $y$ from a given label multiset $S$
by weighting the typical supervised learning loss.
\subsection{Risk-Consistency \& Estimation Error Bound}
We begin by confirming that our method is risk-consistent.
It is shown from \cref{eq:risk_rc} that the proposed risk is the same as that in supervised learning.
We emphasize that each instance takes the same loss as supervised learning by \cref{eq:PXY}.

In the following section, we analyze the estimation error bounds for this loss.
It is assumed that $P(y|X^{(k)})$ is fixed.
Let $\hat f_{rc} = \min_{f\in \mathcal{F}}\hat{R}_{rc}(f)$ and $f_{rc}^{*} = \min_{f\in \mathcal{F}} R_{rc}(f)$ be hypotheses
to minimize the empirical risk and predictive risk, respectively.
Let the hypothesis space be $\mathcal{H}_{y} : \{h : x \rightarrow f_{y}(x) | f\in \mathcal{F}\}$,
and let $\mathfrak{R}_{n}(\mathcal{H}_{y})$ be the expected Rademacher complexity of $\mathcal{H}_{y}$ \cite{journals/jmlr/BartlettM02}.
Suppose that the loss function $\ell(f(x), y)$ is $\rho$-Lipschitz with respect to the inputs and bounded by $M$.
\begin{theorem}
\label{th:rc_bound}
For any $\delta > 0$, we have with probability at least $1 - \delta$,
\begin{align*}
  R_{rc}(\hat f_{rc}) &- R_{rc}(f_{rc}^{*})
    \leq 4 \sqrt{2} \rho \sum_{y=1}^{C} \mathfrak{R}_{n} (\mathcal{H}_{y}) + 2M \sqrt{\frac{\log \frac{2}{\delta}}{2 n}}.
\end{align*}
\end{theorem}
The proof is provided in \cref{proof:rc_bound}.
In general, $\mathfrak{R}_{n}(\mathcal{H}_{y})$ is bounded by some positive constant $\mathcal{C}_{\mathcal{H}}/\sqrt{n}$ \cite{pmlr-v75-golowich18a},
which indicates the convergence of $f_{rc}$ to $f_{rc}^{*}$ as $n\rightarrow \infty$.

\subsection{Learning Method}
In the actual learning process, given that $P(y|X^{(k)})$ values are unknown during the learning process,
it is necessary to estimate them.
As a learning method, we progressively update $P(y|X^{(k)})$ simultaneously, similar to PRODEN \cite{pmlr-v119-lv20a}.
This method is presented in \cref{alg:rc}.
We define this method as RC method.

\begin{algorithm}[tb]
   \caption{RC / RC\_Approx Algorithm}
   \label{alg:rc}
\begin{algorithmic}
  \STATE {\bfseries Input:} Model $f$, epoch $T_{max}$, proportions labeled training set $\tilde{D} = \{(X_{i}, S_{i})\}_{i=1}^{n}$
  \STATE Initialize $P(y|X_{i}^{(k)}) = \frac{1}{K} \sum_{y \in S_{i}} \mathbbm{1}\{y = c\}$.
   \FOR{$t = 1$ {\bfseries to} $T_{max}$}
   \STATE Shuffle $\tilde{D}$ into $B$ mini-batches.
   \FOR{$b=1$ {\bfseries to} $B$}
   \STATE COMPUTE $\hat P$ = softmax($f(X_{i}^{k})$).
   \STATE UPDATE $f$ by \cref{eq:risk_rc}.
   \STATE UPDATE $P(Y_{i}^{(k)} = y|X_{i}^{(k)})$ by $\hat P$.
   \ENDFOR
   \ENDFOR
\end{algorithmic}
\end{algorithm}

\section{Per-Bag Method}
\label{sec:cc}
In this section, we describe a classifier-consist method that involves per-bag label proportion classification
and provide a detailed description of the learning procedure, including the calculation of the estimation error bounds.

\subsection{Risk Estimation}
For a given label multiset variable $S$, there are ${}_{K}H_{C} = \frac{(K + C - 1)!}{K!(C-1)!}$ candidates,
and the space of the label multiset $\mathcal{S}^{K}$ can be represented as $\mathcal{S}^{K} = \{T_{j} | j \in \{1, \cdots, {}_{K}H_{C}\}\}$.
We propose classifying the multiset $T_{j}$ and estimating the probability $P(S= T_{j}|X)$
using each instance output $e(y = c | X^{(k)})$, where $e$ denotes the softmax of the outputs.
In accordance with \cref{eq:PSX}, we design a multivalued function $q$ such that the $j$th output is $q_{j}(X) = P(S = T_{j} | X)$, as follows:
\begin{align}
  \label{eq:cc_q}
  q_{j}(X) = \sum_{y \in \mathit{set}(T_{j})} e(y | X^{(1)}) P(T_{j} \backslash y | X \backslash X^{(1)}).
\end{align}
Using this $q$, we propose the following per-bag loss by letting $s(X, i)$ be the operation of swapping the first and \textit{i}-th instances of bag $X$:
\begin{align}
  \label{eq:risk_cc}
  R_{cc}(f) = \underset{p(X,S)} {\mathbb{E}} \frac{1}{K} \sum_{k} \mathcal{L}(q(s(X,k)),S),
\end{align}
where we utilize loss function which takes scalar input for $\mathcal{L} \colon (\mathbb{R} \times \mathcal{Y}) \rightarrow \mathbb{R}$, e.g., cross-entropy.
This eliminates the need to estimate ${}_{K}H_{C}$ candidate multisets; only for a given multiset, $S$ is required.
\subsection{Classifier-Consistency \& Estimation Error Bound}
First, we discuss the classifier-consistency.
The following lemma, which was presented in several works on PLL \cite{Yu_2018_ECCV,pmlr-v119-lv20a,NEURIPS2020_7bd28f15},
is crucial for the discussion of classifier-consistency.
\begin{lemma}
  \label{lem:crossent}
  \cite{Yu_2018_ECCV} If the cross-entropy or mean squared error loss is used for $\ell$ in \cref{eq:risk_supervised},
  the optimal classifier $f^{*}$ satisfies $f^{*}_{i}(x) = P(y = i | x)$.
\end{lemma}
From this lemma, we obtain the following theorem.
The proof is provided in \cref{proof:cc_consistency}.
We emphasize that this proof is discussed by instance-wise outputs.
\begin{theorem}
  \label{th:cc}
  Under the assumption that we use the cross-entropy or mean-squared error loss,
  the hypotheses $f$ that minimize $R(f)$ in \cref{eq:risk_supervised} and $R_{cc}(f)$ are equal.
\end{theorem}
In the following section, we analyze the estimation error bounds.
We assume that $P(y|X^{(k)})$ is fixed.
Let $\hat f_{cc} = \min_{f\in \mathcal{F}}\hat{R}_{cc}(f)$ and $f_{cc}^{*} = \min_{f\in \mathcal{F}} R_{cc}(f)$ be hypotheses
to minimize the empirical risk and predictive risk.
Let the hypothesis space be $\mathcal{H}_{y} : \{h : x \rightarrow f_{y}(x) | f\in \mathcal{F}\}$,
and let $\mathfrak{R}_{n}(\mathcal{H}_{y})$ be the expected Rademacher complexity of $\mathcal{H}_{y}$ \cite{journals/jmlr/BartlettM02}.
Suppose that the loss function $\mathcal{L} \colon (\mathbb{R} \times \mathcal{Y}) \rightarrow \mathbb{R}_{\geq 0}$
is $\rho$-Lipschitz with respect to the inputs and bounded by $M$.
\begin{theorem}
\label{th:cc_bound}
For any $\delta > 0$, we have with probability at least $1 - \delta$,
\begin{align*}
  R_{cc}(\hat f_{cc}) &- R_{cc}(f_{cc}^{*})
    \leq 4 \sqrt{2} \rho \sum_{y=1}^{C} \mathfrak{R}_{n} (\mathcal{H}_{y}) + 2M \sqrt{\frac{\log \frac{2}{\delta}}{2 n}}.
\end{align*}
\end{theorem}
The proof is provided in \cref{proof:cc_bound}.
Thus, we demonstrate through the theorem that $f_{cc}$ converges to $f_{cc}^{*}$
by employing the appropriate loss function and $n\rightarrow \infty$.
\subsection{Learning Method}
In the actual learning process, we must estimate $P(y | X^{(k)})$ as in \cref{sec:rc}.
In this method, we propose directly using the output of the current classifier to estimate $P(y | X^{(k)})$.
Thus, $P(T_{j} \backslash y | X \backslash X^{(1)})$ in \cref{eq:cc_q} can be rewritten as
$\sum_{Y^{\prime} \in \mathcal{Y}_{\sigma(T_{j} \backslash y)}^{K - 1}} \prod_{i=1}^{K-1} e(Y^{\prime (i)} | (X \backslash X^{(1)})^{(i)})$.
If this term is optimized by the EM algorithm using a logarithmic function for loss,
the loss function is the same as in \citet{9897895}, and our method is a generalization of this method.
The learning algorithm is described in \cref{alg:cc}.
We define this method as CC method.

\begin{algorithm}[tb]
   \caption{CC / CC\_Approx Algorithm}
   \label{alg:cc}
\begin{algorithmic}
  \STATE {\bfseries Input:} Model $f$, epoch $T_{max}$, proportions labeled training set $\tilde{D} = \{(X_{i}, S_{i})\}_{i=1}^{n}$
   \FOR{$t = 1$ {\bfseries to} $T_{max}$}
   \STATE Shuffle $\tilde{D}$ into $B$ mini-batches.
   \FOR{$b=1$ {\bfseries to} $B$}
   \STATE UPDATE $f$ by (\ref{eq:risk_cc})
   \ENDFOR
   \ENDFOR
\end{algorithmic}
\end{algorithm}

\section{Approximation Method}
\label{sec:approx}
In this section, we present a computational complexity-reduction technique for the method outlined in \cref{sec:rc,sec:cc}.
We commence our analysis by considering DLLP \cite{ArdehalyC17a} as a likelihood maximization method
that is approximated by a multinomial distribution.
We utilized this heuristic approximation in the implementation of the proposed methods.
\subsection{DLLP as likelihood maximization}
To consider DLLP as a likelihood maximization method,
we first introduce the concept of representing the label multiset $S$ by using a random variable $n_{1}, \cdots n_{C}$ that corresponds to the number of instances of each class within the bag, that is, $P(S) = P(n_{1}, \cdots n_{C})$,
$n_{c} = \sum_{y \in S} \mathbbm{1}\{y = c\}$.
Then, the multinomial distribution for the parameters $n, p_{1},\cdots,p_{C}$ $(\sum p_{i} = 1, p_{i} > 0)$
and the support $n_{1},\cdots,n_{C} \in \{1,...,n\}$ $(\sum n_{i} = n)$ can be expressed as follows:
\begin{align}
  \label{eq:multinomial}
  P(n_{1},\cdots,n_{C}) = \frac{n!}{n_{1}!\cdots n_{C}!} p_{1}^{n_{1}}\cdots p_{C}^{n_{C}}.
\end{align}
Subsequently, we assume a multinomial distribution with $n=K$ for $P(S)$.
With the class posterior probability $P(y|X^{(k)})$ of each instance,
we can estimate the expected value $\bar p_{c}$ of the class fraction in the bag:
\begin{align*}
  \bar p_{c} &= \frac{1}{K} {\mathbb{E}} [n_{c}] = \frac{1}{K} \sum_{k} P(Y^{(k)} = c|X^{(k)}).
\end{align*}
Thus, we can express $P(S|X)$ approximated by a multinomial distribution with $\bar p_{c}$ as its parameters, as follows:
\begin{align}
  \label{eq:approx_PSX}
  \tilde P(S|X) = \frac{K!}{n_{1}!\cdots n_{C}!} \bar p_{1}^{n_{1}}\cdots \bar p_{C}^{n_{C}}.
\end{align}
Furthermore, calculating the negative log-likelihood for \cref{eq:approx_PSX} yields the following relationship,
indicating that DLLP implicitly minimizes the negative log-likelihood of $\tilde P(S|X)$,
which approximates $P(S|X)$ with a multinomial distribution:
\begin{align*}
  -\log \tilde P(S|X) &= - \sum_{c} n_{c} \log \bar{p_{c}} - \log \frac{n!}{n_{1}!\cdots n_{C}!} \\
                      &= K ~ L_{prop} + \text{const.}
\end{align*}

\subsection{Approximation with Mean Operation}
Here, we briefly introduce a technique for reducing the computational complexity of the methods presented in \cref{sec:rc,sec:cc}
by utilizing a multinomial distribution as in DLLP.
The proposed methods for estimating $P(S \backslash y | X \backslash X^{(k)})$ have a computational complexity of $\mathcal{O}(K^{C})$.
Thus, their implementation becomes computationally challenging when the number of instances in the bag increases.
Therefore, we approximate it
as $\tilde P(S \backslash y | X \backslash X^{(k)})$ from \cref{eq:approx_PSX}, which can be computed in $\mathcal{O}(KC)$.
These approximation methods are defined as RC\_Approx and CC\_Approx.
Although the proposed approximation is heuristic in \cref{sec:exp},
we validated it through experiments.

\section{Experiments}
\label{sec:exp}

\input exp_table_1.tex

\input exp_fig_llpfc.tex

\input exp_fig_ot.tex

\input exp_fig_weight_diff.tex
We evaluated the proposed risk estimators and their approximation methods experimentally.
\textbf{Settings~~~}
We used four commonly used datasets:
MNIST \cite{lecun-gradientbased-learning-applied-1998},
Fashion-MNIST \cite{DBLP:journals/corr/abs-1708-07747},
Kuzushiji-MNIST \cite{DBLP:journals/corr/abs-1812-01718}, and
CIFAR-10 \cite{Krizhevsky09learningmultiple}.
We conducted the experiments by randomly generating bags and varying the number of instances, i.e., $K = \{2,4,8,16,32,64,128\}$.
We employed a linear model (Linear), a 5-layer perceptron (MLP),
a 12-layer ConvNet \cite{convnet}, and a 32-layer ResNet \cite{7780459} in our experiments.
We utilized 10\% of the training data for validation, including searching for the optimal
learning rate from $\{10^{-6},...,10^{-1}\}$ and determining the optimal number of training epochs.
We set the batch size such that the total number of instances in a mini-batch was 256 and utilized the cross-entropy as the loss function and Adam \cite{adam} with weight decay $10^{-5}$ as the optimizer.
A detailed description of the experimental setup is presented in the Appendix.
The experiments were carried out with NVIDIA Tesla V100 GPU.

\textbf{Methods~~~}
To confirm the effectiveness of the proposed methods, we compared them with several baselines:
\begin{itemize}
  \setlength{\itemsep}{0cm}
  \item Supervised: utilize supervised information \cref{eq:risk_supervised}
  \item LLPFC: per-instance label classification with label noise correction \cite{zhang2022learning}
  \item OT: per-instance label classification using pseudo-labels made by optimal transport \cite{ijcai2021-377}
  \item DLLP: per-bag label proportion regression with mean operation \cite{ArdehalyC17a}
\end{itemize}

\textbf{Results~~~}
The accuracy of the test data was presented in \cref{tab:mnists-mlp}--\ref{tab:cifar10-convnet} and \ref{tab:mnists-linear}.
We also report the paired $t$-test results at a $5\%$ significance level for the best of the proposed methods and each comparison method.
$\bullet$ and $\circ$ indicate that our best method was significantly better and worse, respectively.
First, we compare the proposed methods with LLPFC and OT, which use the loss per instance.
For LLPFC, the proposed methods performed better in many settings in our experiments.
As shown in \cref{fig:train-accuracy-llpfc},
for LLPFC, the training accuracy increased as the loss decreased for simple linear models.
However, for complex models, such as deep learning, despite a reduction in the loss, the accuracy decreased---particularly when K was large.
A similar phenomenon, in which the accuracy decreases despite decreasing losses when complex models are used,
has been observed in various weakly supervised learning studies involving risk estimation,
which holds only for the expected values \cite{NIPS2017_7cce53cf, pmlr-v108-lu20c, pmlr-v119-chou20a}.
Thus, LLPFC may have resulted in biased loss estimations for similar reasons.
In contrast, no such trend was observed for the proposed methods.
Compared with OT, which generates pseudo-labels through optimal transport, the proposed methods exhibited equal or superior performance, except for the linear model settings.
Considering the linear setting results, OT sometimes performed unstably in relatively simple settings such as $K=2$ and 4,
as shown in \cref{fig:train-accuracy-curve-ot}, suggesting that the optimal transport labels were unstable during the training.
Subsequently, we compared the proposed methods with DLLP, which takes per-bag losses.
While the proposed methods outperformed DLLP,
no significant difference was observed among the results of CC, CC\_Approx, and DLLP methods in many settings.
As discussed in \cref{sec:approx}, we can consider DLLP as an approximation method for CC.
The results suggest that the same level of performance can be achieved through approximation of the bag-wise loss.
Finally, an approximation method is discussed.
The proposed approximation method addresses the memory limitations in experimental setups with $K \geq 16$,
and its performance is comparable to that of non-approximate methods in settings where $K=2,4$ or $8$.
The average difference between the label weights with and without approximation using the RC method is shown in \cref{fig:approx-weight-diff}.
For $K=2$, the two methods produced the same results, except for the computer error.
For $K=4$ and $8$, the results of the two methods differed, but the difference decreased as the learning converged.

\section{Conclusion}
\label{sec:conclusion}
We propose learning methods for MCLLP based on statistical learning theory.
First, we introduced a method based on per-instance label classification and examined its risk-consistency.
We then proposed a method based on the classification of per-bag label proportions and analyzed its classifier-consistency.
In addition, we presented a heuristic approximation using a multinomial distribution for the label proportions, which was implicitly used in the previous methods.
We discussed partially incorporating this approximation into the proposed methods for reducing the computational complexity.
Experimental results confirmed the effectiveness of the proposed methods.
Future work will involve integrating the two proposed methods and studying an approximation method with provable guarantees.

\section*{Acknowledgements}
This work was partially supported by JST AIP Acceleration Research JPMJCR20U3, Moonshot R\&D Grant Number JPMJPS2011, CREST Grant Number JPMJCR2015, JSPS KAKENHI Grant Number JP19H01115 and Basic Research Grant (Super AI) of Institute for AI and Beyond of the University of Tokyo.

\bibliographystyle{unsrtnat}
\bibliography{paper}

\newpage
\appendix

\section{Proofs}
\subsection{Proof of \cref{th:rc_bound}}
\label{proof:rc_bound}
First, we derived the estimation error bounds for the risk-consistent method proposed in \cref{sec:rc}.
We begin by defining the function set $\mathcal{G},\mathcal{G}_{rc}$ as follows:
\begin{align*}
  \mathcal{G} = \{(x,y) \mapsto \ell(f(x), y) | f \in \mathcal{F} \},
\end{align*}
\begin{align*}
  \mathcal{G}_{rc} = \{(X,S) \mapsto
  \frac{1}{K} \sum_{Y \in \mathcal{Y}_{\sigma(S)}^{K}} \frac{P(Y|X)}{P(S|X)}
  \sum_{k} \ell(f(X^{(k)}), Y^{(k)}) | f \in \mathcal{F} \}.
\end{align*}

\begin{lemma}
  \label{lem:rc_1}
  Let $\ell$ denote the loss function and let $M$ denote an upper bound on this function.
  Then, $\forall \delta > 0$ with probability at least $1 - \delta$, we have:
  \begin{align*}
    \underset{f \in \mathcal{F}}\sup |R_{rc}(f) - \hat R_{rc}(f) |
    \leq 2 \mathfrak{R}_{n} (\mathcal{G}_{rc}) + M\sqrt{\frac{\log \frac{2}{\delta}}{2n}}.
  \end{align*}
\end{lemma}

\begin{proof}
  First, consider $\underset{f \in \mathcal{F}}\sup R_{rc}(f) - \hat R_{rc}(f)$.
  Because the loss function satisfies the condition $0 \leq \ell(x,y) \leq M$,
  replacing one sample $(X,S)$ with another sample $(X^{\prime}, S^{\prime})$ does not result in a change
  in $\underset{f \in \mathcal{F}}\sup R_{rc}(f) - \hat R_{rc}(f)$ exceeding $\frac{M}{n}$.
  Then, using McDiarmid's inequality \cite{mcdiarmid1989method}, we can assert that $\forall \delta > 0$ with a probability of at least $1 - \frac{\delta}{2}$, we have:
  \begin{align*}
    \underset{f \in \mathcal{F}}\sup R_{rc}(f) - \hat R_{rc}(f)
    \leq \mathbb{E}\left[\underset{f \in \mathcal{F}}
    \sup R_{rc}(f) - \hat R_{rc}(f)\right] + M \sqrt{\frac{\log \frac{2}{\delta}}{2n}}.
  \end{align*}
  Furthermore, by utilizing the technique of symmetrization, as described in \citet{Vapnik1998},
  we demonstrate the following bound:
  \begin{align*}
    \mathbb{E}\left[\underset{f \in \mathcal{F}} \sup R_{rc}(f) - \hat R_{rc}(f)\right]
    \leq 2 \mathfrak{R}_{n} (\mathcal{G}_{rc}).
  \end{align*}
  Additionally, we show that the same bounds hold with a probability of at least $1-\frac{\delta}{2}$
  for $\underset{f \in \mathcal{F}}\sup \hat R_{rc}(f) - R_{rc}(f)$.
\end{proof}

\begin{lemma}
  \label{lem:rc_2}
  Assuming that the loss function $\ell$ is $\rho$-Lipschitz, that instances are independently and identically distributed,
  and that the conditional probability $P(Y|X),P(S|X)$ does not depend on hypothesis $f$, we can show that the following holds:
  \begin{align*}
    \mathfrak{R}_{n} (\mathcal{G}_{rc}) \leq \sqrt{2} \rho \sum_{y=1}^{C} \mathfrak{R}_{n}(H_{y}).
  \end{align*}
\end{lemma}
\begin{proof}
  Let $\phi_{X,S} : \mathbb{R}^{C} \rightarrow \mathbb{R}_{+}$ be defined as
  \begin{align*}
    \phi_{X,S}(\bm{z}) = \sum_{Y\in\mathcal{Y}_{\sigma(S)}^{K}} \frac{P(Y|X)}{P(S|X)} \ell(\bm{z}, Y^{(k)}).
  \end{align*}
  In this case, $\sum_{Y\in\mathcal{Y}_{\sigma(S)}^{K}} \frac{P(Y|X)}{P(S|X)} = 1$,
  and $\phi_{X,S}$ is a $\rho$-Lipschitz function.
  Therefore, the following holds:
  \begin{align*}
    \mathfrak{R}_{n}(\mathcal{G}_{rc})
    &= \mathbb{E} \left[\underset{g \in \mathcal{G}_{rc}} \sup \sum_{i}^{n} \sigma_{i} g(X_{i}, S_{i}) \right] \\
    &= \mathbb{E} \left[\underset{f \in \mathcal{F}} \sup \sum_{i}^{n} \sigma_{i} \frac{1}{K} \sum_{k}
    \sum_{Y\in\mathcal{Y}_{\sigma(S)}^{K}} \frac{P(Y|X)}{P(S|X)} \ell(f(X^{(k)}), Y^{(k)}) \right] \\
    &= \mathbb{E} \left[\underset{f \in \mathcal{F}} \sup \sum_{i}^{n} \sigma_{i} \frac{1}{K} \sum_{k} \phi_{X,S} (f(X^{(k)})) \right] \\
    &= \frac{1}{K} \sum_{k} \mathbb{E} \left[\underset{f \in \mathcal{F}} \sup \sum_{i}^{n} \sigma_{i} \phi_{X,S} (f(X^{(k)})) \right] \\
    &\leq \sqrt{2} \rho \mathbb{E} \left[\underset{f \in \mathcal{F}} \sup \sum_{i}^{n} \sum_{c} \sigma_{ic} f(X^{(1)}) \right] \\
    &= \sqrt{2} \rho \sum_{c} \mathfrak{R}_{n}(\mathcal{H}_{c}).
  \end{align*}
  When deriving this result, the Rademacher vector contraction inequality \cite{vectorrademachercontraction} was employed for the second line from the end.
\end{proof}

From \cref{lem:rc_1,lem:rc_2}, we obtain the following estimated error bounds:
\begin{align*}
  R_{rc}(\hat f) - R_{rc}(f^{*})
  &\leq R_{rc}(\hat f) - \hat R_{rc}(\hat f) + \hat R_{rc} (\hat f) - R_{rc}(f^{*}) \\
  &\leq R_{rc}(\hat f) - \hat R_{rc}(\hat f) + R_{rc} (\hat f) - R_{rc}(f^{*}) \\
  &\leq 2 \underset{f\in F} \sup | \hat R_{rc}(f) - R_{rc}(f) | \\
  &\leq 4 \mathfrak{R}_{n}(\mathcal{G}_{rc}) + 2M \sqrt{\frac{\log \frac{2}{\delta}}{2 n}} \\
  &\leq 4 \sqrt{2} \rho \sum_{y=1}^{C} \mathfrak{R}_{n} (\mathcal{H}_{y})
  + 2M \sqrt{\frac{\log \frac{2}{\delta}}{2 n}}.
\end{align*}
\hfill $\Box$

\subsection{Proof of \cref{th:cc}}
\label{proof:cc_consistency}

\begin{proof}
First, we set the quantity $\bm{Q} \in \mathbb{R}^{{}_{K}H_{C} \times C}$ as follows:
\begin{align*}
  \bm{Q}_{ij}(X \backslash X^{(k)}) =
  \begin{cases}
    P(T_{i} \backslash j | X \backslash X^{(k)}) & \text{if~} j \in T_{i}, \\
    0 & \text{otherwise}.
  \end{cases}
\end{align*}
Let $\bm{v}_{S} = \left[ P(S = T_{1} | X), \cdots,  P(S = T_{{}_{K}H_{C}} | X) \right]^{\mathsf{T}} \in \mathbb{R}^{{}_{K}H_{C}}$
and $\bm{v}_{y} = \left[ P(y = 1 | X^{(k)}), \cdots,  P(y = C | X^{(k)}) \right]^{\mathsf{T}} \in \mathbb{R}^{C}$.
Then, $\forall k \in \{1, \cdots, C\}$, $\bm{v}_{S}$ and $\bm{v}_{y}$ have the following relationship:
\begin{align*}
  \bm{v}_{S} = \bm{Q} \bm{v}_{y}.
\end{align*}
Furthermore, as a result of \cref{lem:crossent} \cite{Yu_2018_ECCV},
it can be shown that the optimal classifier $q^{*}$ learned with the cross-entropy loss and mean squared error loss satisfies
$q_{i}^{*} = P(S = T_{i} | X)$, and $q^{*}(X) = \bm{v}_{S}$ holds.
Let $e^{*}(X^{(k)})$ denote the softmax output of the classifier that minimizes the $R_{cc}$ as specified in \cref{eq:risk_cc}.
Then $q^{*}(X) = \bm{Q} e^{*}(X^{(k)})$ holds.
Additionally, because $\bm{Q}$ is full column rank, $e^{*}_{i}(X^{(k)}) = P(y = i | X^{(k)})$ holds.
\end{proof}

\subsection{Proof of \cref{th:cc_bound}}
\label{proof:cc_bound}
We derive the estimation error bounds for the classifier-consistent method proposed in \cref{sec:cc}
by following the same procedure used in \cref{proof:cc_bound}.
We begin by defining the function set $\mathcal{G}_{cc}$ as follows:
\begin{align*}
  \mathcal{G}_{cc} = \{(X,S) \mapsto \frac{1}{K}
  \sum_{k} \mathcal{L}(q(s(X^{(k)}, k)), S) | f \in \mathcal{F} \}.
\end{align*}

\begin{lemma}
  \label{lem:cc_1}
  Let $\mathcal{L}$ denote the loss function and let $M$ denote an upper bound on this function.
  Then, $\forall \delta > 0$ with a probability of at least $1 - \delta$, we have
  \begin{align*}
    \underset{f \in \mathcal{F}}\sup |R_{cc}(f) - \hat R_{cc}(f) |
    \leq 2 \mathfrak{R}_{n} (\mathcal{G}_{cc}) + M\sqrt{\frac{\log \frac{2}{\delta}}{2n}}.
  \end{align*}
\end{lemma}

\begin{proof}
  We first consider $\underset{f \in \mathcal{F}}\sup R_{cc}(f) - \hat R_{cc}(f)$.
  As the loss function satisfies the condition $0 \leq \mathcal{L}(x,y) \leq M$,
  replacing one sample $(X,S)$ with another sample $(X^{\prime}, S^{\prime})$ does not result in a change
  in $\underset{f \in \mathcal{F}}\sup R_{cc}(f) - \hat R_{cc}(f)$ exceeding $\frac{M}{n}$.
  Then, using McDiarmid's inequality \cite{mcdiarmid1989method}, we can assert that $\forall \delta > 0$ with a probability of at least $1 - \frac{\delta}{2}$, we have:
  \begin{align*}
    \underset{f \in \mathcal{F}}\sup R_{cc}(f) - \hat R_{cc}(f)
    \leq \mathbb{E}\left[\underset{f \in \mathcal{F}}
    \sup R_{cc}(f) - \hat R_{cc}(f)\right] + M \sqrt{\frac{\log \frac{2}{\delta}}{2n}}.
  \end{align*}
  Furthermore, by utilizing the technique of symmetrization, as described in \citet{Vapnik1998},
  we demonstrate the following bound:
  \begin{align*}
    \mathbb{E}\left[\underset{f \in \mathcal{F}} \sup R_{cc}(f) - \hat R_{cc}(f)\right]
    \leq 2 \mathfrak{R}_{n} (\mathcal{G}_{cc}).
  \end{align*}
  Additionally, we show that the same bounds hold with a probability of at least $1-\frac{\delta}{2}$
  for $\underset{f \in \mathcal{F}}\sup \hat R_{cc}(f) - R_{cc}(f)$.
\end{proof}

\begin{lemma}
  \label{lem:cc_2}
  Assuming that the loss function $\mathcal{L}$ is $\rho$-Lipschitz, that the instances are independently and identically distributed,
  and that $\bm{Q}$ does not depend on the hypothesis $f$, it can be shown that the following holds:
  \begin{align*}
    \mathfrak{R}_{n} (\mathcal{G}_{cc}) \leq \sqrt{2} \rho \sum_{y=1}^{C} \mathfrak{R}_{n}(H_{y}).
  \end{align*}
\end{lemma}
\begin{proof}
  Let $\phi_{X,S} : \mathbb{R}^{C} \rightarrow \mathbb{R}_{+}$ be defined as
  \begin{align*}
    \phi_{X,T_{j}}(\bm{z}) = \mathcal{L}(\bm{Q}[j]^{\mathsf{T}}\bm{z}, T_{j}).
  \end{align*}
  In this case, $\bm{Q}$ has the property that the sum of the elements in each row is less than 1,
  and $\phi_{X,S}$ is a $\rho$-Lipschitz function.
  Therefore, the following holds:
  \begin{align*}
    \mathfrak{R}_{n}(\mathcal{G}_{cc})
    &= \mathbb{E} \left[\underset{g \in \mathcal{G}_{cc}} \sup \sum_{i}^{n} \sigma_{i} g(X_{i}, S_{i}) \right] \\
    &= \mathbb{E} \left[\underset{f \in \mathcal{F}} \sup \sum_{i}^{n} \sigma_{i} \frac{1}{K} \sum_{k}
    \mathcal{L}(q(s(X^{(k)}, k)), S) \right] \\
    &= \mathbb{E} \left[\underset{f \in \mathcal{F}} \sup \sum_{i}^{n} \sigma_{i} \frac{1}{K} \sum_{k} \phi_{X,S} (f(s(X^{(k)}, k))) \right] \\
    &= \frac{1}{K} \sum_{k} \mathbb{E} \left[\underset{f \in \mathcal{F}} \sup \sum_{i}^{n} \sigma_{i} \phi_{X,S} (f(s(X^{(k)}, k))) \right] \\
    &\leq \sqrt{2} \rho \mathbb{E} \left[\underset{f \in \mathcal{F}} \sup \sum_{i}^{n} \sum_{c} \sigma_{ic} f(X^{(1)}) \right] \\
    &= \sqrt{2} \rho \sum_{c} \mathfrak{R}_{n}(\mathcal{H}_{c}).
  \end{align*}
  We use the Rademacher vector contraction inequality \cite{vectorrademachercontraction} in the second line from the end.
\end{proof}

From \cref{lem:cc_1,lem:cc_2}, we obtain the following estimated error bounds:
\begin{align*}
  R_{cc}(\hat f) - R_{cc}(f^{*})
  &\leq R_{cc}(\hat f) - \hat R_{cc}(\hat f) + \hat R_{cc} (\hat f) - R_{cc}(f^{*}) \\
  &\leq R_{cc}(\hat f) - \hat R_{cc}(\hat f) + R_{cc} (\hat f) - R_{cc}(f^{*}) \\
  &\leq 2 \underset{f\in F} \sup | \hat R_{cc}(f) - R_{cc}(f) | \\
  &\leq 4 \mathfrak{R}_{n}(\mathcal{G}_{cc}) + 2M \sqrt{\frac{\log \frac{2}{\delta}}{2 n}} \\
  &\leq 4 \sqrt{2} \rho \sum_{y=1}^{C} \mathfrak{R}_{n} (\mathcal{H}_{y})
  + 2M \sqrt{\frac{\log \frac{2}{\delta}}{2 n}}.
\end{align*}
\hfill $\Box$

\newpage

\section{Experimental Settings}

\subsection{Datasets}
In this study, we conducted experiments with four commonly used datasets.
\begin{itemize}
  \item MNIST \cite{lecun-gradientbased-learning-applied-1998}: 10-class datasets of handwritten digits. Each image is grayscale and has a size of 28 $\times$ 28.
  \item Fashion-MNIST \cite{DBLP:journals/corr/abs-1708-07747}: 10-class datasets of fashion items. Each image is grayscale and has a size of 28 $\times$ 28.
  \item Kuzushiji-MNIST \cite{DBLP:journals/corr/abs-1812-01718}: 10-class datasets of Japanese handwritten kuzushiji, i.e., cursive writing style letters. Each image is grayscale and has a size of 28 $\times$ 28.
  \item CIFAR-10 \cite{Krizhevsky09learningmultiple}: 10-class datasets of vehicles and animals. Each image has an RGB channel and has a size of 32 $\times$ 32 pixels.
\end{itemize}

\subsection{Models}
We used linear and MLP models for MNIST, Fashion-MNIST, Kuzushiji-MNIST, and ConvNet and ResNet models for the CIFAR-10 dataset.
The linear model refers to a $d$–$10$ linear function, where $d$ represents the input size.
The MLP was a 5-layer perceptron $d$–$300$–$300$–$300$–$300$–$10$ with ReLU activation. Each dense layer was subjected to batch normalization.
The ConvNet architecture is described in \cref{tab:convnet}.

\subsection{comparison methods}
We performed comparative experiments with the LLPFC \cite{zhang2022learning}, OT \cite{ijcai2021-377}, and DLLP \cite{ArdehalyC17a} methods.
In this section, we describe the experimental setup---particularly for LLPFC and OT, which had specific parameters.
\textbf{LLPFC~~~} \cite{zhang2022learning}
LLPFC is a method that performs LLP via forward correction of label noise \cite{patrini2017making}.
We implemented the LLPFC-uniform algorithm according to the implementation provided on Github
\footnote{\url{https://github.com/Z-Jianxin/LLPFC}}.
As a specific parameter, the groups were updated every 20 epochs following the reference implementation.
\textbf{OT~~~} \cite{ijcai2021-377}
OT is a method that takes into account the classification loss per instance using pseudo-labels created by optimal transport.
\citet{ijcai2021-377} proposed using per-instance losses with pseudo-labels generated by optimal transport after utilizing per-bag losses.
However, in our study, we implemented OT to update the weights during learning in the same way as the $R_{rc}$ method
to facilitate a direct comparison with our methods.
We implemented optimal transport using the POT \cite{flamary2021pot} Sinkhorn algorithm.
The number of optimization was set to 75, and the entropy constraint factor $\epsilon$ was set as 1, following the guidelines of \cite{abs-1905-12909}.
\input exp_convnet.tex

\section{Experimental Results}
We present the test accuracy for the linear models in \cref{tab:mnists-linear}
and the test accuracy curve for the proposed methods in \cref{fig:test-accuracy-curve2}.

\input exp_fig_ours_tst.tex

\end{document}

%% file: exp_table_1.tex
\begin{table*}[hbtp]
\caption{MNIST, F-MNIST, K-MNIST (Linear)}\label{tab:mnists-linear}\footnotesize\centering
\begin{tabular}{ccccccccc}
\toprule
MNIST &                           K=2 &                             4 &                        8 &                            16 &                            32 &                       64 &                      128 \\
\midrule
Supervised &                $91.1 \pm 0.1$ &                             - &                        - &                             - &                             - &                        - &                        - \\
\midrule
RC         &                $91.3 \pm 0.3$ &                $91.2 \pm 0.1$ &           $91.2 \pm 0.1$ &                           NA &                           NA &                      NA &                      NA \\
RC\_Approx &                $91.3 \pm 0.3$ &                $91.2 \pm 0.2$ &           $90.6 \pm 0.2$ &                $89.6 \pm 0.3$ &                $86.5 \pm 0.9$ &           $69.3 \pm 1.8$ &           $38.7 \pm 3.1$ \\
CC         &                $91.4 \pm 0.4$ &                $91.3 \pm 0.3$ &           $91.1 \pm 0.2$ &                           NA &                           NA &                      NA &                      NA \\
CC\_Approx &                $91.4 \pm 0.4$ &                $91.2 \pm 0.2$ &           $90.5 \pm 0.3$ &                $89.9 \pm 0.2$ &               $77.4 \pm 11.6$ &          $70.9 \pm 14.6$ &  $\mathbf{81.4 \pm 8.6}$ \\
\midrule
DLLP       &                $91.2 \pm 0.2$ &         $90.7 \pm 0.5\bullet$ &    $90.1 \pm 0.5\bullet$ &                $89.8 \pm 0.5$ &        $53.9 \pm 13.6\bullet$ &           $71.5 \pm 9.6$ &          $72.2 \pm 14.9$ \\
LLPFC      &  $\mathbf{92.5 \pm 0.2\circ}$ &  $\mathbf{92.1 \pm 0.2\circ}$ &           $85.8 \pm 5.5$ &               $71.6 \pm 12.1$ &               $80.6 \pm 11.2$ &           $78.9 \pm 6.8$ &    $58.3 \pm 9.3\bullet$ \\
OT         &                $90.6 \pm 0.5$ &                $91.4 \pm 0.3$ &  $\mathbf{91.3 \pm 0.1}$ &  $\mathbf{91.4 \pm 0.2\circ}$ &  $\mathbf{90.1 \pm 0.5\circ}$ &  $\mathbf{79.8 \pm 1.7}$ &    $44.9 \pm 3.4\bullet$ \\
\bottomrule
\end{tabular}

\begin{tabular}{ccccccccc}
\toprule
F-MNIST &                      K=2 &                        4 &                        8 &                            16 &                            32 &                            64 &                           128 \\
\midrule
Supervised &           $82.5 \pm 0.3$ &                        - &                        - &                             - &                             - &                             - &                             - \\
\midrule
RC         &           $82.2 \pm 0.3$ &           $82.4 \pm 0.5$ &           $82.2 \pm 0.7$ &                           NA &                           NA &                           NA &                           NA \\
RC\_Approx &  $\mathbf{82.6 \pm 0.2}$ &           $82.3 \pm 0.2$ &           $81.6 \pm 0.5$ &                $80.1 \pm 0.6$ &                $74.3 \pm 0.6$ &                $62.5 \pm 2.6$ &                $38.3 \pm 5.1$ \\
CC         &           $82.5 \pm 0.2$ &           $82.4 \pm 0.2$ &           $82.0 \pm 0.3$ &                           NA &                           NA &                           NA &                           NA \\
CC\_Approx &           $82.5 \pm 0.2$ &  $\mathbf{82.5 \pm 0.3}$ &           $81.9 \pm 0.3$ &                $81.1 \pm 0.5$ &               $63.4 \pm 10.8$ &               $61.0 \pm 22.4$ &                $19.1 \pm 8.4$ \\
\midrule
DLLP       &           $82.6 \pm 0.2$ &           $82.1 \pm 0.3$ &           $81.2 \pm 0.3$ &         $80.3 \pm 0.2\bullet$ &               $53.0 \pm 15.4$ &               $58.0 \pm 12.0$ &               $17.8 \pm 10.1$ \\
LLPFC      &           $82.4 \pm 0.1$ &    $81.2 \pm 0.2\bullet$ &    $79.1 \pm 0.6\bullet$ &         $77.2 \pm 0.7\bullet$ &                $71.9 \pm 3.9$ &               $61.6 \pm 10.4$ &                $45.1 \pm 6.7$ \\
OT         &           $78.9 \pm 2.1$ &           $81.2 \pm 0.6$ &  $\mathbf{83.1 \pm 0.1}$ &  $\mathbf{82.6 \pm 0.7\circ}$ &  $\mathbf{80.2 \pm 1.4\circ}$ &  $\mathbf{68.6 \pm 1.9\circ}$ &  $\mathbf{49.0 \pm 3.9\circ}$ \\
\bottomrule
\end{tabular}

\begin{tabular}{ccccccccc}
\toprule
K-MNIST &                      K=2 &                        4 &                        8 &                            16 &                            32 &                            64 &                      128 \\
\midrule
Supervised &           $66.3 \pm 0.3$ &                        - &                        - &                             - &                             - &                             - &                        - \\
\midrule
RC         &  $\mathbf{66.5 \pm 0.6}$ &           $66.1 \pm 0.6$ &  $\mathbf{65.4 \pm 0.1}$ &                           NA &                           NA &                           NA &                      NA \\
RC\_Approx &           $66.5 \pm 0.6$ &           $65.9 \pm 0.7$ &           $63.0 \pm 0.4$ &                $59.5 \pm 0.5$ &                $51.0 \pm 1.1$ &                $37.3 \pm 1.1$ &           $25.2 \pm 1.3$ \\
CC         &           $66.2 \pm 0.6$ &  $\mathbf{66.2 \pm 0.7}$ &           $65.0 \pm 0.6$ &                           NA &                           NA &                           NA &                      NA \\
CC\_Approx &           $66.2 \pm 0.6$ &           $66.2 \pm 0.8$ &           $63.0 \pm 0.6$ &                $60.6 \pm 0.8$ &                $36.0 \pm 7.7$ &                $32.9 \pm 5.5$ &  $\mathbf{41.9 \pm 4.9}$ \\
\midrule
DLLP       &           $66.1 \pm 0.4$ &    $64.8 \pm 0.6\bullet$ &    $62.2 \pm 0.6\bullet$ &                $59.1 \pm 1.3$ &         $28.7 \pm 5.5\bullet$ &                $30.4 \pm 8.9$ &           $41.5 \pm 4.3$ \\
LLPFC      &           $66.0 \pm 2.3$ &           $61.1 \pm 5.1$ &           $62.4 \pm 3.8$ &                $53.3 \pm 4.4$ &                $52.0 \pm 3.4$ &  $\mathbf{50.2 \pm 2.4\circ}$ &           $40.0 \pm 5.2$ \\
OT         &           $65.2 \pm 1.0$ &           $65.9 \pm 0.6$ &           $65.3 \pm 0.4$ &  $\mathbf{64.2 \pm 0.5\circ}$ &  $\mathbf{57.0 \pm 0.9\circ}$ &                $39.2 \pm 1.2$ &    $27.6 \pm 0.8\bullet$ \\
\bottomrule
\end{tabular}

\end{table*}

\begin{table*}[hbtp]
\caption{MNIST, F-MNIST, K-MNIST (MLP)}\label{tab:mnists-mlp}\footnotesize\centering
\begin{tabular}{ccccccccc}
\toprule
MNIST &                      K=2 &                        4 &                        8 &                       16 &                       32 &                       64 &                      128 \\
\midrule
Supervised &           $98.5 \pm 0.1$ &                        - &                        - &                        - &                        - &                        - &                        - \\
\midrule
RC         &           $98.6 \pm 0.1$ &           $98.5 \pm 0.1$ &           $98.5 \pm 0.1$ &                      NA &                      NA &                      NA &                      NA \\
RC\_Approx &           $98.6 \pm 0.1$ &           $98.6 \pm 0.1$ &  $\mathbf{98.5 \pm 0.1}$ &  $\mathbf{98.5 \pm 0.2}$ &  $\mathbf{98.5 \pm 0.1}$ &  $\mathbf{98.3 \pm 0.1}$ &           $97.6 \pm 0.1$ \\
CC         &           $98.5 \pm 0.0$ &           $98.5 \pm 0.1$ &           $98.5 \pm 0.1$ &                      NA &                      NA &                      NA &                      NA \\
CC\_Approx &           $98.5 \pm 0.0$ &           $98.6 \pm 0.1$ &           $98.5 \pm 0.1$ &           $98.4 \pm 0.1$ &           $98.3 \pm 0.2$ &           $98.2 \pm 0.1$ &  $\mathbf{97.9 \pm 0.1}$ \\
\midrule
DLLP       &           $98.5 \pm 0.1$ &           $98.5 \pm 0.1$ &           $98.5 \pm 0.1$ &           $98.4 \pm 0.1$ &    $98.3 \pm 0.0\bullet$ &           $98.2 \pm 0.1$ &           $97.8 \pm 0.1$ \\
LLPFC      &    $98.3 \pm 0.1\bullet$ &    $97.8 \pm 0.1\bullet$ &    $97.0 \pm 0.1\bullet$ &    $95.6 \pm 0.3\bullet$ &    $93.1 \pm 0.3\bullet$ &    $90.6 \pm 0.2\bullet$ &    $87.3 \pm 1.0\bullet$ \\
OT         &  $\mathbf{98.6 \pm 0.1}$ &  $\mathbf{98.6 \pm 0.2}$ &           $98.5 \pm 0.1$ &           $98.5 \pm 0.1$ &           $98.5 \pm 0.1$ &    $97.1 \pm 0.1\bullet$ &    $95.5 \pm 0.2\bullet$ \\
\bottomrule
\end{tabular}

\begin{tabular}{ccccccccc}
\toprule
F-MNIST &                      K=2 &                        4 &                        8 &                       16 &                       32 &                       64 &                      128 \\
\midrule
Supervised &           $90.1 \pm 0.2$ &                        - &                        - &                        - &                        - &                        - &                        - \\
\midrule
RC         &           $89.5 \pm 0.4$ &  $\mathbf{89.9 \pm 0.2}$ &  $\mathbf{89.6 \pm 0.1}$ &                      NA &                      NA &                      NA &                      NA \\
RC\_Approx &  $\mathbf{90.0 \pm 0.1}$ &           $89.5 \pm 0.3$ &           $89.5 \pm 0.1$ &  $\mathbf{89.0 \pm 0.2}$ &  $\mathbf{87.7 \pm 0.2}$ &  $\mathbf{86.8 \pm 0.2}$ &  $\mathbf{85.2 \pm 0.4}$ \\
CC         &           $89.5 \pm 0.4$ &           $89.7 \pm 0.1$ &           $89.5 \pm 0.1$ &                      NA &                      NA &                      NA &                      NA \\
CC\_Approx &           $89.5 \pm 0.4$ &           $89.8 \pm 0.1$ &           $89.4 \pm 0.1$ &           $88.7 \pm 0.2$ &           $87.6 \pm 0.2$ &           $86.3 \pm 0.1$ &           $84.9 \pm 0.7$ \\
\midrule
DLLP       &           $89.8 \pm 0.2$ &           $89.7 \pm 0.1$ &    $89.2 \pm 0.2\bullet$ &           $88.5 \pm 0.2$ &           $87.5 \pm 0.1$ &           $86.5 \pm 0.3$ &           $84.7 \pm 0.3$ \\
LLPFC      &    $89.0 \pm 0.4\bullet$ &    $88.2 \pm 0.2\bullet$ &    $86.8 \pm 0.2\bullet$ &    $85.4 \pm 0.2\bullet$ &    $83.3 \pm 0.2\bullet$ &    $81.2 \pm 0.5\bullet$ &    $78.0 \pm 0.7\bullet$ \\
OT         &    $89.6 \pm 0.2\bullet$ &           $89.7 \pm 0.2$ &           $89.4 \pm 0.1$ &           $88.8 \pm 0.1$ &           $87.2 \pm 0.2$ &    $85.0 \pm 0.3\bullet$ &    $82.8 \pm 0.4\bullet$ \\
\bottomrule
\end{tabular}

\begin{tabular}{ccccccccc}
\toprule
K-MNIST &                      K=2 &                        4 &                        8 &                       16 &                       32 &                       64 &                      128 \\
\midrule
Supervised &           $92.7 \pm 0.2$ &                        - &                        - &                        - &                        - &                        - &                        - \\
\midrule
RC         &           $92.8 \pm 0.2$ &           $92.6 \pm 0.3$ &           $92.3 \pm 0.1$ &                      NA &                      NA &                      NA &                      NA \\
RC\_Approx &           $92.8 \pm 0.2$ &  $\mathbf{92.9 \pm 0.2}$ &           $92.6 \pm 0.2$ &  $\mathbf{92.6 \pm 0.2}$ &  $\mathbf{91.9 \pm 0.3}$ &  $\mathbf{90.7 \pm 0.4}$ &           $85.1 \pm 0.4$ \\
CC         &           $92.5 \pm 0.3$ &           $92.7 \pm 0.2$ &           $92.5 \pm 0.3$ &                      NA &                      NA &                      NA &                      NA \\
CC\_Approx &           $92.5 \pm 0.3$ &           $92.6 \pm 0.3$ &           $92.6 \pm 0.2$ &           $92.1 \pm 0.3$ &           $91.3 \pm 0.1$ &           $90.6 \pm 0.4$ &  $\mathbf{88.3 \pm 0.3}$ \\
\midrule
DLLP       &  $\mathbf{92.9 \pm 0.1}$ &           $92.4 \pm 0.4$ &           $92.5 \pm 0.1$ &    $91.9 \pm 0.2\bullet$ &           $91.5 \pm 0.2$ &           $90.4 \pm 0.5$ &           $88.0 \pm 0.5$ \\
LLPFC      &    $91.7 \pm 0.4\bullet$ &    $89.7 \pm 0.3\bullet$ &    $86.7 \pm 0.6\bullet$ &    $81.7 \pm 0.9\bullet$ &    $72.2 \pm 0.8\bullet$ &    $66.9 \pm 1.9\bullet$ &    $61.4 \pm 1.2\bullet$ \\
OT         &           $92.7 \pm 0.3$ &           $92.7 \pm 0.1$ &  $\mathbf{93.0 \pm 0.2}$ &           $92.6 \pm 0.2$ &           $91.7 \pm 0.2$ &    $85.5 \pm 0.4\bullet$ &    $77.9 \pm 1.0\bullet$ \\
\bottomrule
\end{tabular}
\end{table*}

\begin{table*}[hbtp]
\caption{CIFAR-10 (ResNet)}\label{tab:cifar10-resnet}\footnotesize\centering
\begin{tabular}{ccccccccc}
\toprule
CIFAR-10 &                      K=2 &                        4 &                        8 &                       16 &                       32 &                       64 &                      128 \\
\midrule
Supervised &           $81.9 \pm 0.9$ &                        - &                        - &                        - &                        - &                        - &                        - \\
\midrule
RC         &           $81.6 \pm 0.7$ &           $80.2 \pm 0.7$ &           $75.5 \pm 2.0$ &                      NA &                      NA &                      NA &                      NA \\
RC\_Approx &           $81.6 \pm 0.7$ &           $80.1 \pm 0.6$ &           $73.9 \pm 1.2$ &           $65.2 \pm 3.4$ &           $53.7 \pm 2.7$ &           $37.3 \pm 1.7$ &           $30.9 \pm 4.8$ \\
CC         &           $81.6 \pm 0.8$ &           $80.7 \pm 0.4$ &  $\mathbf{78.5 \pm 0.3}$ &                      NA &                      NA &                      NA &                      NA \\
CC\_Approx &           $81.6 \pm 0.8$ &           $80.0 \pm 0.7$ &           $75.1 \pm 0.5$ &  $\mathbf{69.0 \pm 1.0}$ &           $59.3 \pm 2.1$ &           $32.1 \pm 1.7$ &           $22.6 \pm 2.5$ \\
\midrule
DLLP       &           $80.7 \pm 0.7$ &    $79.2 \pm 0.5\bullet$ &    $74.6 \pm 0.2\bullet$ &           $69.0 \pm 0.4$ &  $\mathbf{60.1 \pm 2.1}$ &           $35.5 \pm 2.2$ &           $23.4 \pm 2.4$ \\
LLPFC      &           $80.4 \pm 0.4$ &    $76.8 \pm 0.9\bullet$ &    $71.8 \pm 1.0\bullet$ &    $62.3 \pm 0.9\bullet$ &    $46.0 \pm 3.1\bullet$ &  $\mathbf{41.2 \pm 1.6}$ &  $\mathbf{35.8 \pm 2.3}$ \\
OT         &  $\mathbf{81.9 \pm 0.6}$ &  $\mathbf{81.0 \pm 0.3}$ &    $72.3 \pm 1.8\bullet$ &    $59.4 \pm 3.7\bullet$ &    $48.9 \pm 1.9\bullet$ &           $36.2 \pm 2.4$ &    $28.3 \pm 4.0\bullet$ \\
\bottomrule
\end{tabular}
\end{table*}

\begin{table*}[hbtp]
\caption{CIFAR-10 (ConvNet)}\label{tab:cifar10-convnet}\footnotesize\centering
\begin{tabular}{ccccccccc}
\toprule
CIFAR-10 &                      K=2 &                        4 &                        8 &                       16 &                       32 &                       64 &                      128 \\
\midrule
Supervised &           $86.7 \pm 0.4$ &                        - &                        - &                        - &                        - &                        - &                        - \\
\midrule
RC         &           $86.2 \pm 0.2$ &           $86.2 \pm 0.2$ &           $84.9 \pm 0.3$ &                      NA &                      NA &                      NA &                      NA \\
RC\_Approx &           $86.5 \pm 0.3$ &  $\mathbf{86.6 \pm 0.4}$ &           $85.9 \pm 0.2$ &           $82.2 \pm 0.3$ &           $72.3 \pm 0.8$ &           $59.9 \pm 1.4$ &           $48.0 \pm 1.1$ \\
CC         &           $86.4 \pm 0.3$ &           $86.3 \pm 0.2$ &           $85.4 \pm 0.2$ &                      NA &                      NA &                      NA &                      NA \\
CC\_Approx &           $86.4 \pm 0.3$ &           $86.2 \pm 0.2$ &           $85.4 \pm 0.4$ &  $\mathbf{82.6 \pm 0.3}$ &  $\mathbf{77.0 \pm 0.1}$ &  $\mathbf{69.2 \pm 0.8}$ &  $\mathbf{50.0 \pm 2.6}$ \\
\midrule
DLLP       &  $\mathbf{86.6 \pm 0.3}$ &           $86.1 \pm 0.1$ &           $85.2 \pm 0.2$ &           $82.3 \pm 0.3$ &           $76.9 \pm 0.4$ &           $68.6 \pm 0.7$ &           $48.3 \pm 1.8$ \\
LLPFC      &    $84.7 \pm 0.4\bullet$ &    $81.7 \pm 0.4\bullet$ &    $76.7 \pm 0.3\bullet$ &    $68.5 \pm 0.6\bullet$ &    $60.0 \pm 1.2\bullet$ &    $50.3 \pm 1.5\bullet$ &    $41.7 \pm 0.8\bullet$ \\
OT         &           $86.5 \pm 0.3$ &           $86.1 \pm 0.2$ &  $\mathbf{85.9 \pm 0.2}$ &    $79.4 \pm 0.6\bullet$ &    $66.4 \pm 0.8\bullet$ &    $55.1 \pm 0.5\bullet$ &    $45.2 \pm 1.1\bullet$ \\
\bottomrule
\end{tabular}

\end{table*}

%% file: exp_fig_llpfc.tex
\begin{figure*}[!t]
  \subfigure{
    \begin{minipage}[b]{0.24\columnwidth}
      \includegraphics[width=\columnwidth]{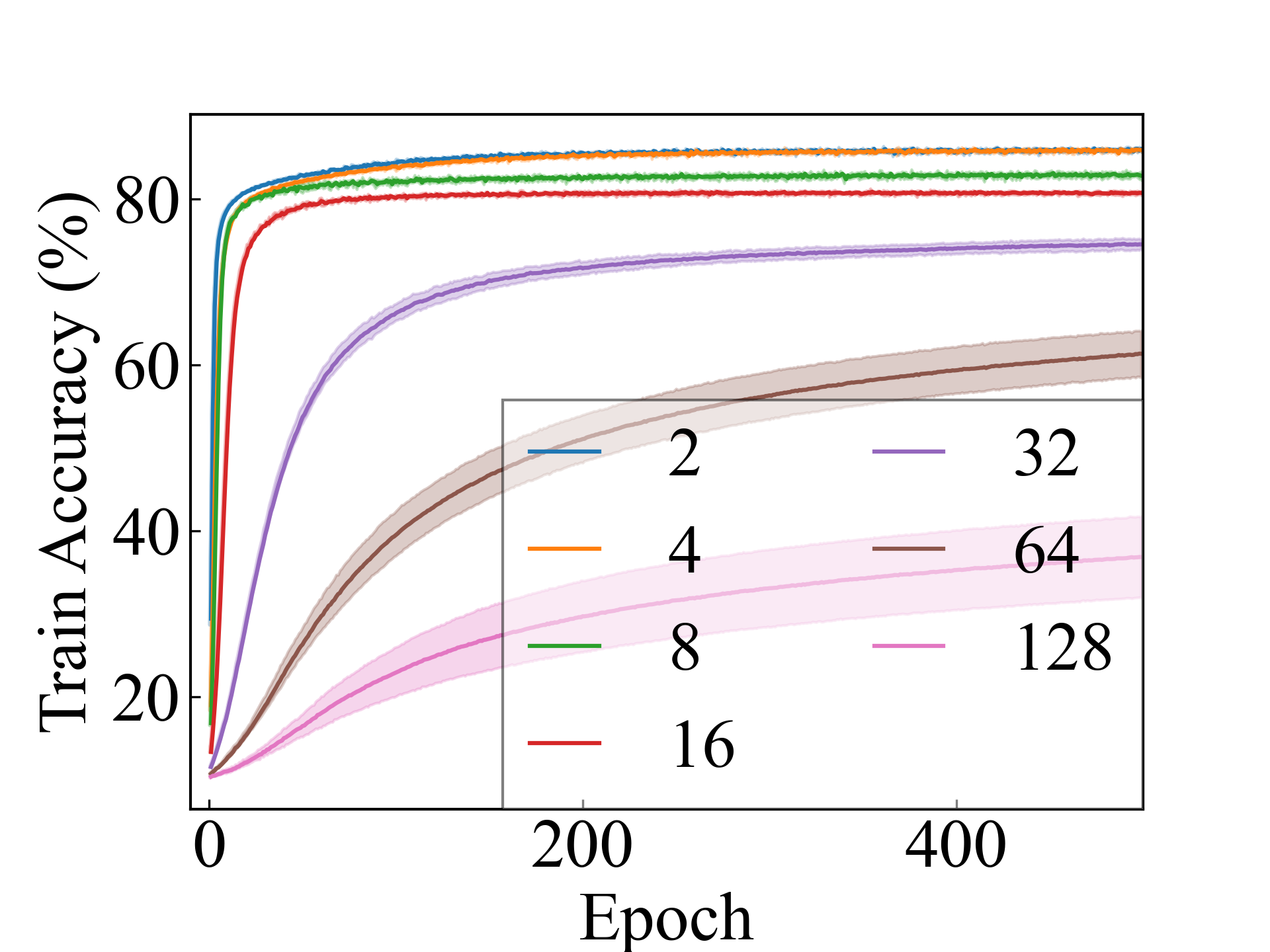}
      \centerline{\quad Linear, RC\_Approx}
  \end{minipage}}
  \subfigure{
    \begin{minipage}[b]{0.24\columnwidth}
      \includegraphics[width=\columnwidth]{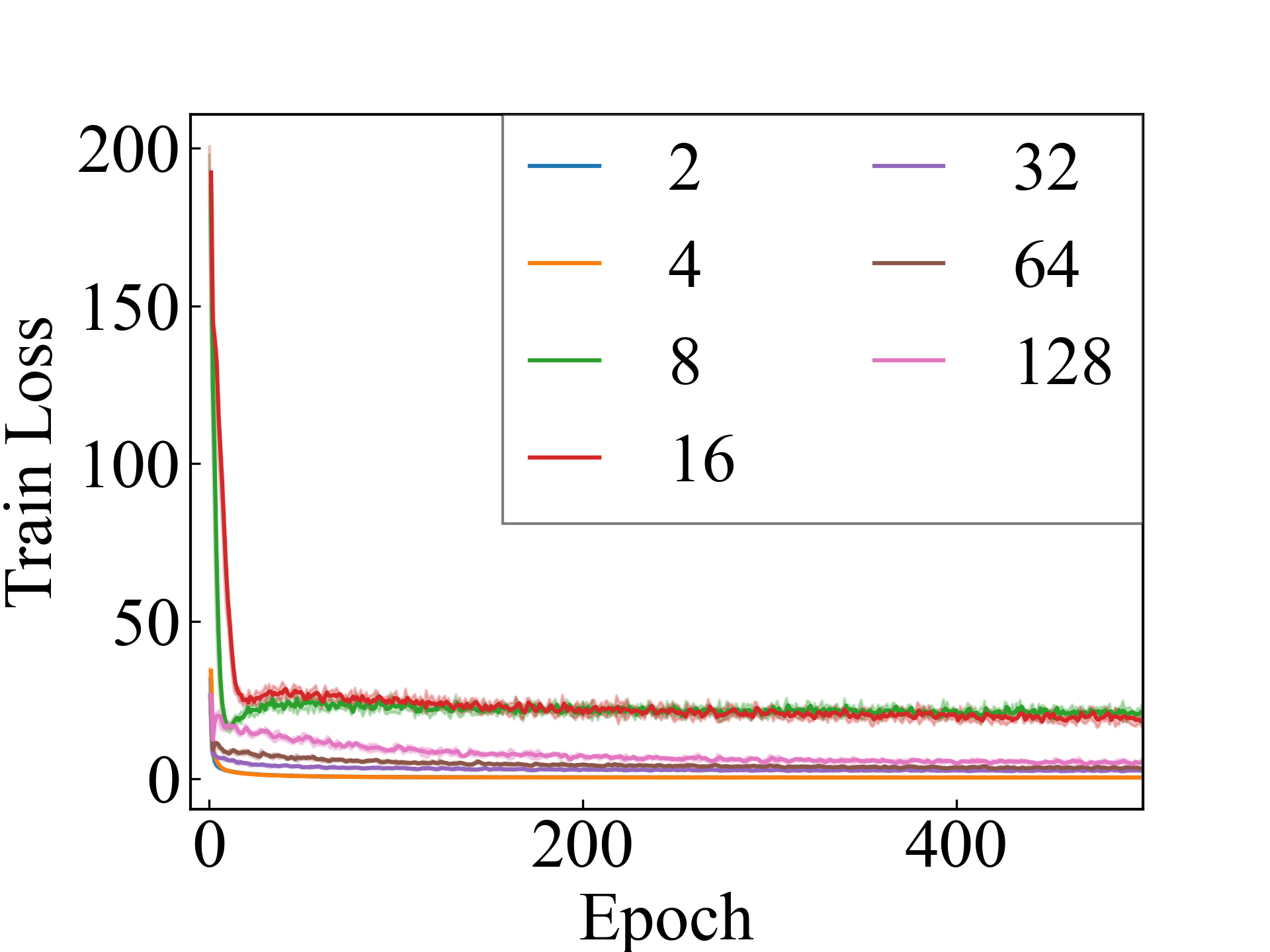}
      \centerline{\quad Linear, RC\_Approx}
  \end{minipage}}
  \subfigure{
    \begin{minipage}[b]{0.24\columnwidth}
      \includegraphics[width=\columnwidth]{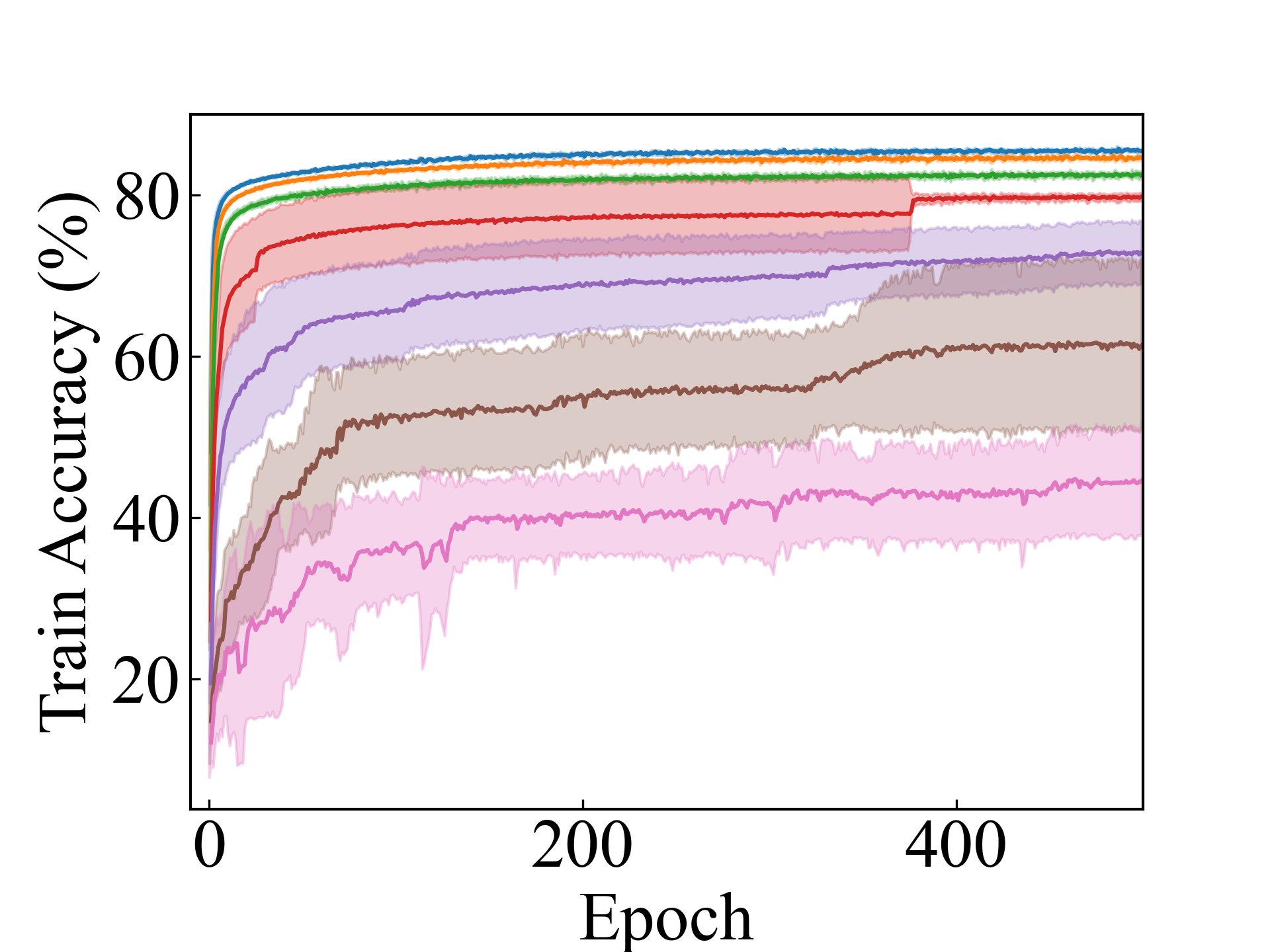}
      \centerline{\quad Linear, LLPFC}
  \end{minipage}}
  \subfigure{
    \begin{minipage}[b]{0.24\columnwidth}
      \includegraphics[width=\columnwidth]{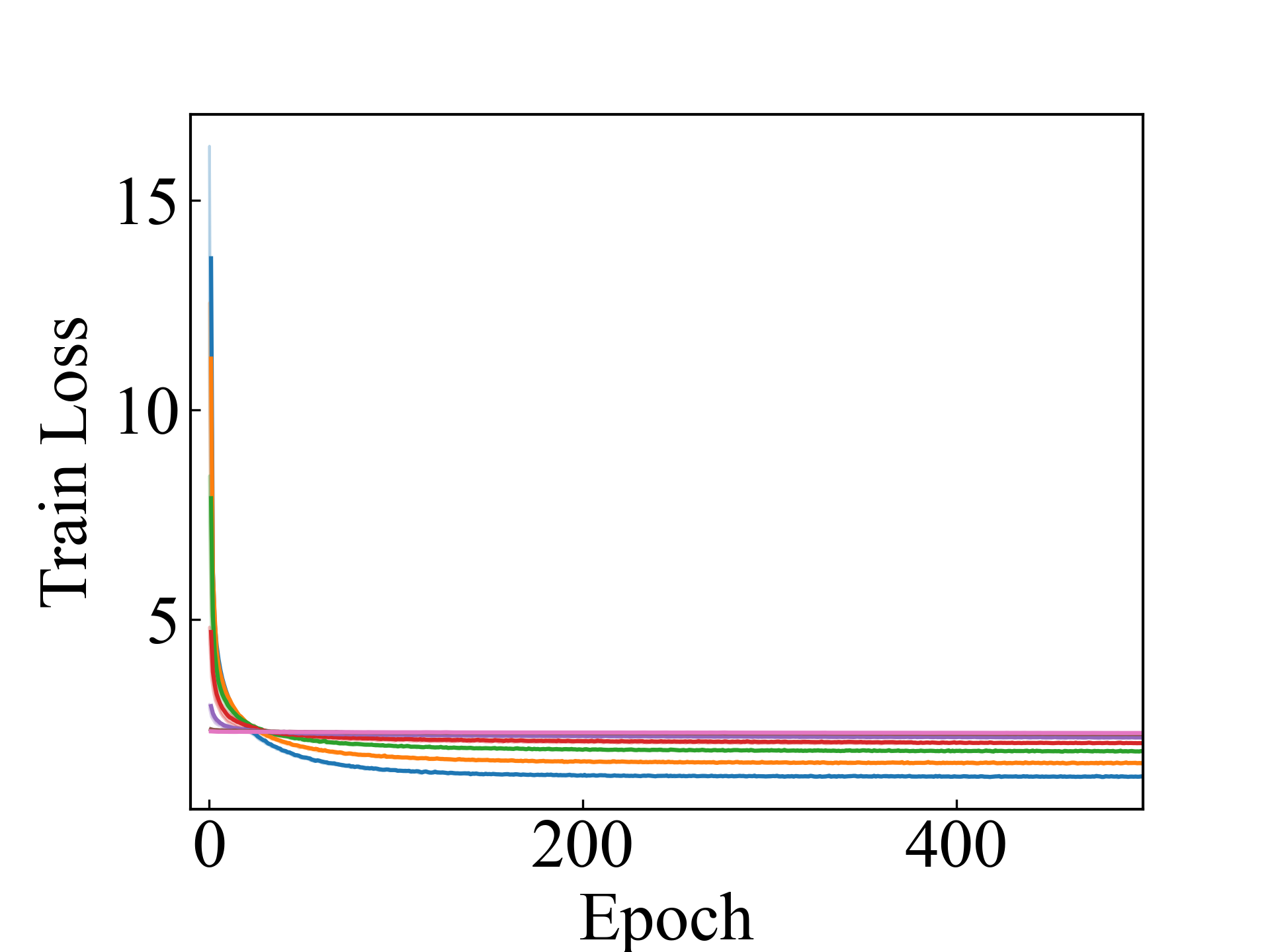}
      \centerline{\quad Linear, LLPFC}
  \end{minipage}}
  \subfigure{
    \begin{minipage}[b]{0.24\columnwidth}
      \includegraphics[width=\columnwidth]{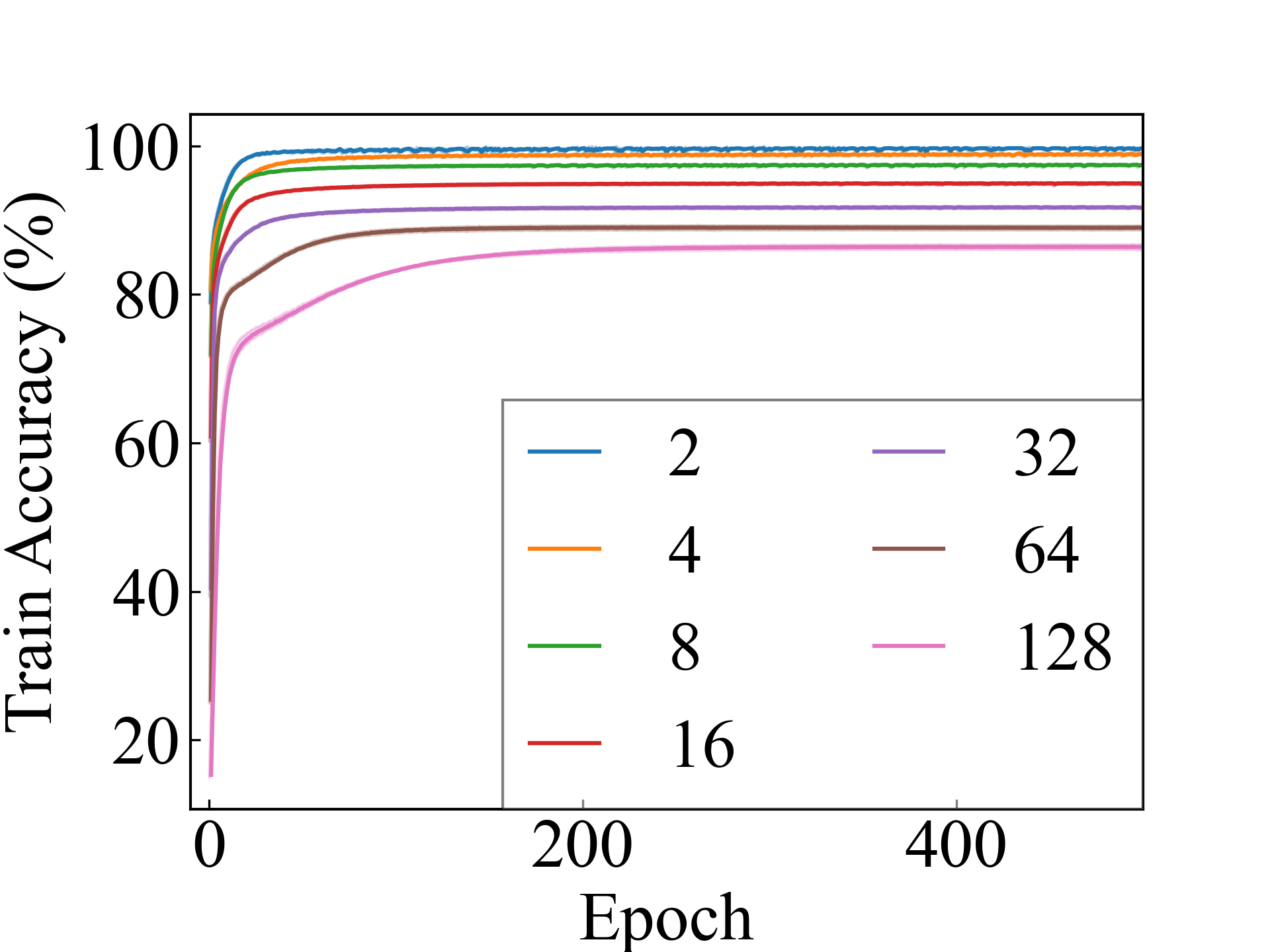}
      \centerline{\quad MLP, RC\_Approx}
  \end{minipage}}
  \subfigure{
    \begin{minipage}[b]{0.24\columnwidth}
      \includegraphics[width=\columnwidth]{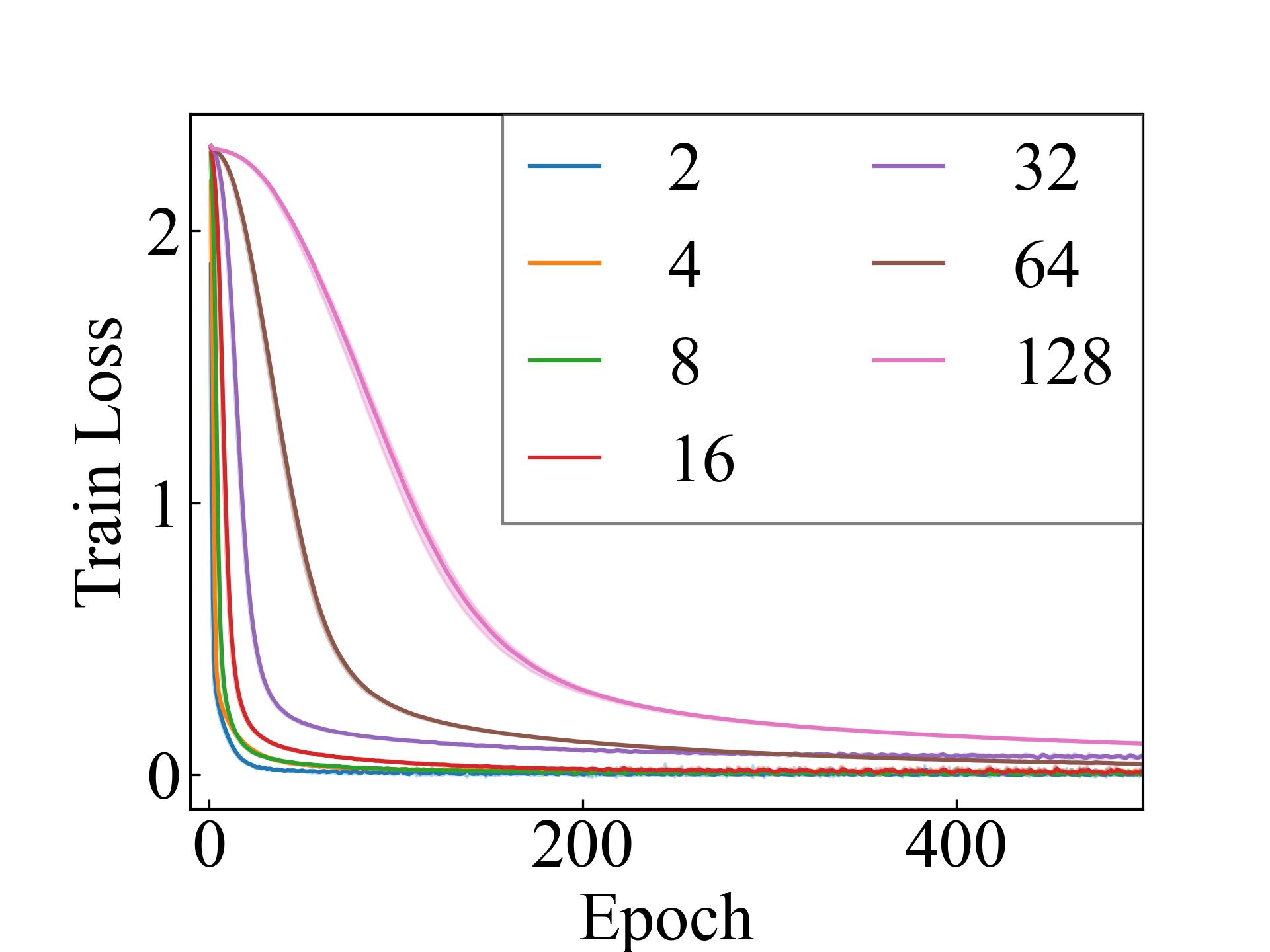}
      \centerline{\quad MLP, RC\_Approx}
  \end{minipage}}
  \subfigure{
    \begin{minipage}[b]{0.24\columnwidth}
      \includegraphics[width=\columnwidth]{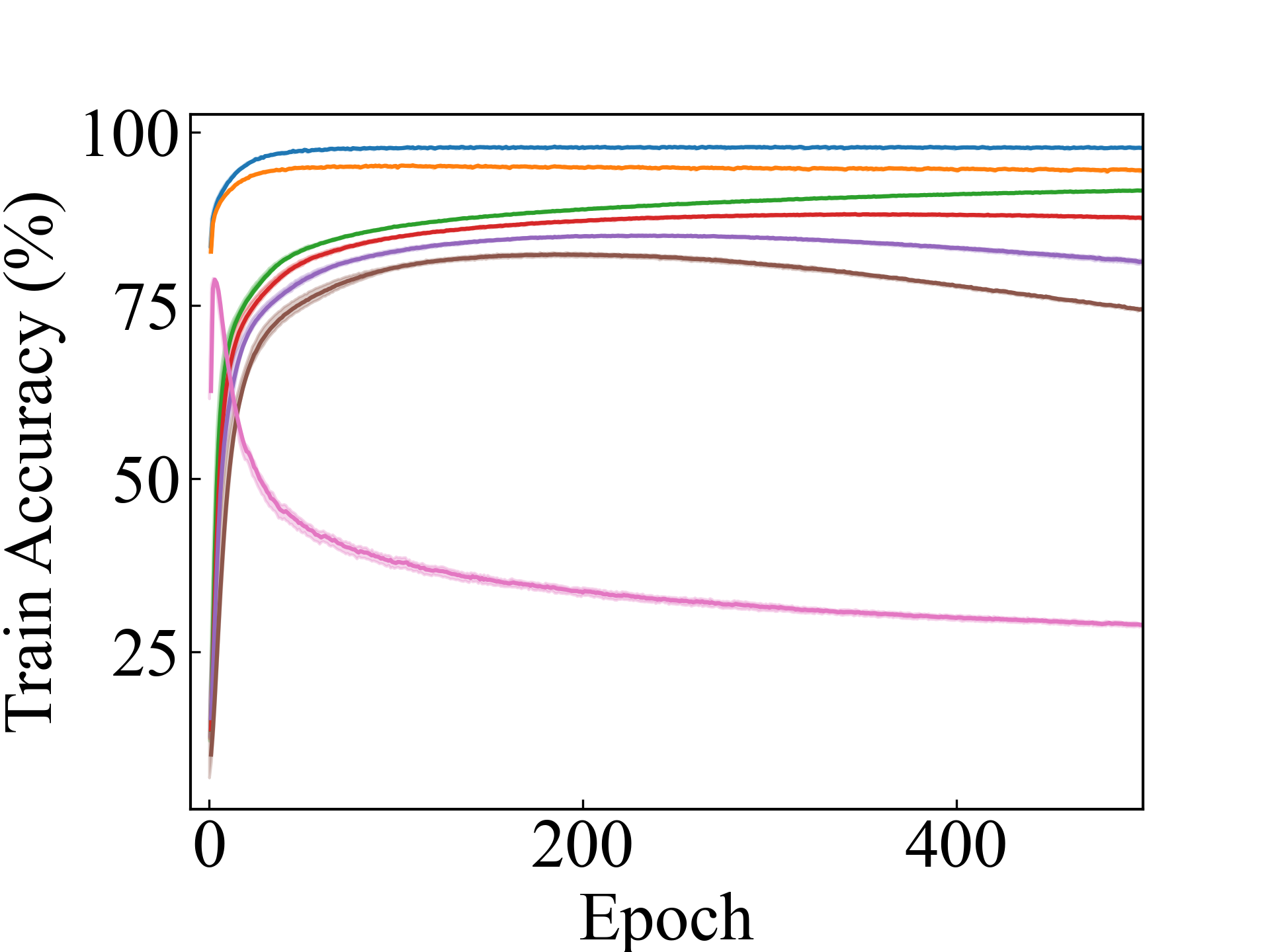}
      \centerline{\quad MLP, LLPFC}
  \end{minipage}}
  \subfigure{
    \begin{minipage}[b]{0.24\columnwidth}
      \includegraphics[width=\columnwidth]{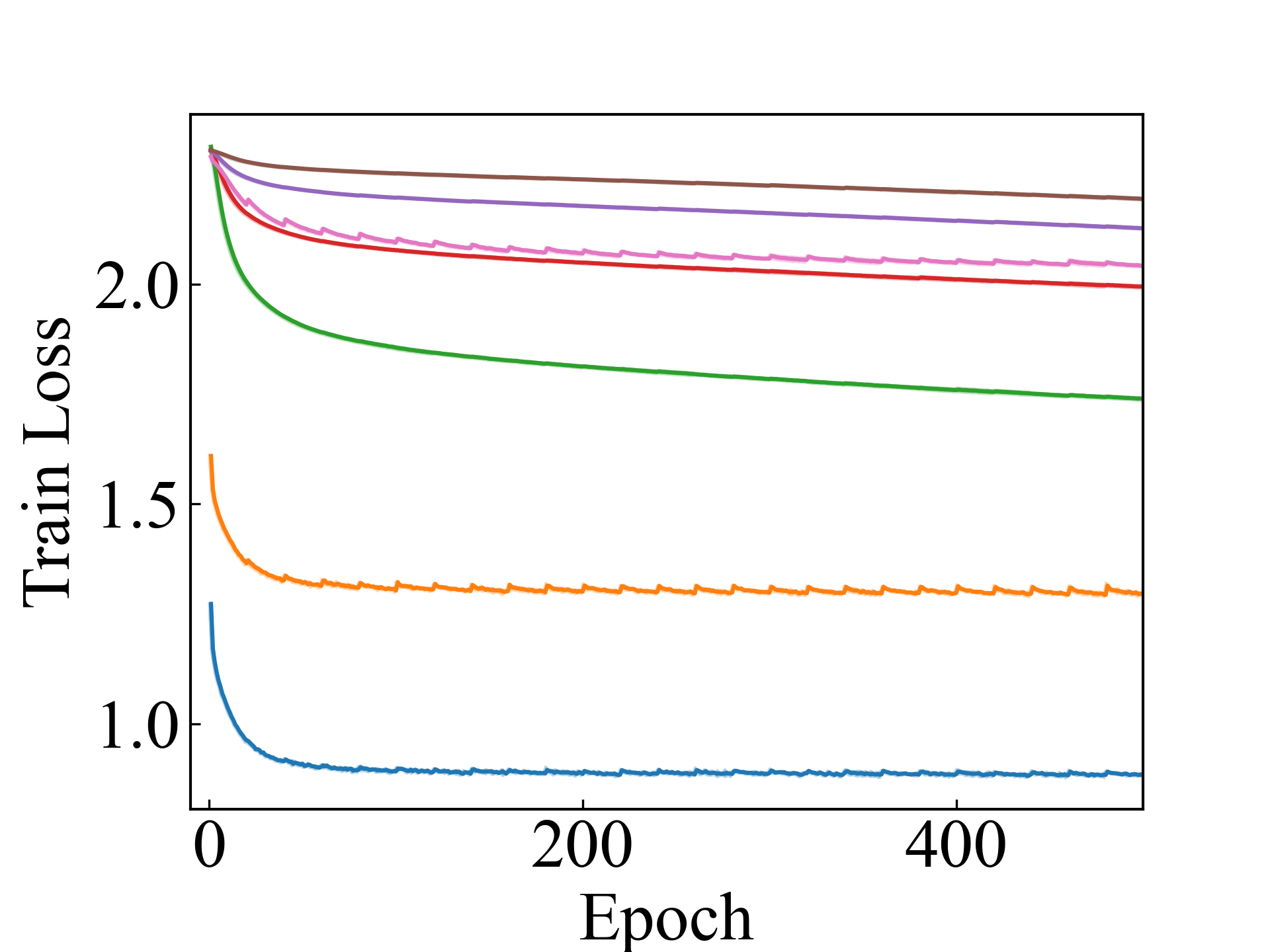}
      \centerline{\quad MLP, LLPFC}
  \end{minipage}}
  \subfigure{
    \begin{minipage}[b]{0.24\columnwidth}
      \includegraphics[width=\columnwidth]{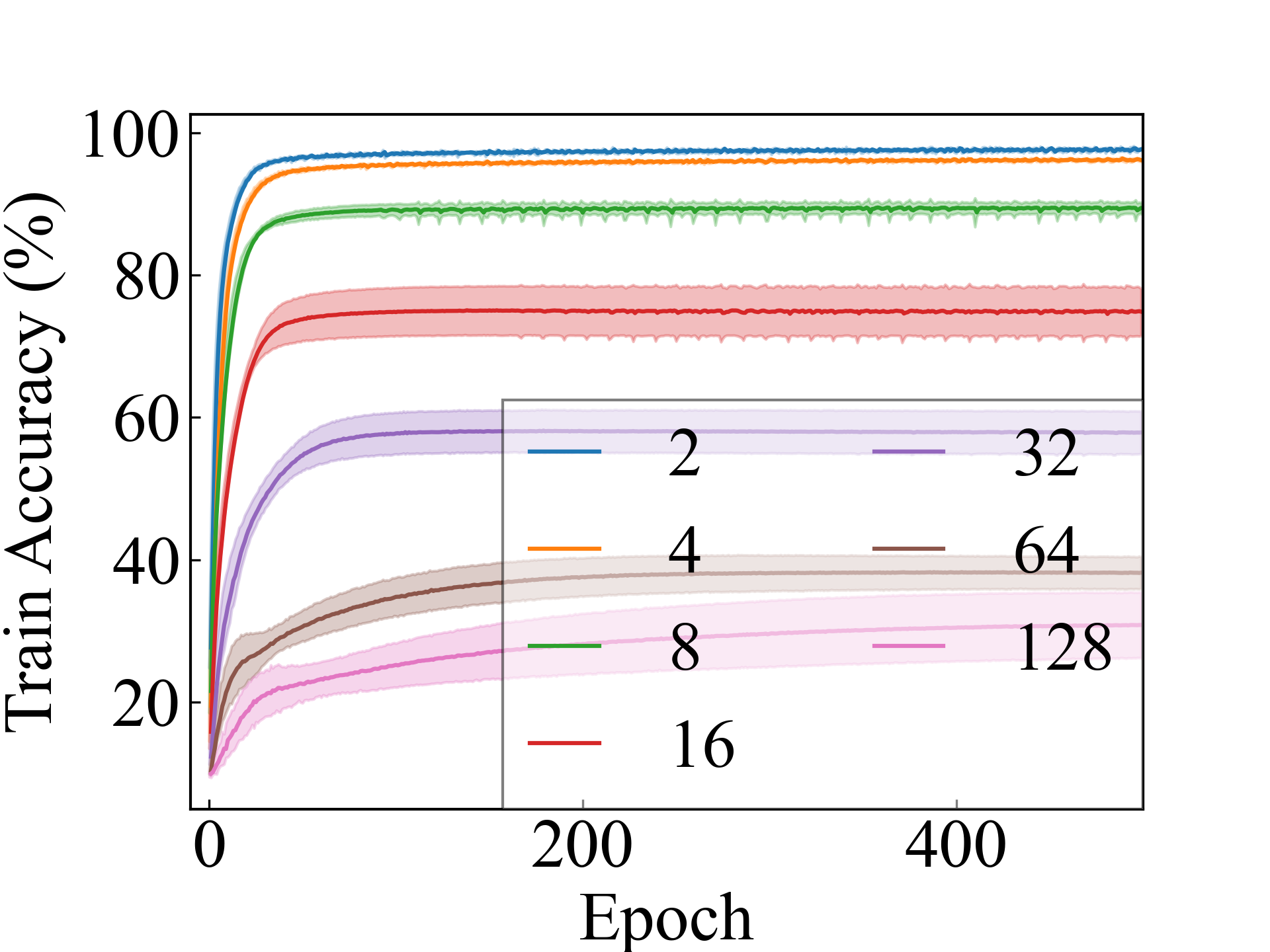}
      \centerline{\quad ResNet, RC\_Approx}
  \end{minipage}}
  \subfigure{
    \begin{minipage}[b]{0.24\columnwidth}
      \includegraphics[width=\columnwidth]{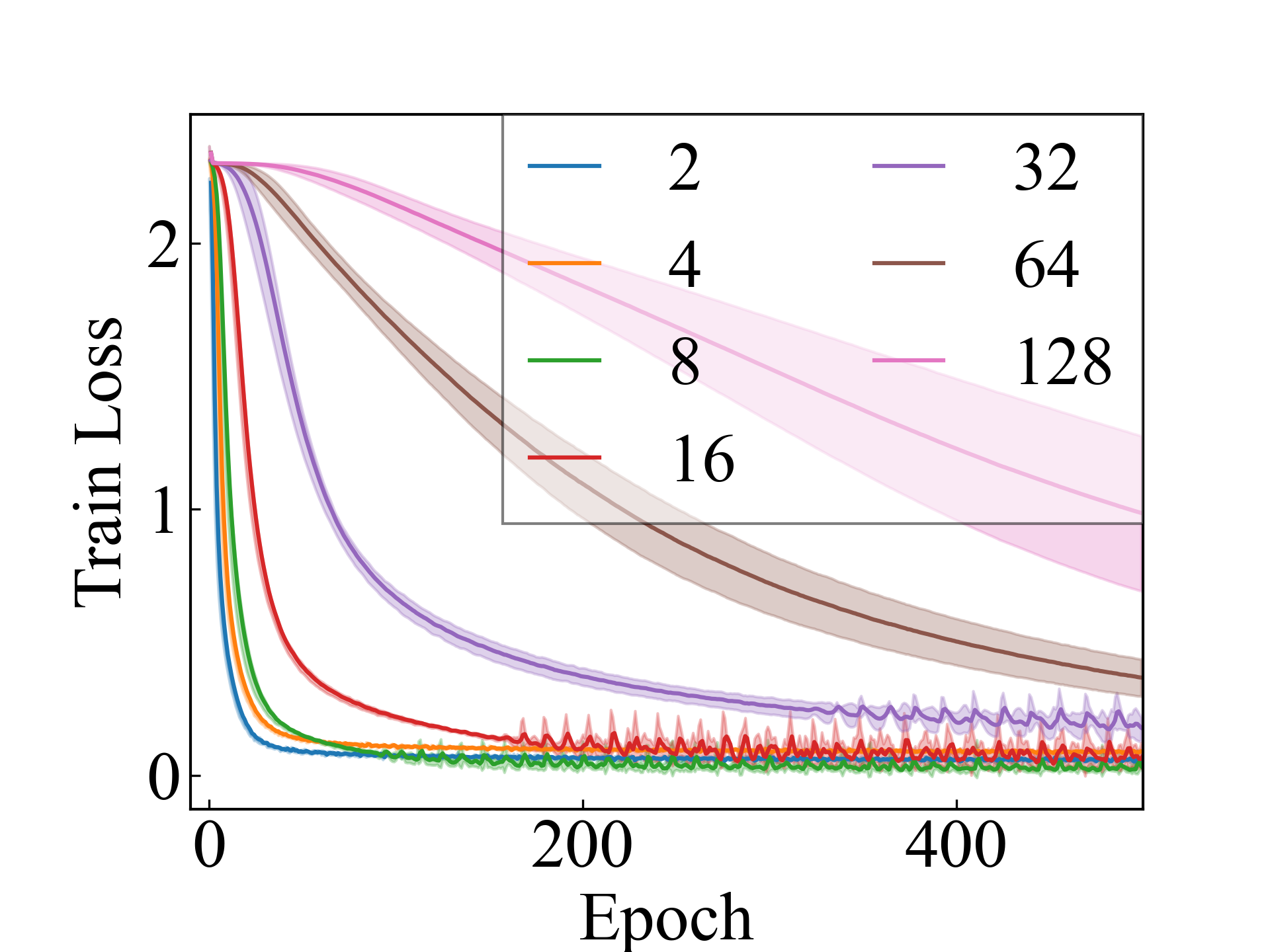}
      \centerline{\quad ResNet, RC\_Approx}
  \end{minipage}}
  \subfigure{
    \begin{minipage}[b]{0.24\columnwidth}
      \includegraphics[width=\columnwidth]{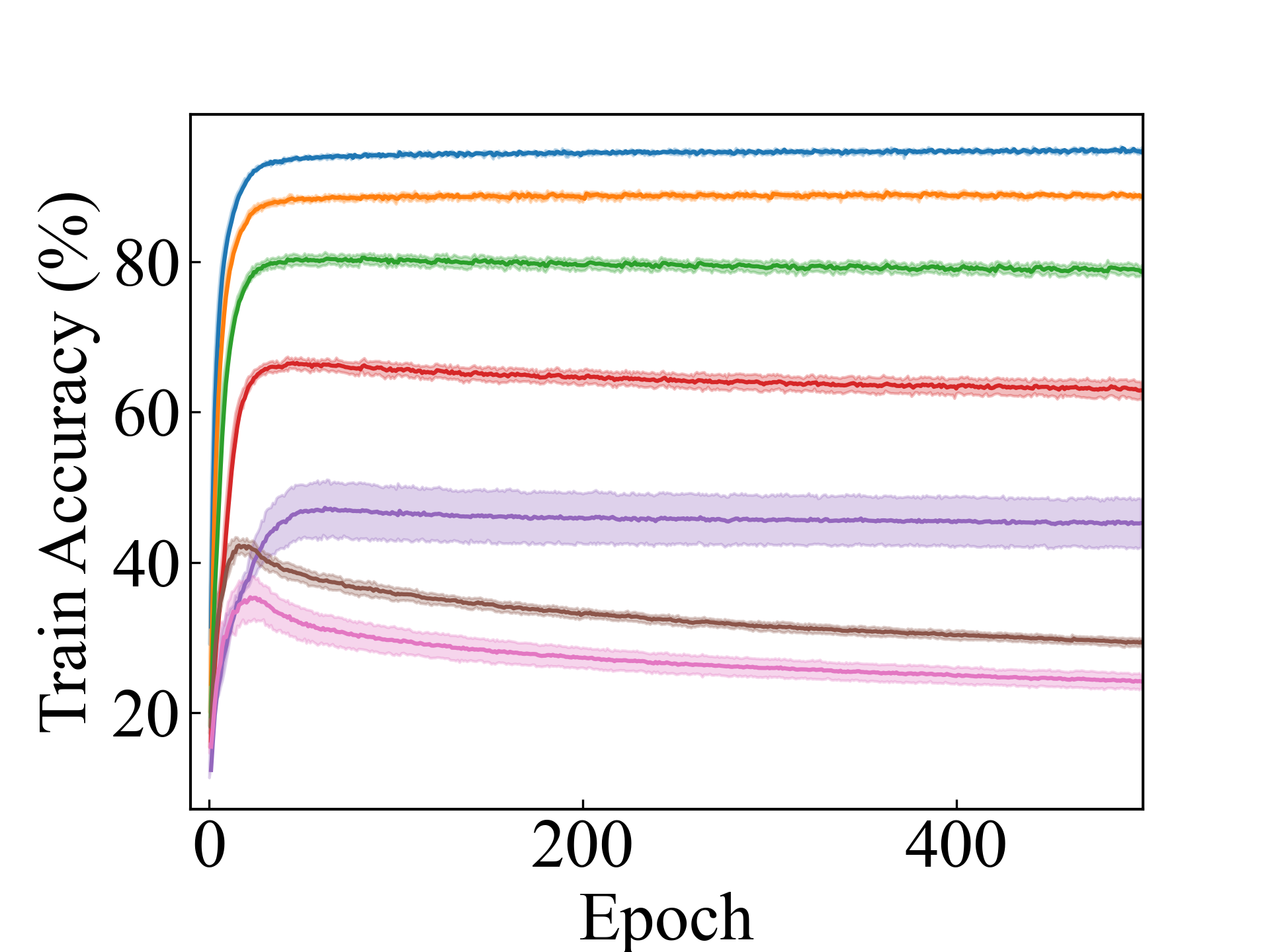}
      \centerline{\quad ResNet, LLPFC}
  \end{minipage}}
  \subfigure{
    \begin{minipage}[b]{0.24\columnwidth}
      \includegraphics[width=\columnwidth]{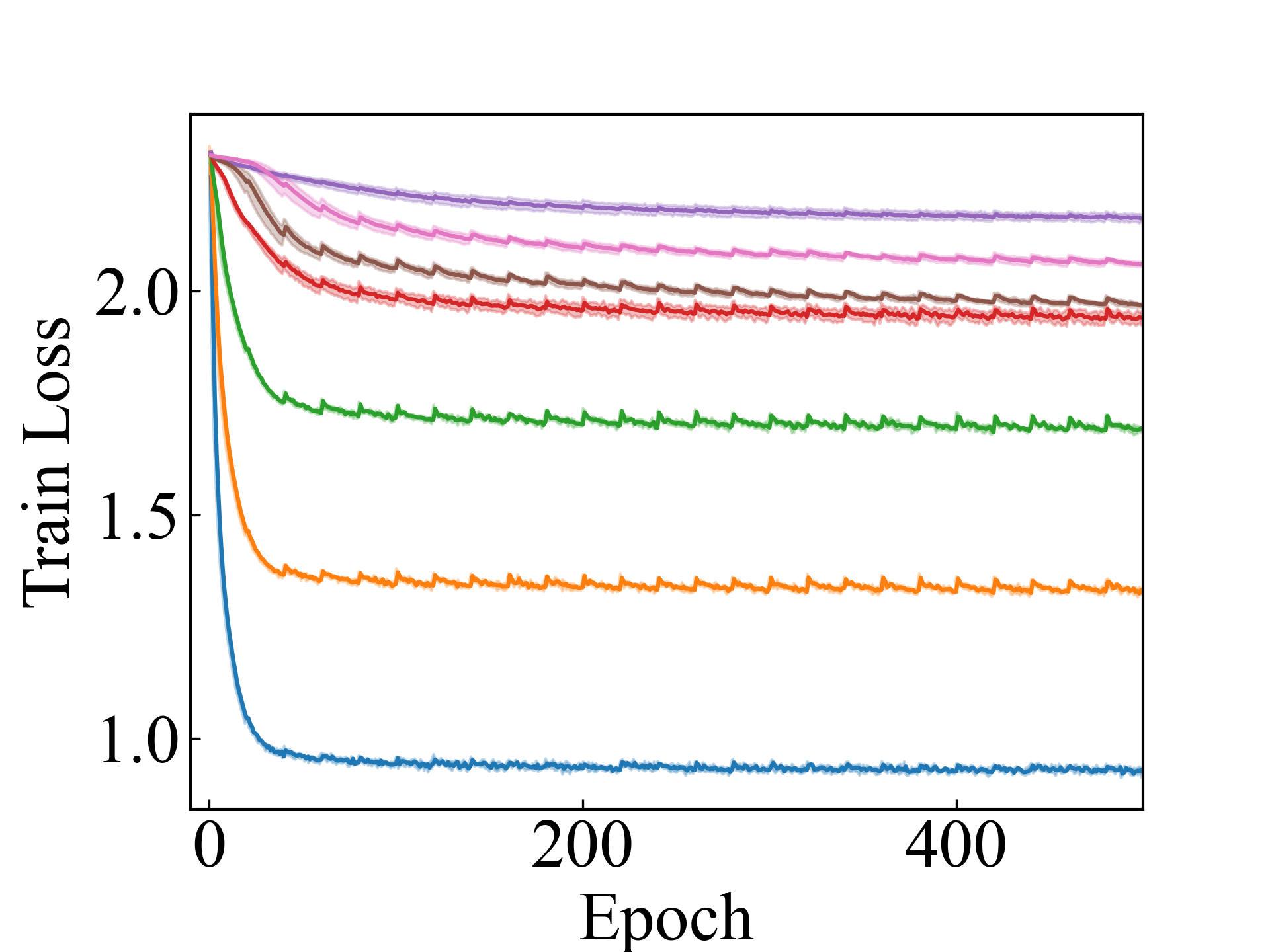}
      \centerline{\quad ResNet, LLPFC}
  \end{minipage}}
  \subfigure{
    \begin{minipage}[b]{0.24\columnwidth}
      \includegraphics[width=\columnwidth]{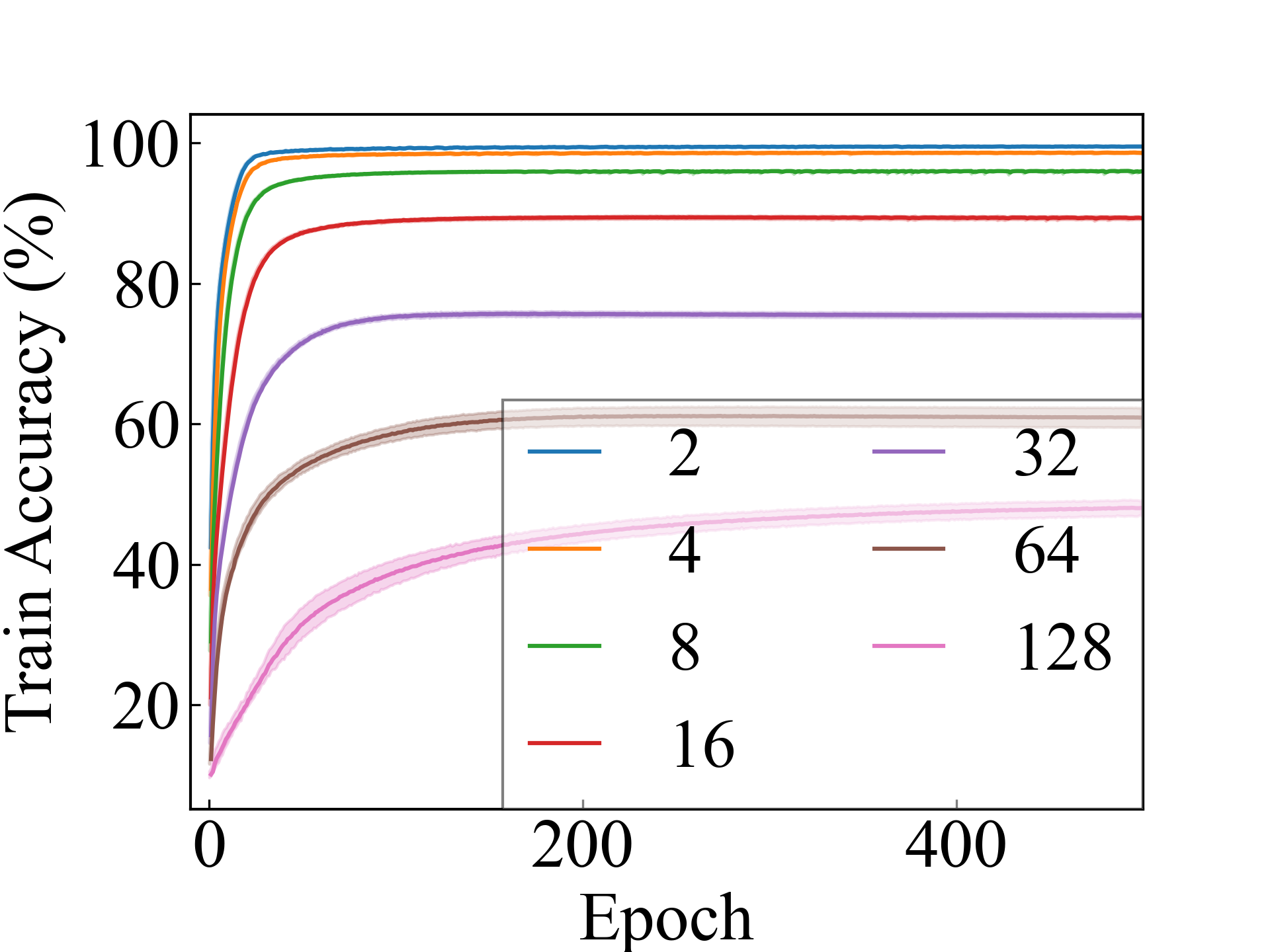}
      \centerline{\quad ConvNet, RC\_Approx}
  \end{minipage}}
  \subfigure{
    \begin{minipage}[b]{0.24\columnwidth}
      \includegraphics[width=\columnwidth]{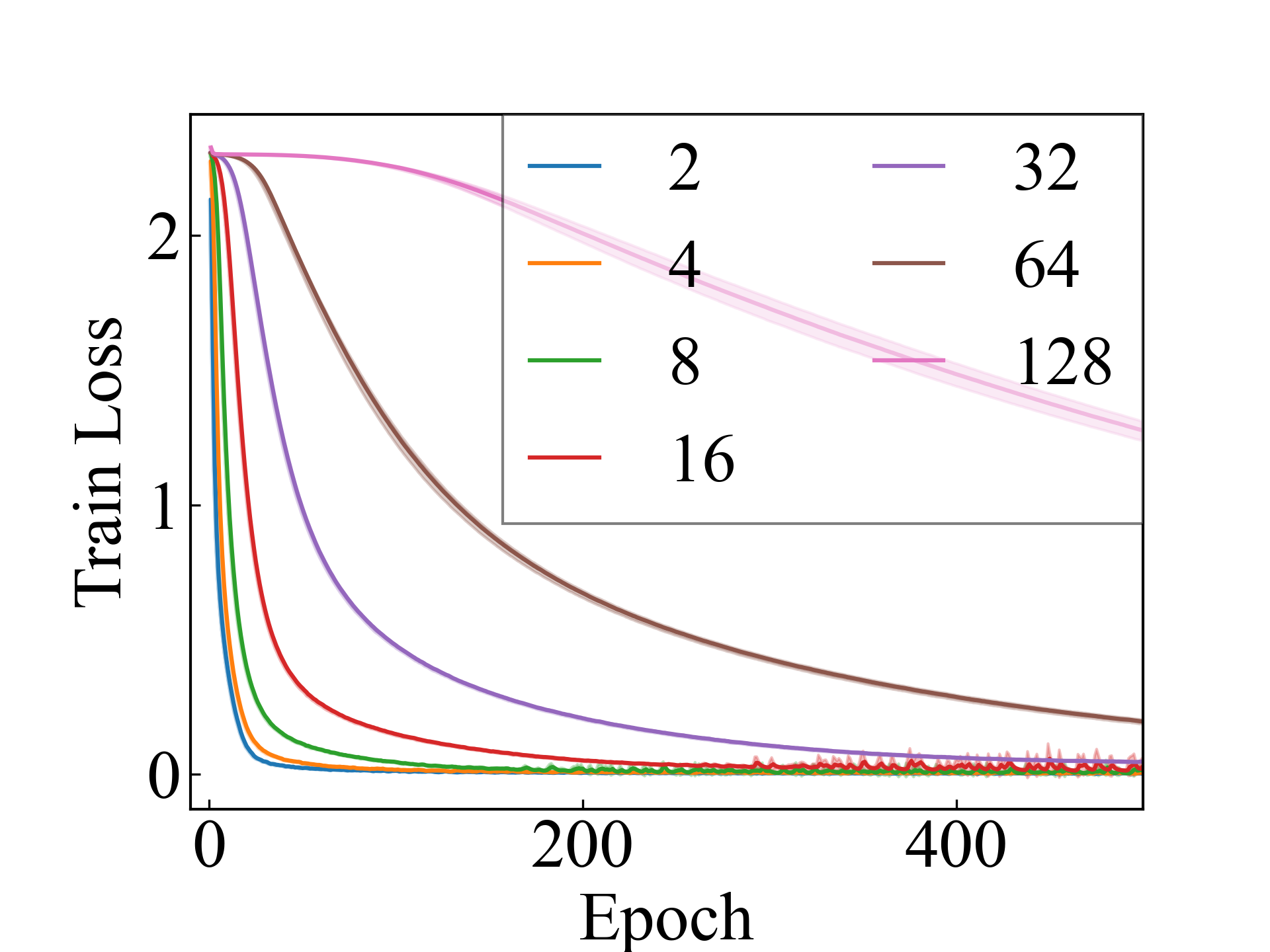}
      \centerline{\quad ConvNet, RC\_Approx}
  \end{minipage}}
  \subfigure{
    \begin{minipage}[b]{0.24\columnwidth}
      \includegraphics[width=\columnwidth]{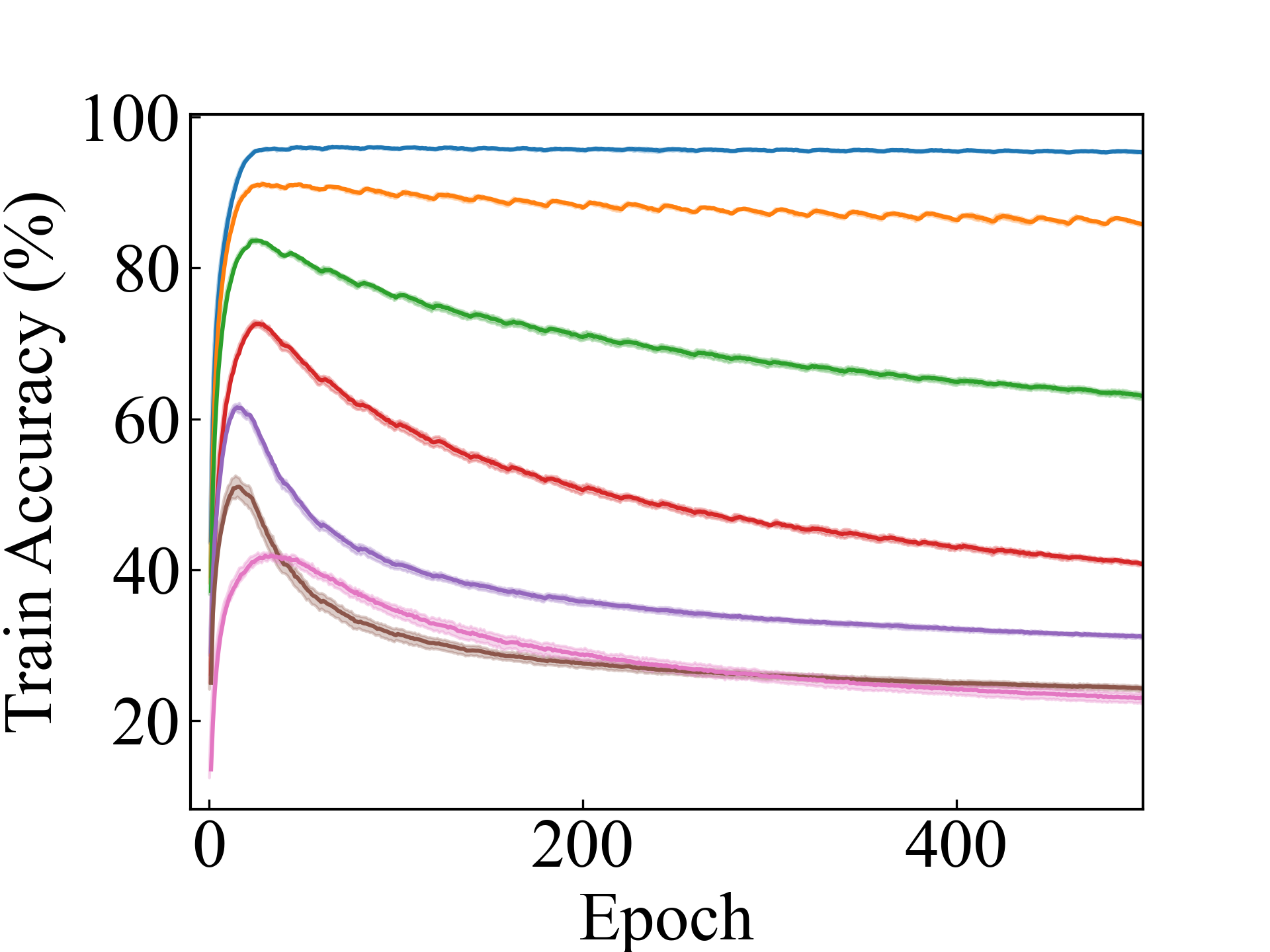}
      \centerline{\quad ConvNet, LLPFC}
  \end{minipage}}
  \subfigure{
    \begin{minipage}[b]{0.24\columnwidth}
      \includegraphics[width=\columnwidth]{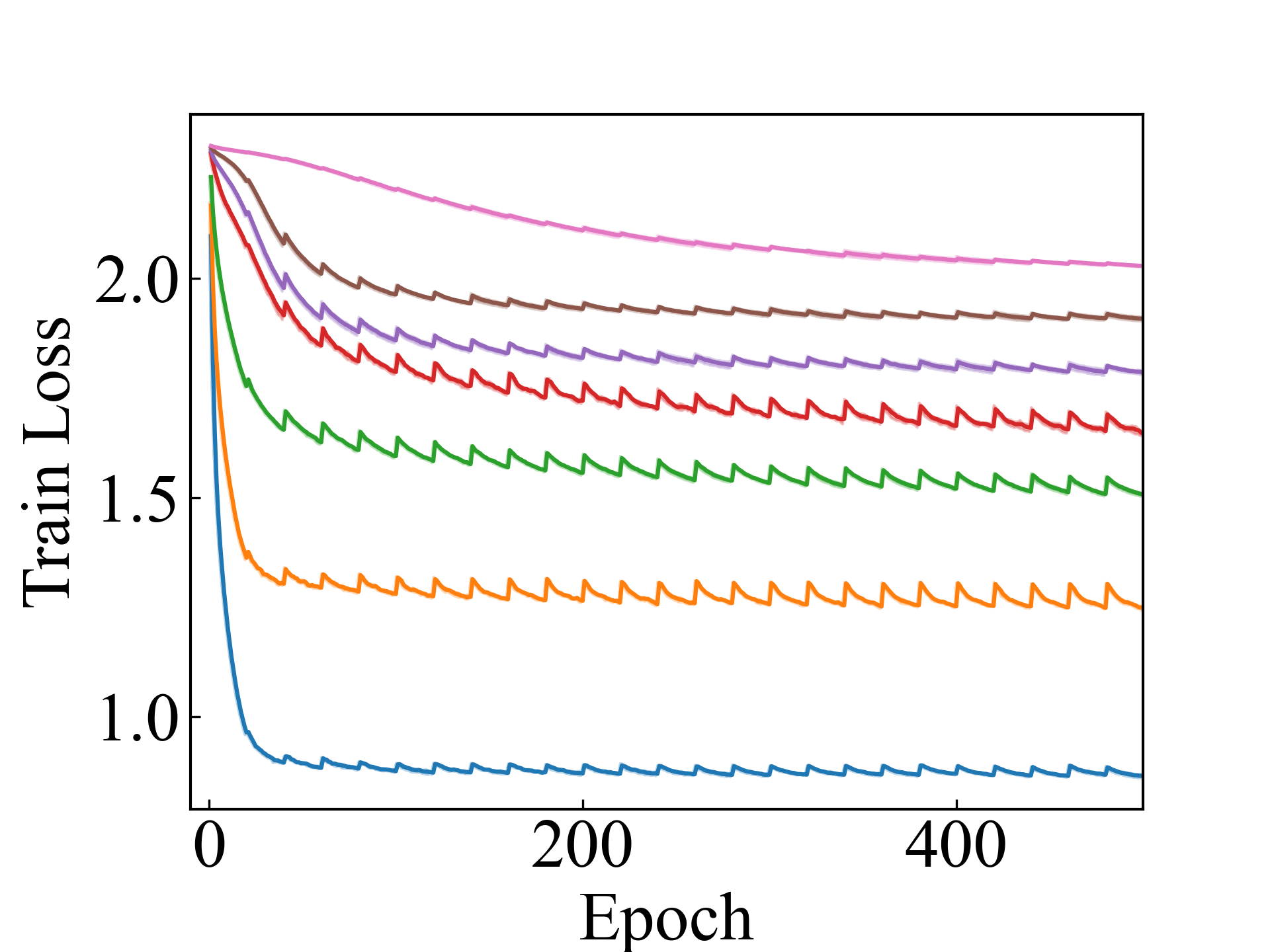}
      \centerline{\quad ConvNet, LLPFC}
  \end{minipage}}
  \caption{
    Train accuracy for Linear model (F-MNIST) and ConvNet (CIFAR-10) with proposed RC\_Approx and LLPFC.
  }
  \label{fig:train-accuracy-llpfc}
\end{figure*}

%% file: exp_fig_ot.tex
\begin{figure*}
  \centering
  \subfigure{
    \begin{minipage}[b]{0.32\columnwidth}
      \includegraphics[width=\columnwidth]{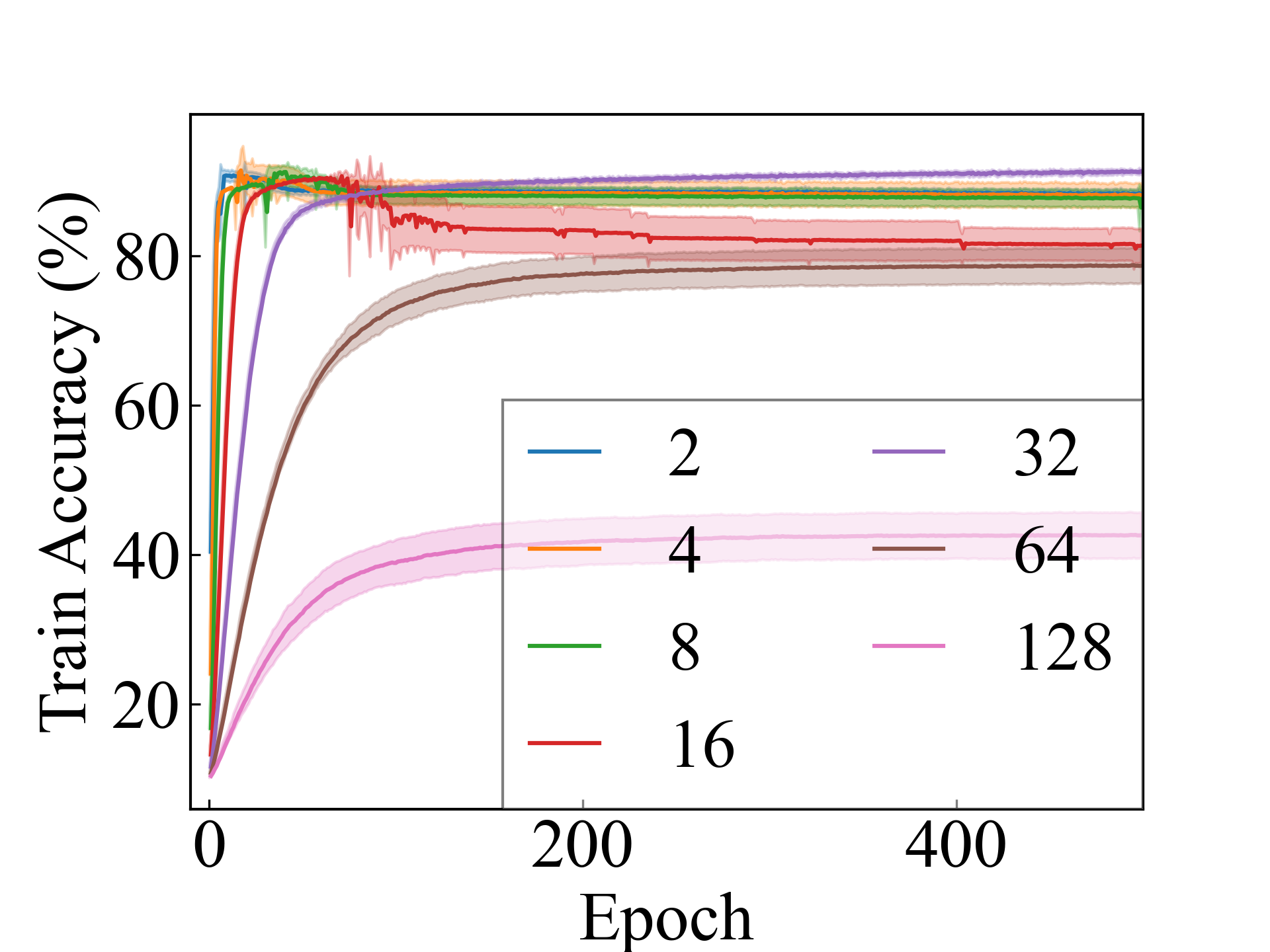}
      \centerline{\quad MNIST, Linear}
  \end{minipage}}
  \subfigure{
    \begin{minipage}[b]{0.32\columnwidth}
      \includegraphics[width=\columnwidth]{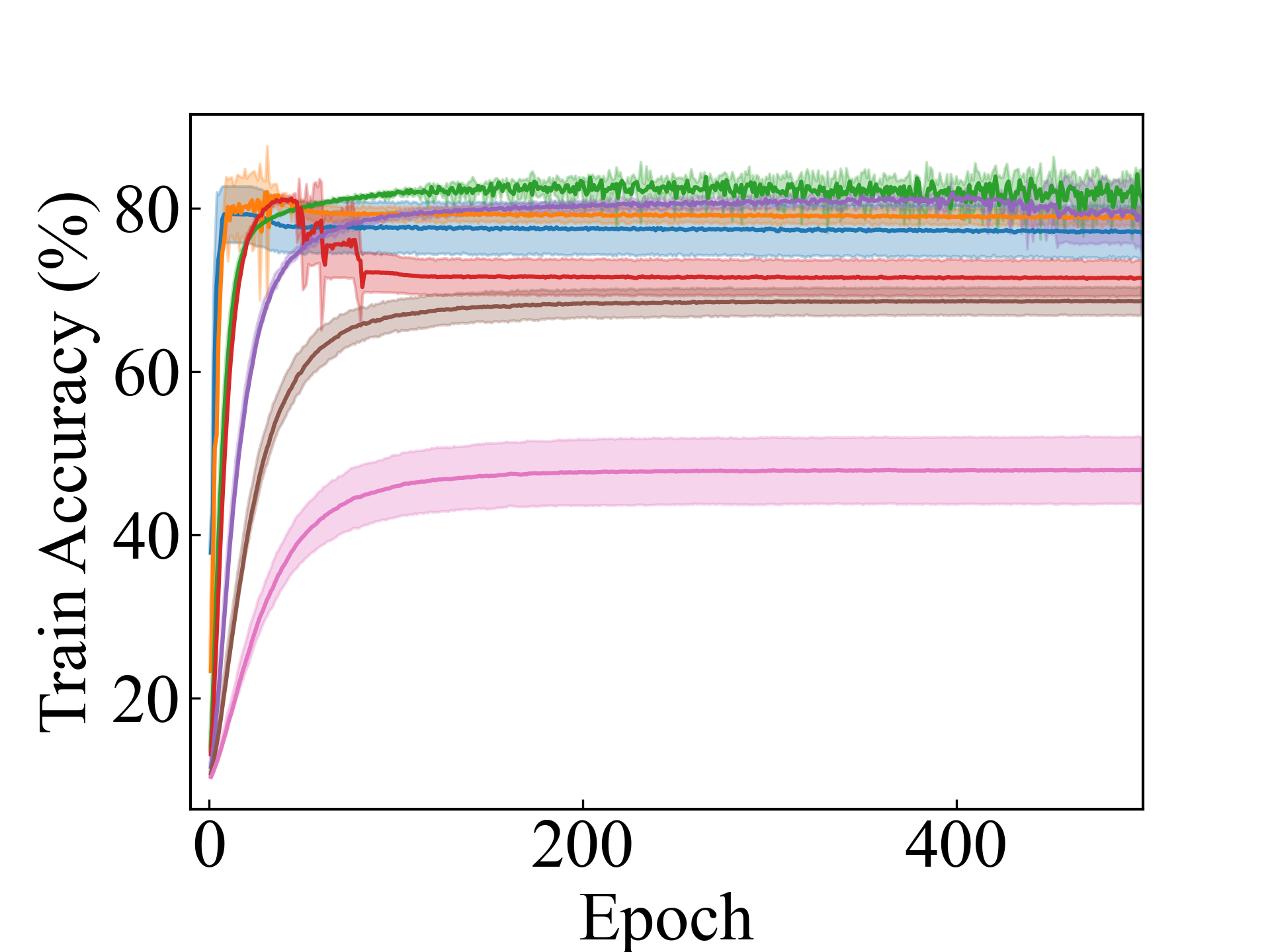}
      \centerline{\quad F-MNIST, Linear}
  \end{minipage}}
  \subfigure{
    \begin{minipage}[b]{0.32\columnwidth}
      \includegraphics[width=\columnwidth]{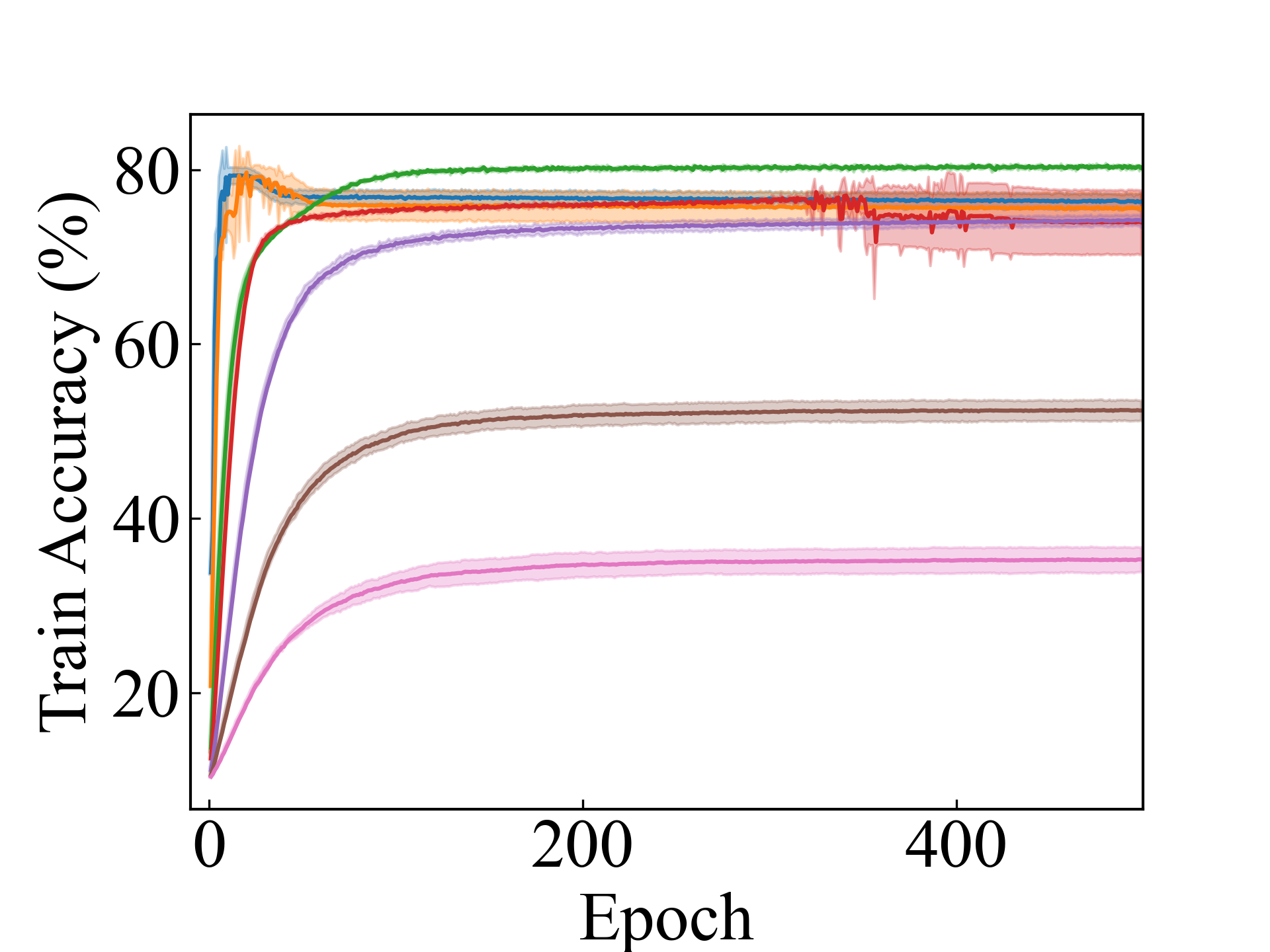}
      \centerline{\quad K-MNIST, Linear}
  \end{minipage}}
  \caption{
  Train accuracy for Linear model (F-MNIST) and with OT method.
  }
  \label{fig:train-accuracy-curve-ot}
\end{figure*}

%% file: exp_fig_weight_diff.tex
\begin{figure*}[!t]
  \subfigure{
    \begin{minipage}[b]{0.24\columnwidth}
      \includegraphics[width=\columnwidth]{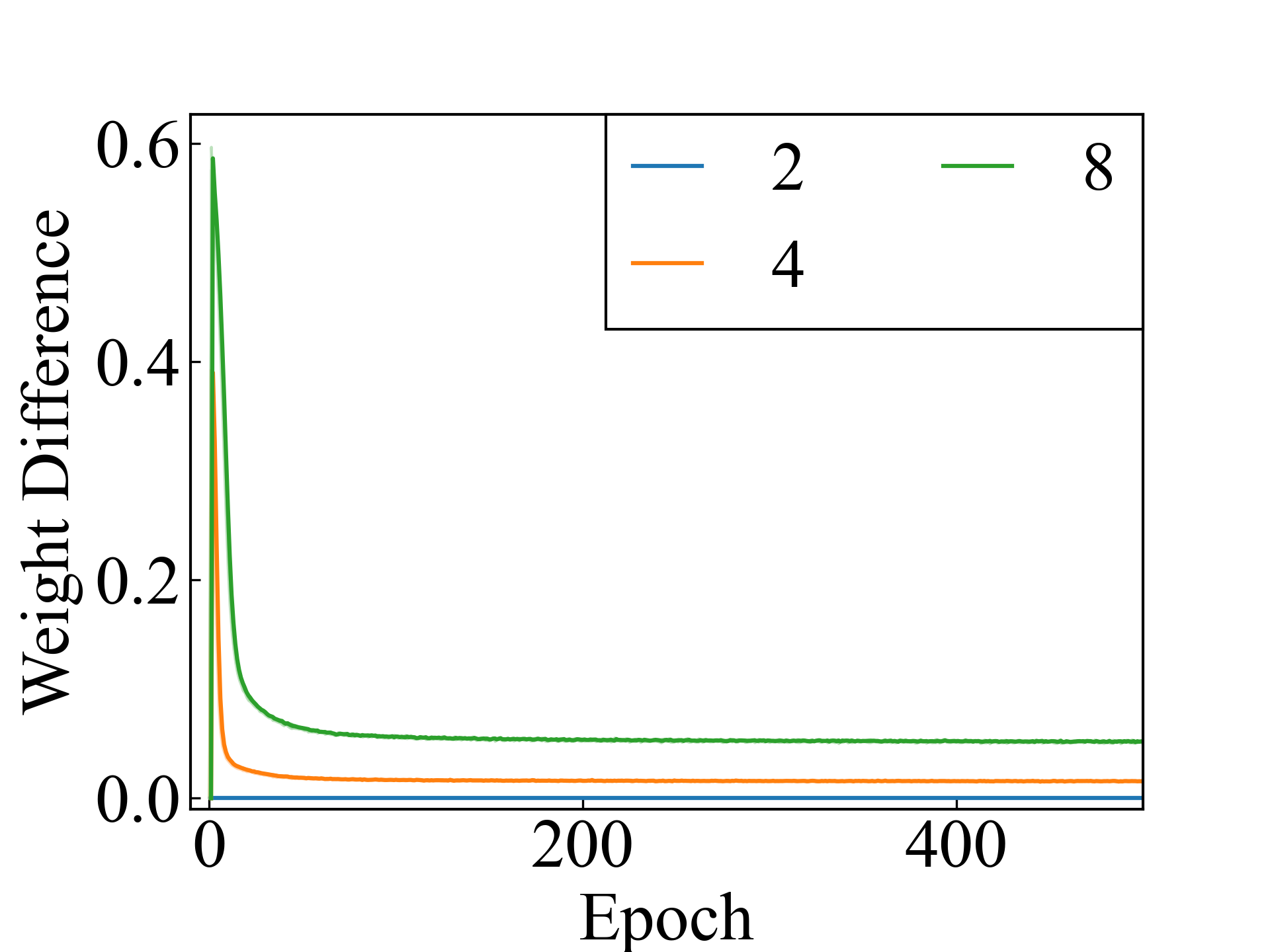}
      \centerline{\quad MNIST, Linear}
  \end{minipage}}
  \subfigure{
    \begin{minipage}[b]{0.24\columnwidth}
      \includegraphics[width=\columnwidth]{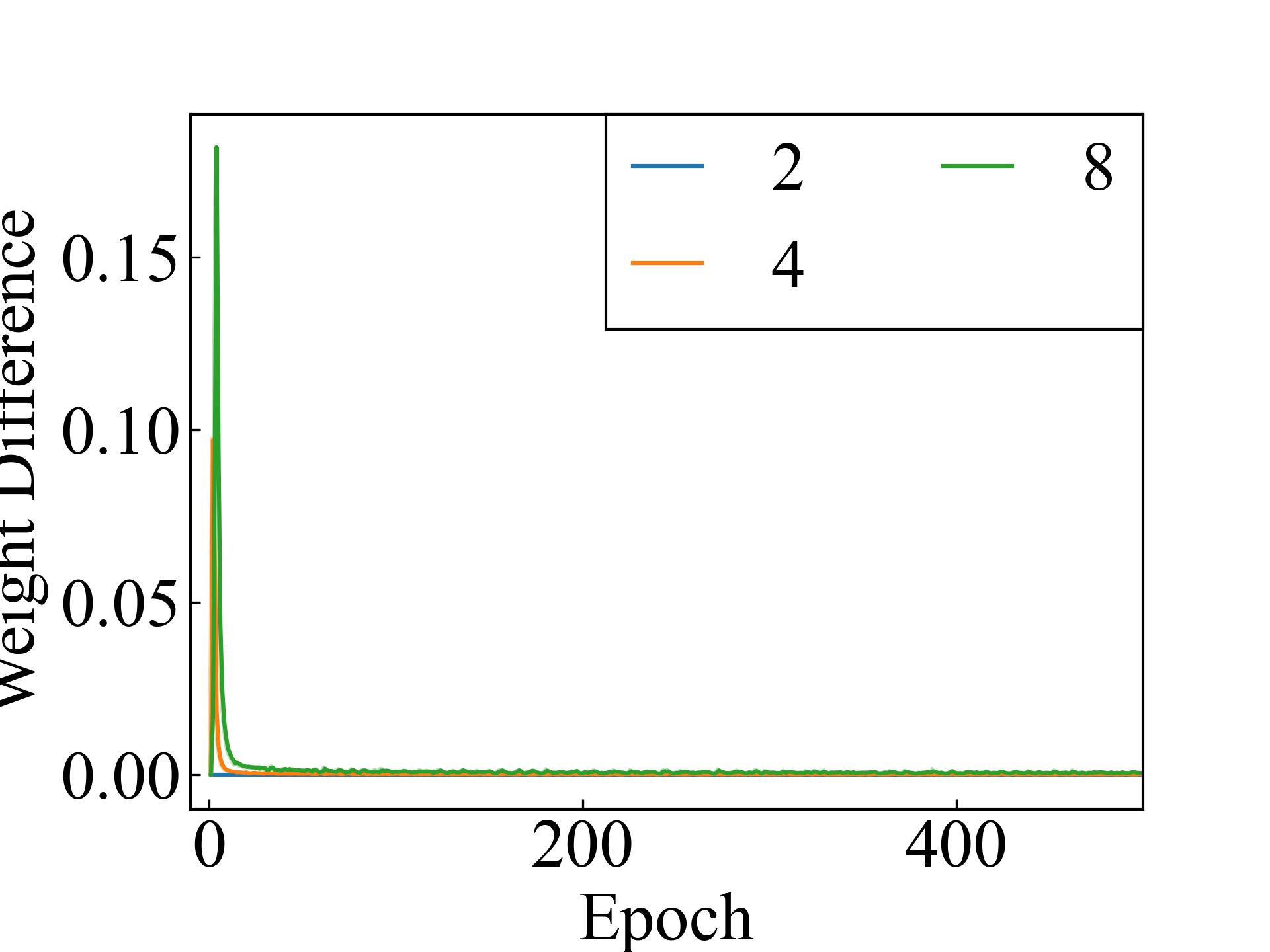}
      \centerline{\quad K-MNIST, MLP}
  \end{minipage}}
  \subfigure{
    \begin{minipage}[b]{0.24\columnwidth}
      \includegraphics[width=\columnwidth]{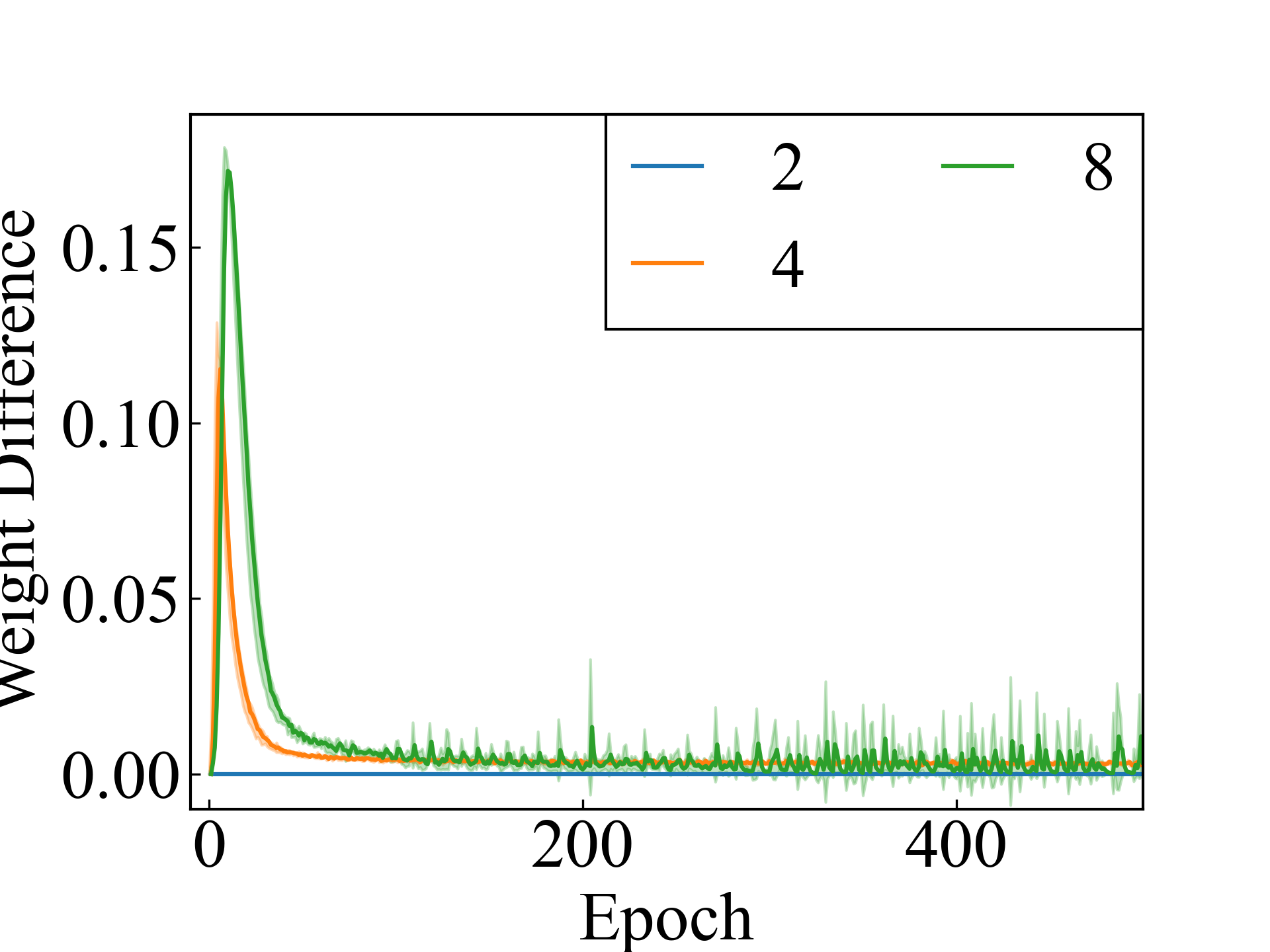}
      \centerline{\quad CIFAR10, ResNet}
  \end{minipage}}
  \subfigure{
    \begin{minipage}[b]{0.24\columnwidth}
      \includegraphics[width=\columnwidth]{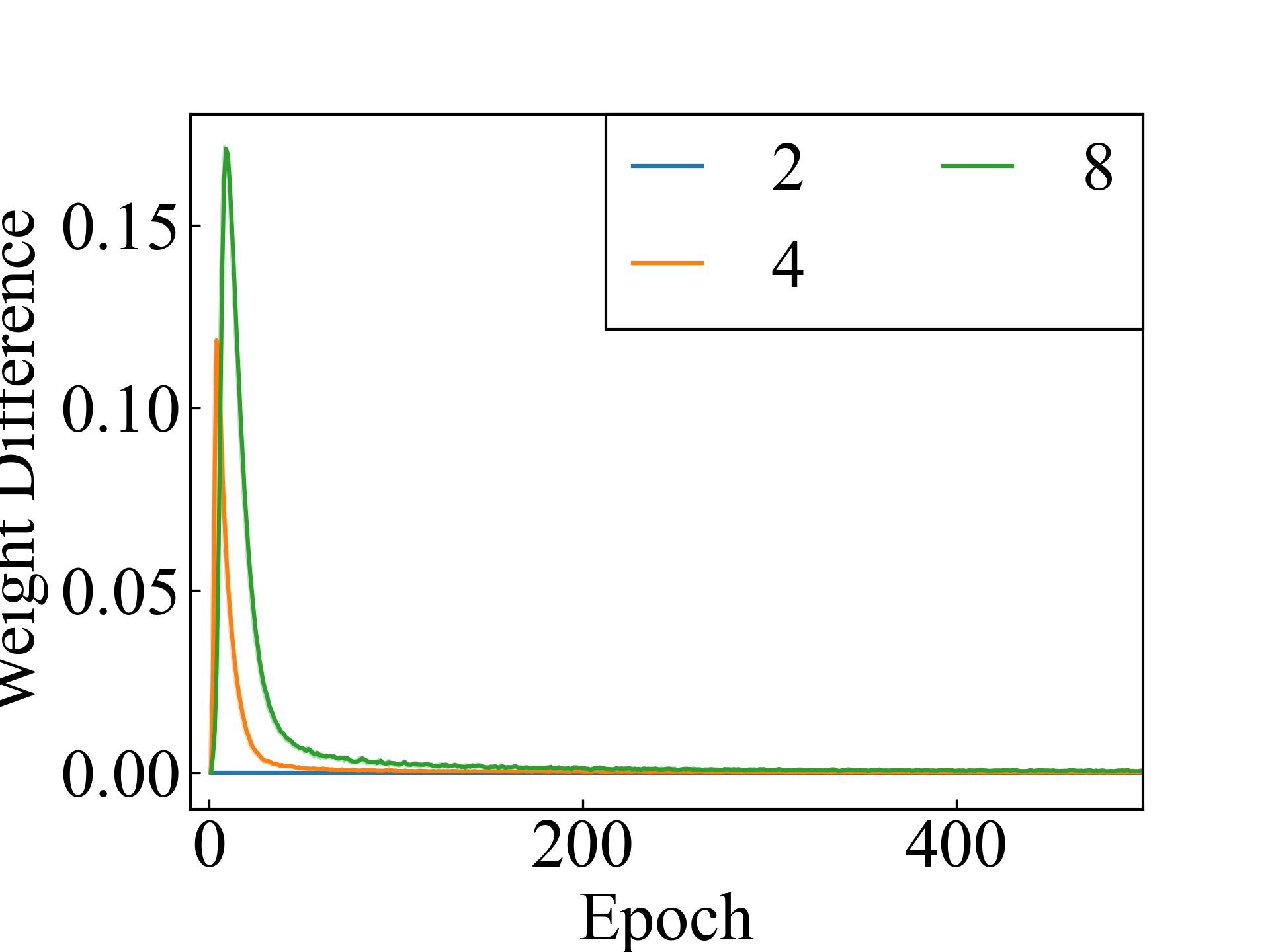}
      \centerline{\quad CIFAR10, ConvNet}
  \end{minipage}}
  \caption{
    Absolute difference between the label weights without and with approximation using RC method.
  }
  \label{fig:approx-weight-diff}
\end{figure*}

%% file: exp_convnet.tex
\begin{table}[h]
\label{tab:convnet}
\vspace{-5pt}
\centering
\begin{tabular}{c}
\specialrule{0.15em}{0pt}{2pt}
Input 3$\times$32$\times$32 image \\
\specialrule{0.05em}{2pt}{2pt}
3$\times$3 conv. 128 followed by LeakyReLU ~$\times$ 3 \\
\specialrule{0.05em}{2pt}{2pt}
max-pooling, dropout with $p=0.25$ \\
\specialrule{0.05em}{2pt}{2pt}
3$\times$3 conv. 256 followed by LeakyReLU ~$\times$ 3 \\
\specialrule{0.05em}{2pt}{2pt}
max-pooling, dropout with $p=0.25$ \\
\specialrule{0.05em}{2pt}{2pt}
3$\times$3 conv. 512 followed by LeakyReLU  ~$\times$ 3 \\
\specialrule{0.05em}{2pt}{2pt}
global mean pooling, Dense 10 \\
\specialrule{0.05em}{2pt}{2pt}
\end{tabular}
\caption{ConvNet model for CIFAR-10.}
\vspace{-5pt}
\end{table}

%% file: exp_fig_ours_tst.tex
\begin{figure*}[!t]
  \subfigure{
    \begin{minipage}[b]{0.24\columnwidth}
      \centering MNIST
      \includegraphics[width=\columnwidth]{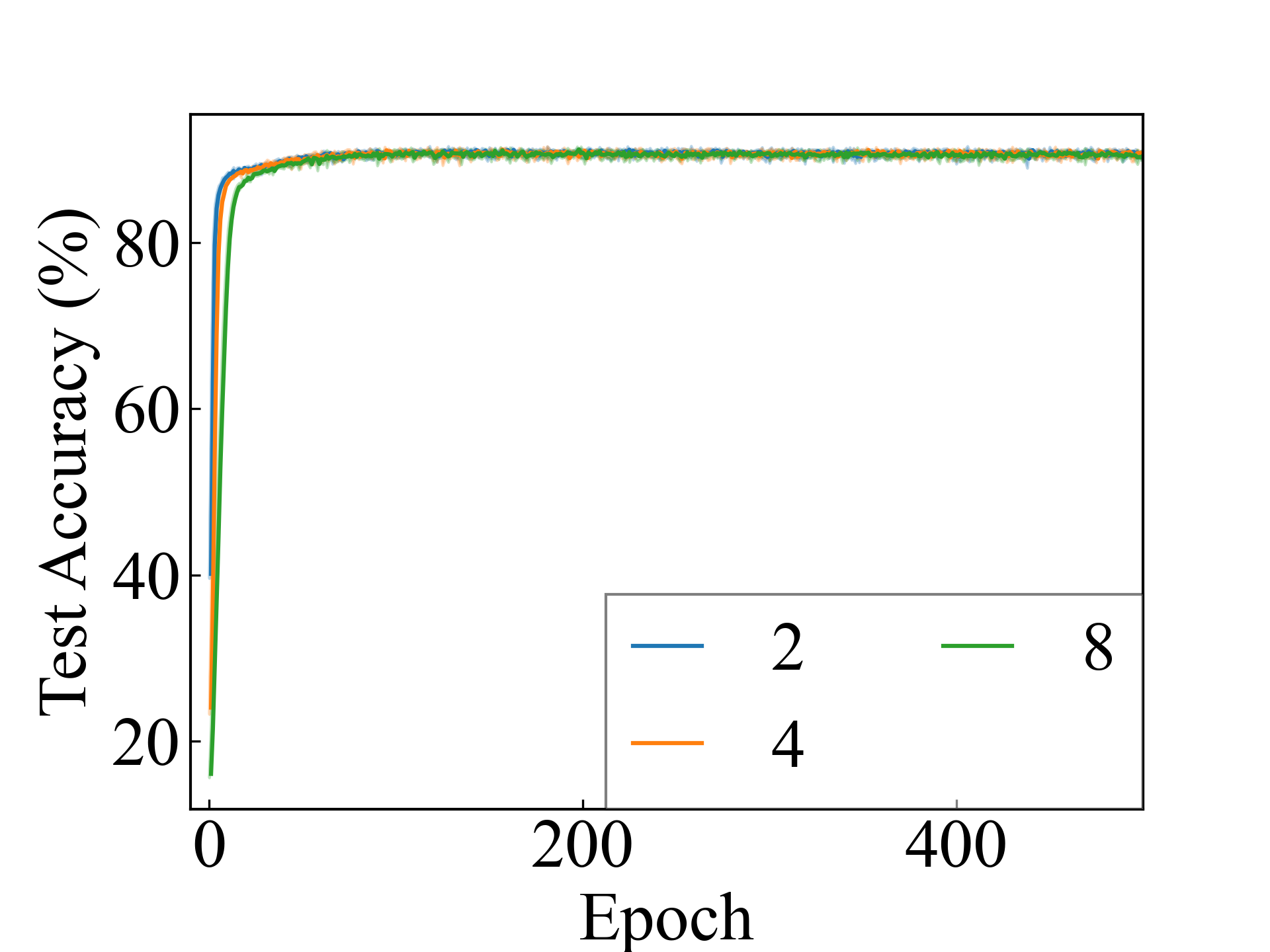}
      \centerline{\quad Linear, RC}
  \end{minipage}}
  \subfigure{
    \begin{minipage}[b]{0.24\columnwidth}
      \includegraphics[width=\columnwidth]{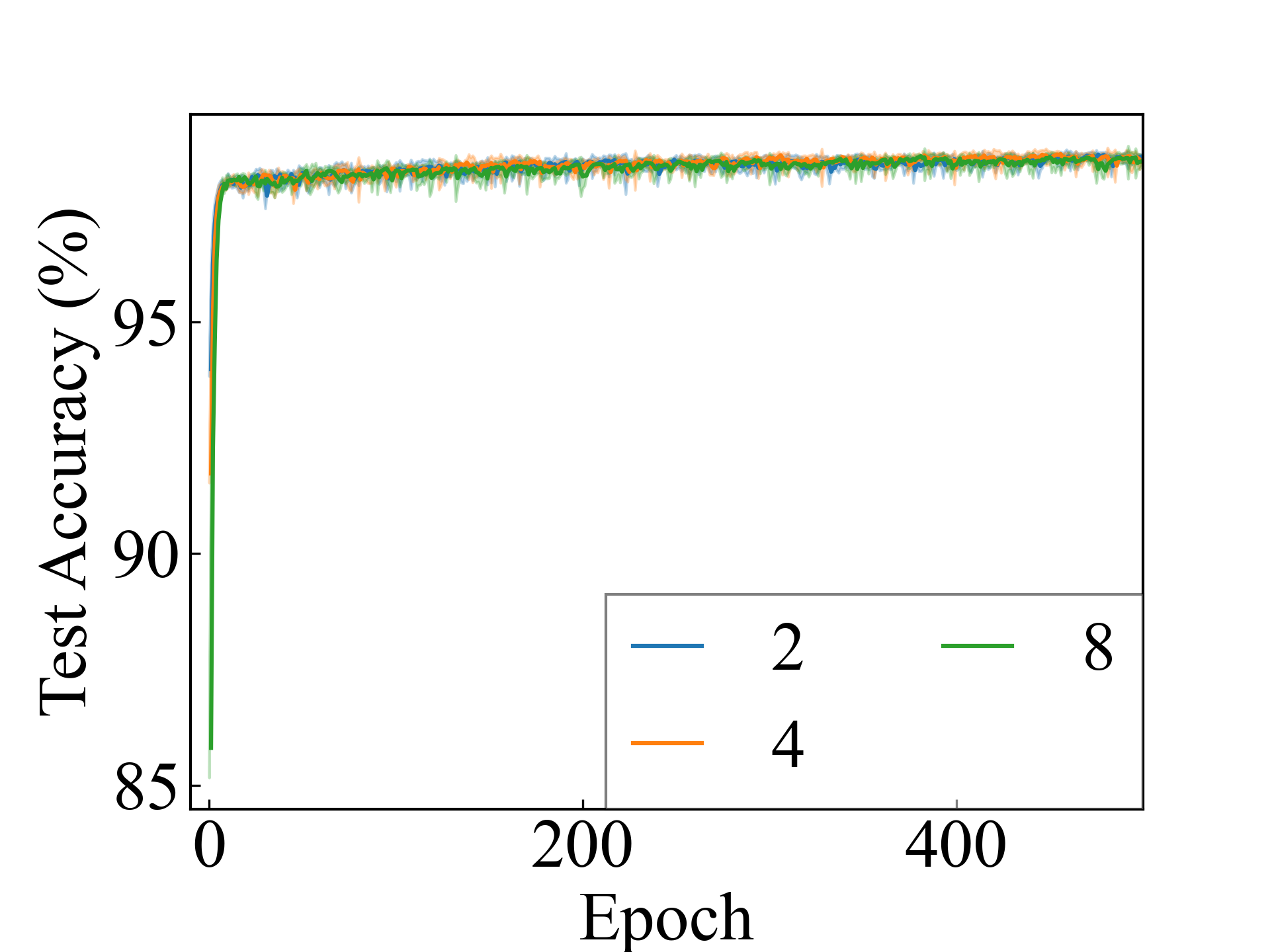}
      \centerline{\quad MLP, RC}
  \end{minipage}}
  \subfigure{
    \begin{minipage}[b]{0.24\columnwidth}
      \includegraphics[width=\columnwidth]{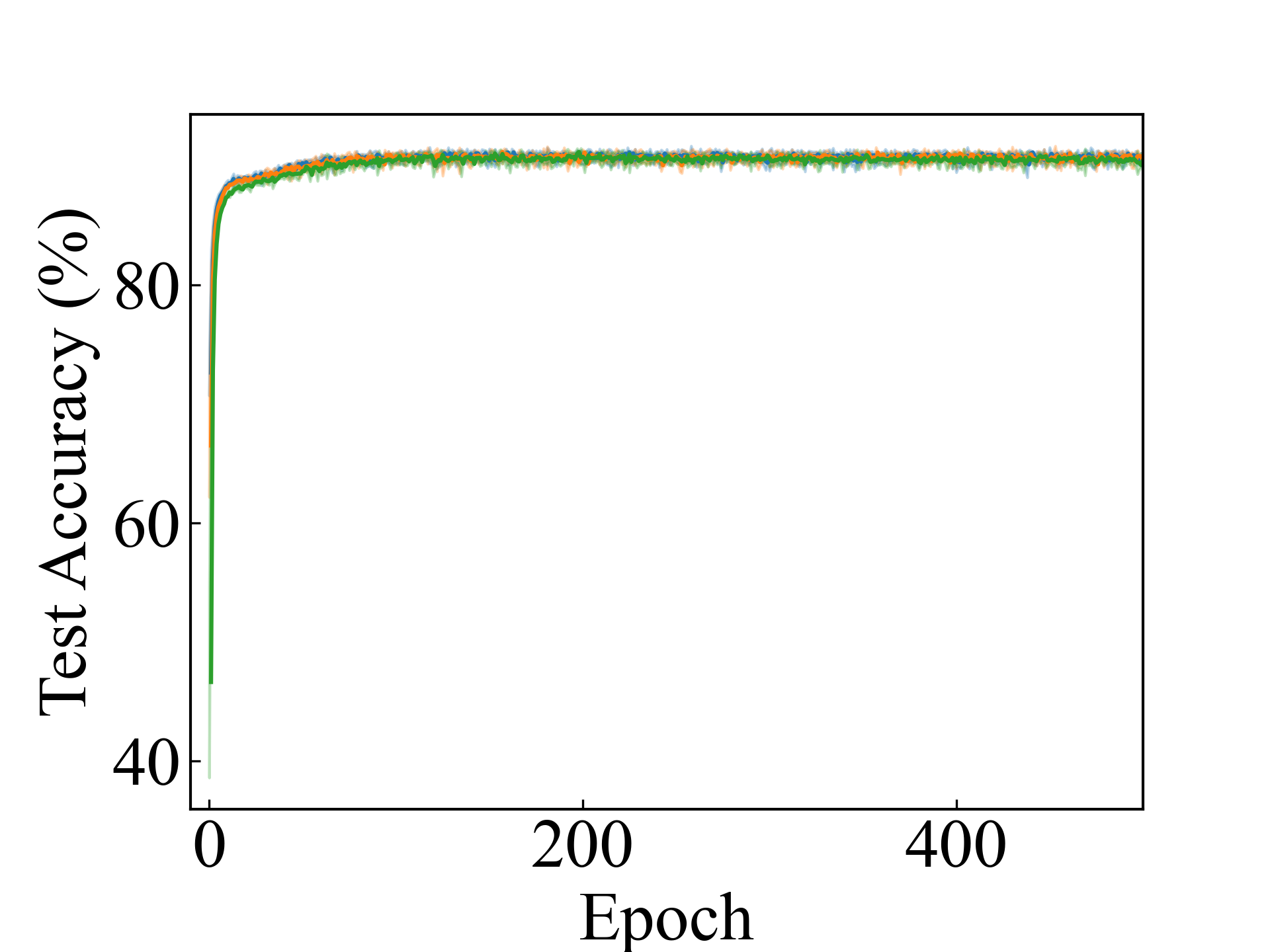}
      \centerline{\quad Linear, CC}
  \end{minipage}}
  \subfigure{
    \begin{minipage}[b]{0.24\columnwidth}
      \includegraphics[width=\columnwidth]{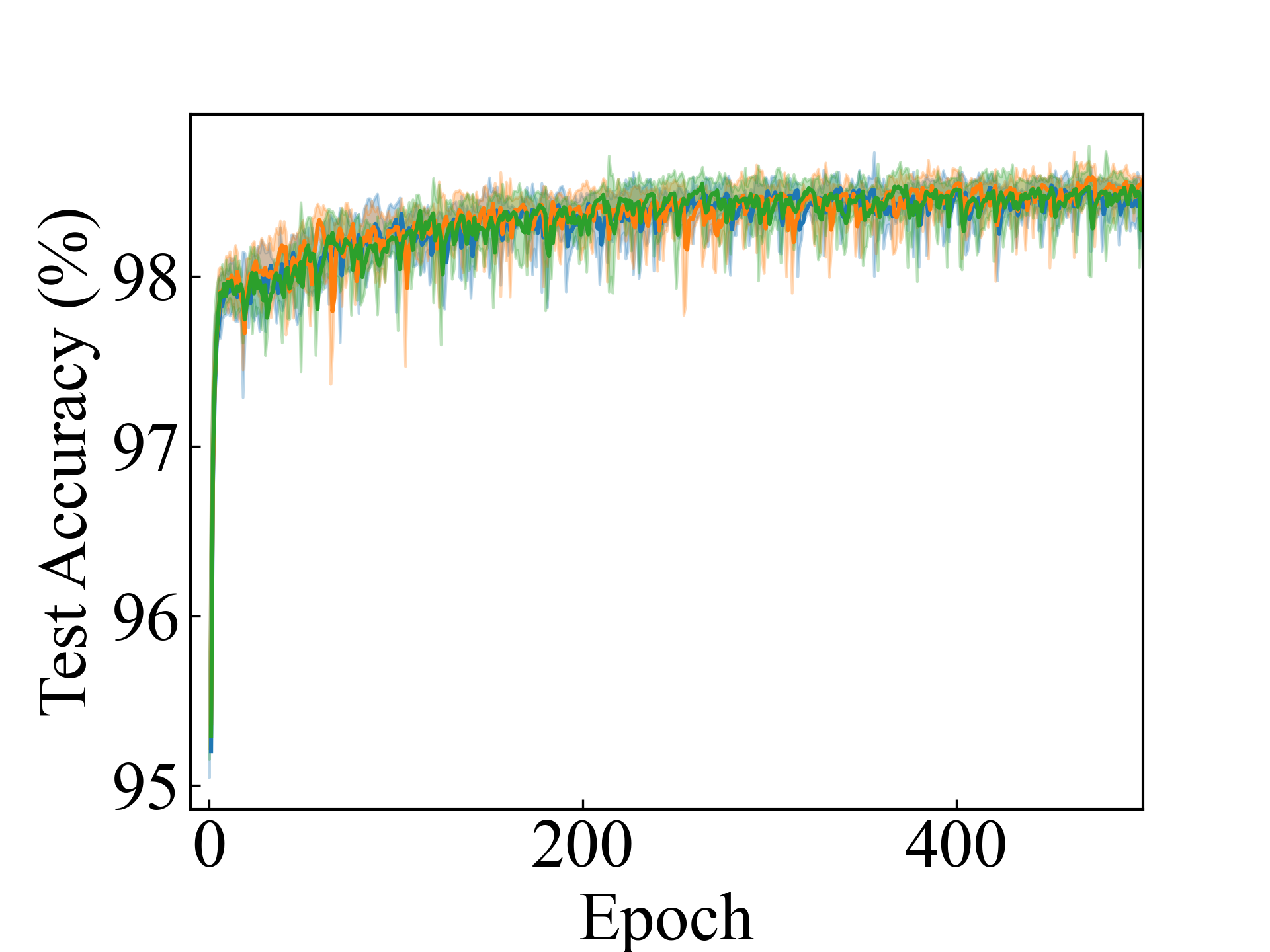}
      \centerline{\quad MLP, CC}
  \end{minipage}}
  \subfigure{
    \begin{minipage}[b]{0.24\columnwidth}
      \centering F-MNIST
      \includegraphics[width=\columnwidth]{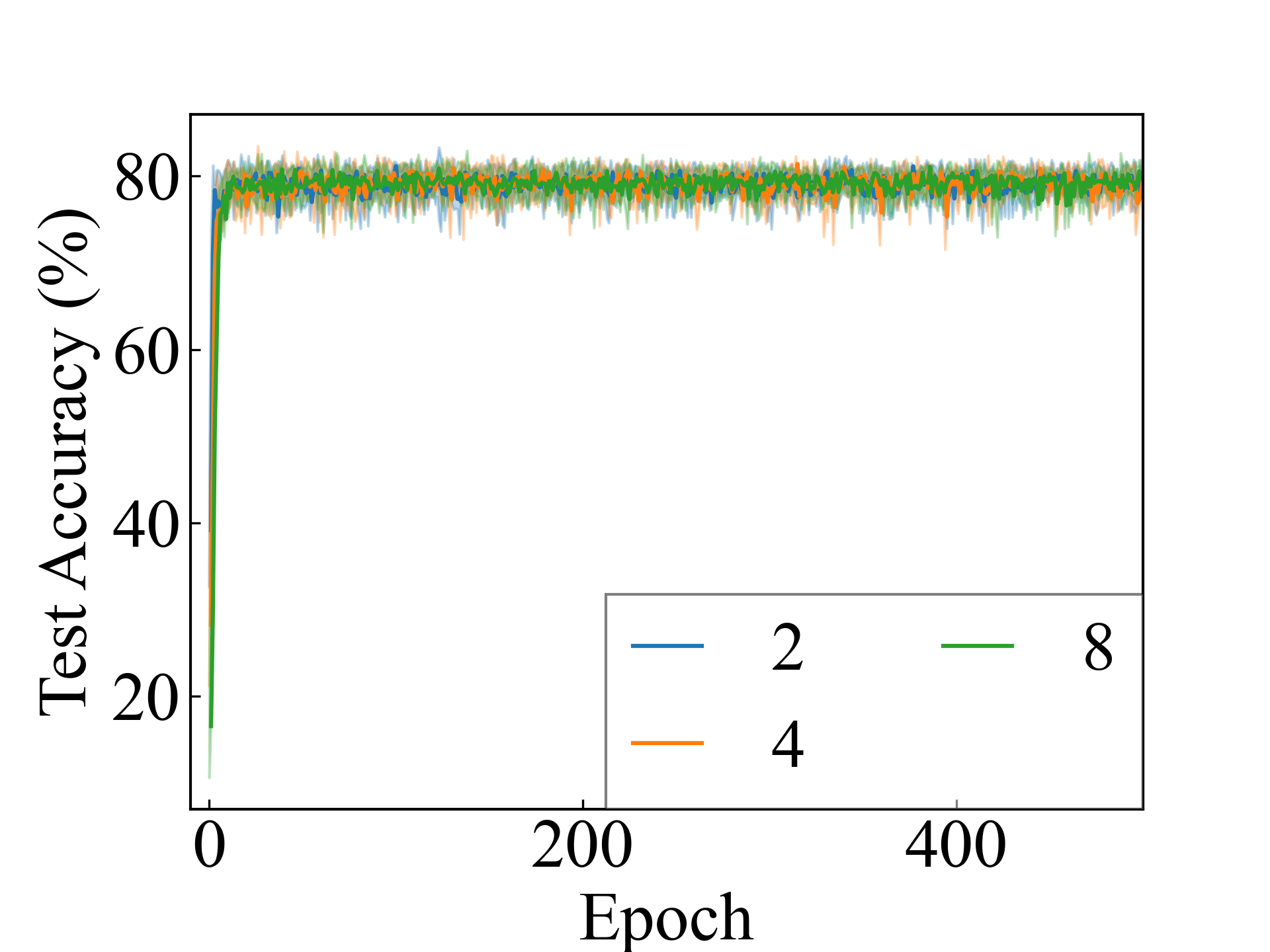}
      \centerline{\quad Linear, RC}
  \end{minipage}}
  \subfigure{
    \begin{minipage}[b]{0.24\columnwidth}
      \includegraphics[width=\columnwidth]{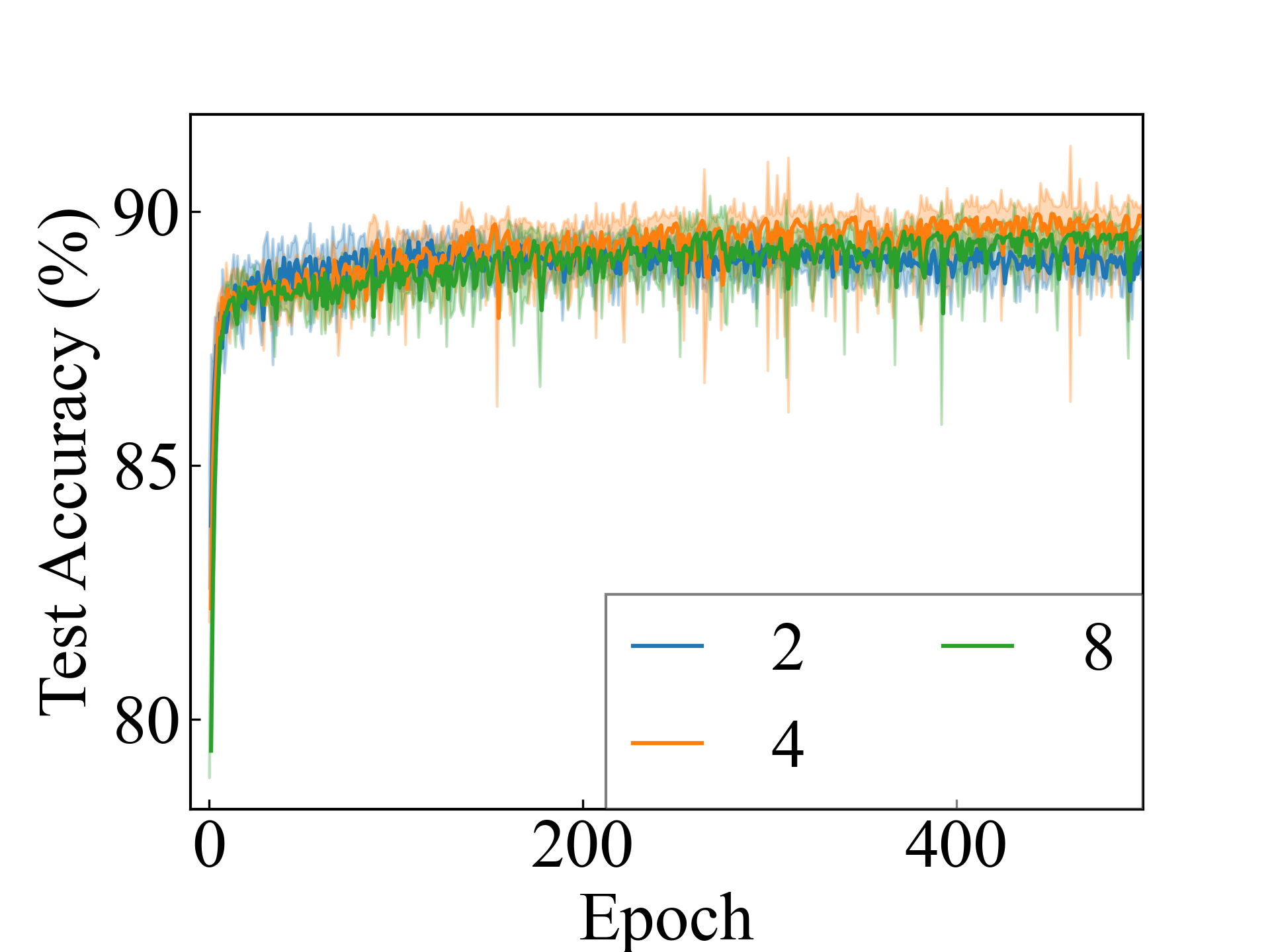}
      \centerline{\quad MLP, RC}
  \end{minipage}}
  \subfigure{
    \begin{minipage}[b]{0.24\columnwidth}
      \includegraphics[width=\columnwidth]{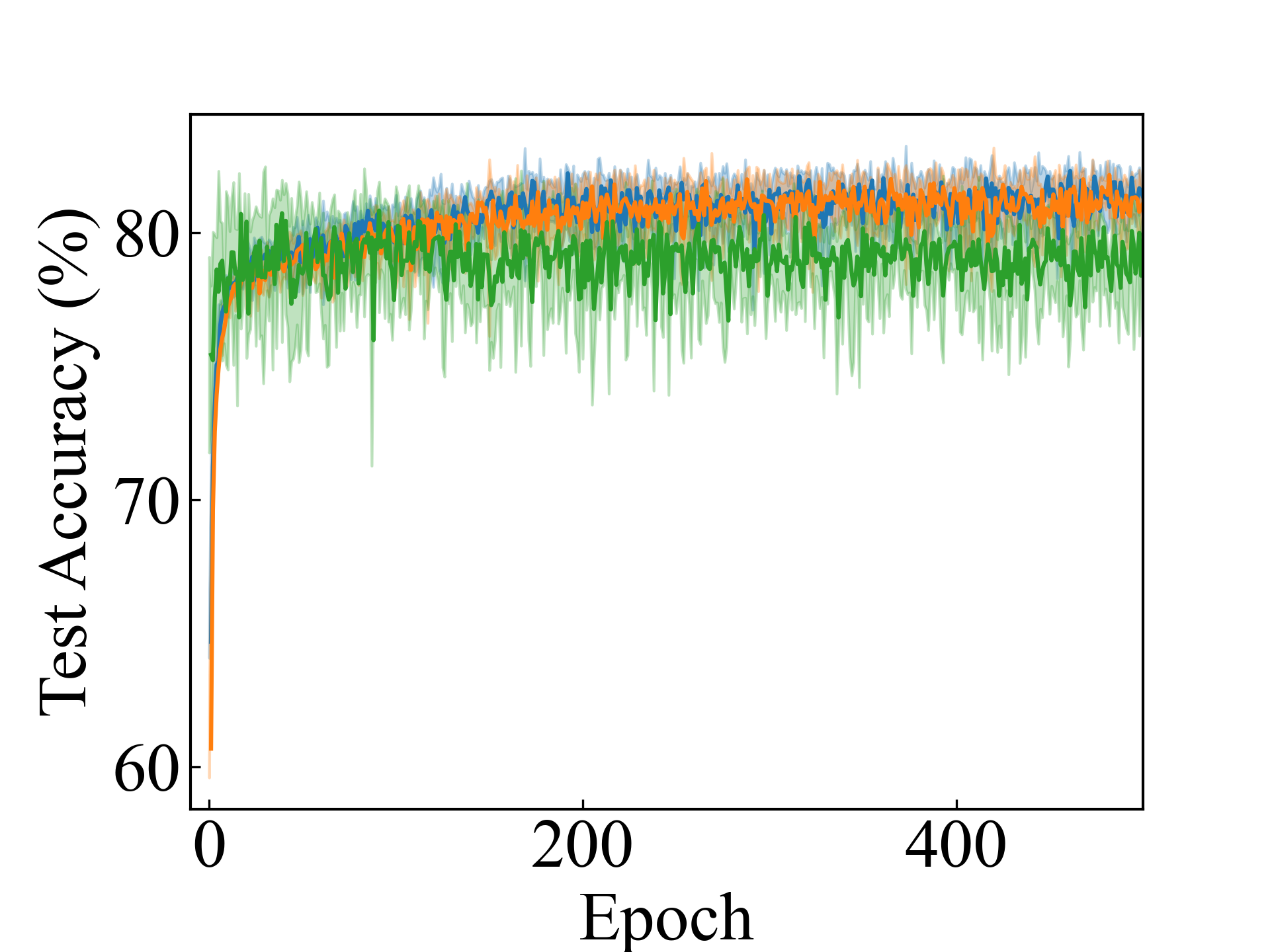}
      \centerline{\quad Linear, CC}
  \end{minipage}}
  \subfigure{
    \begin{minipage}[b]{0.24\columnwidth}
      \includegraphics[width=\columnwidth]{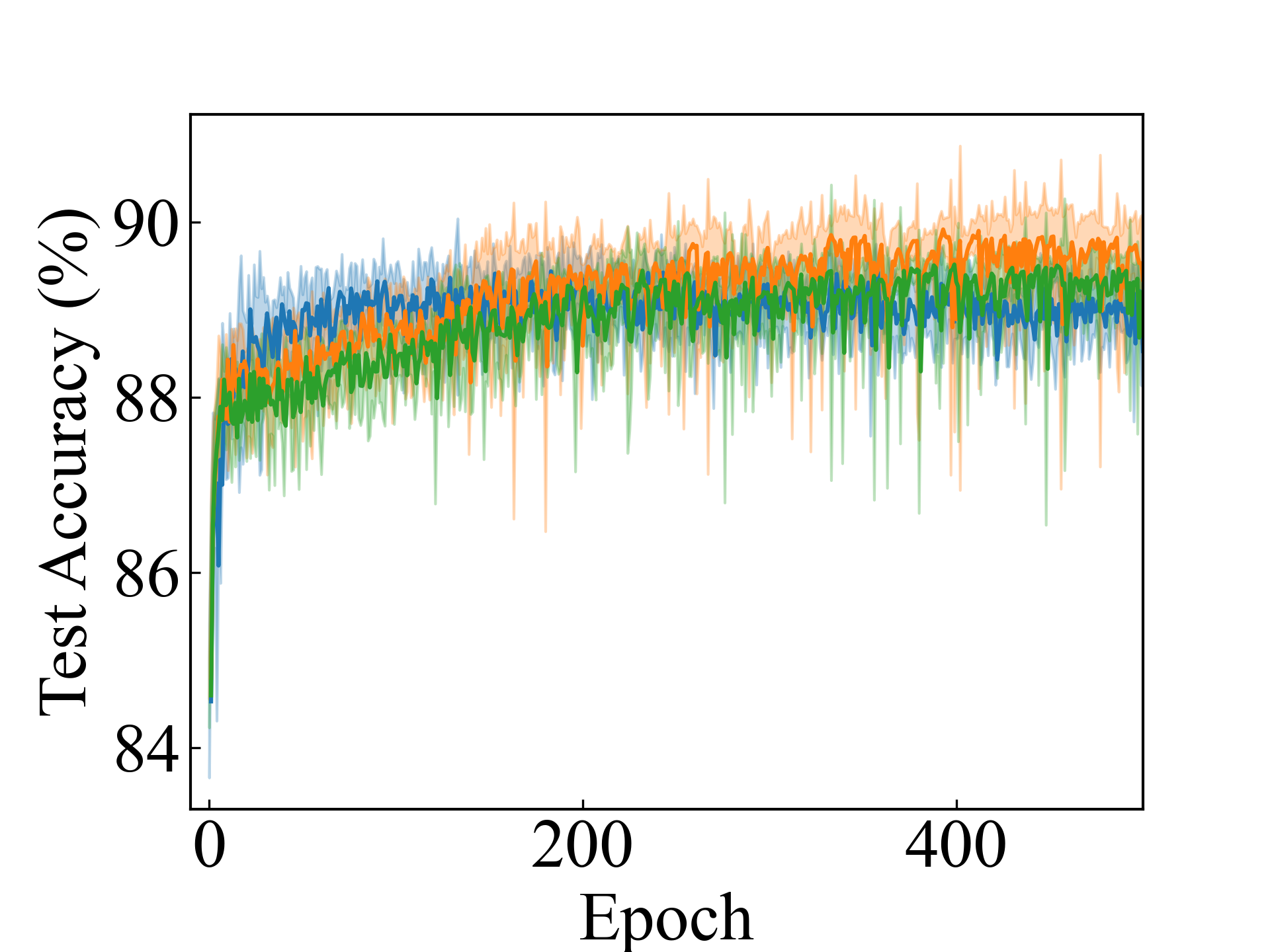}
      \centerline{\quad MLP, CC}
  \end{minipage}}
  \subfigure{
    \begin{minipage}[b]{0.24\columnwidth}
      \centering K-MNIST
      \includegraphics[width=\columnwidth]{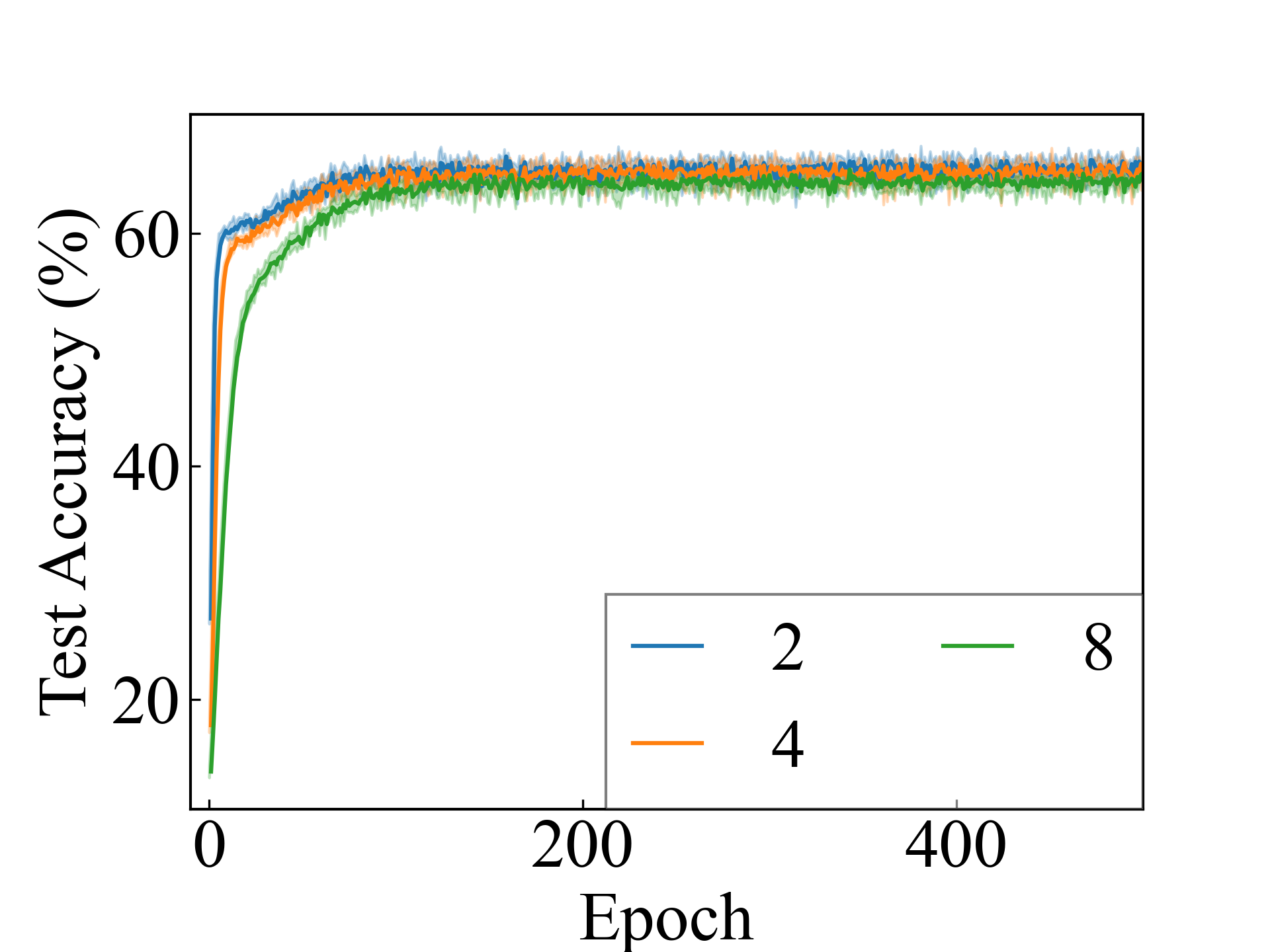}
      \centerline{\quad Linear, RC}
  \end{minipage}}
  \subfigure{
    \begin{minipage}[b]{0.24\columnwidth}
      \includegraphics[width=\columnwidth]{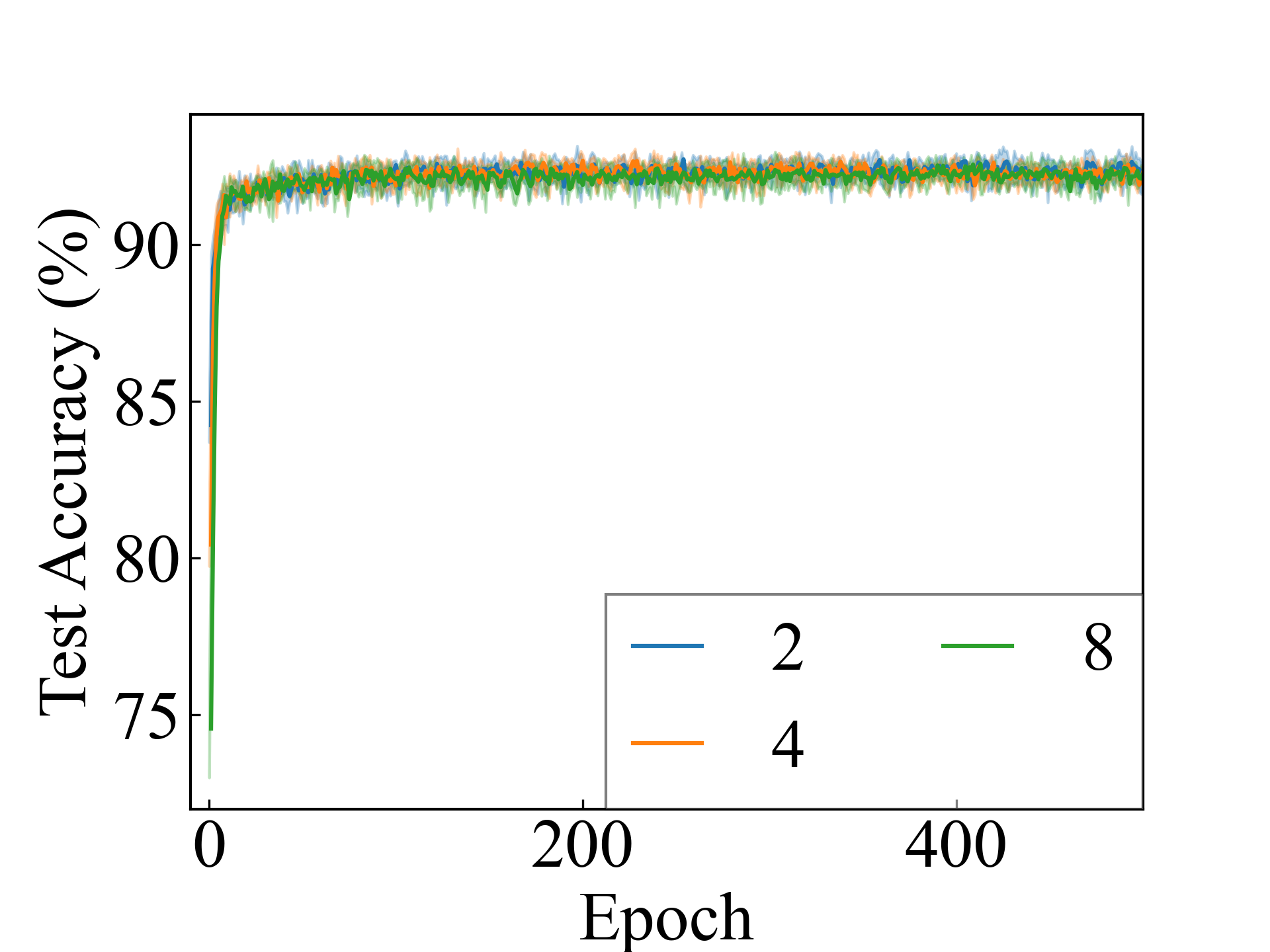}
      \centerline{\quad MLP, RC}
  \end{minipage}}
  \subfigure{
    \begin{minipage}[b]{0.24\columnwidth}
      \includegraphics[width=\columnwidth]{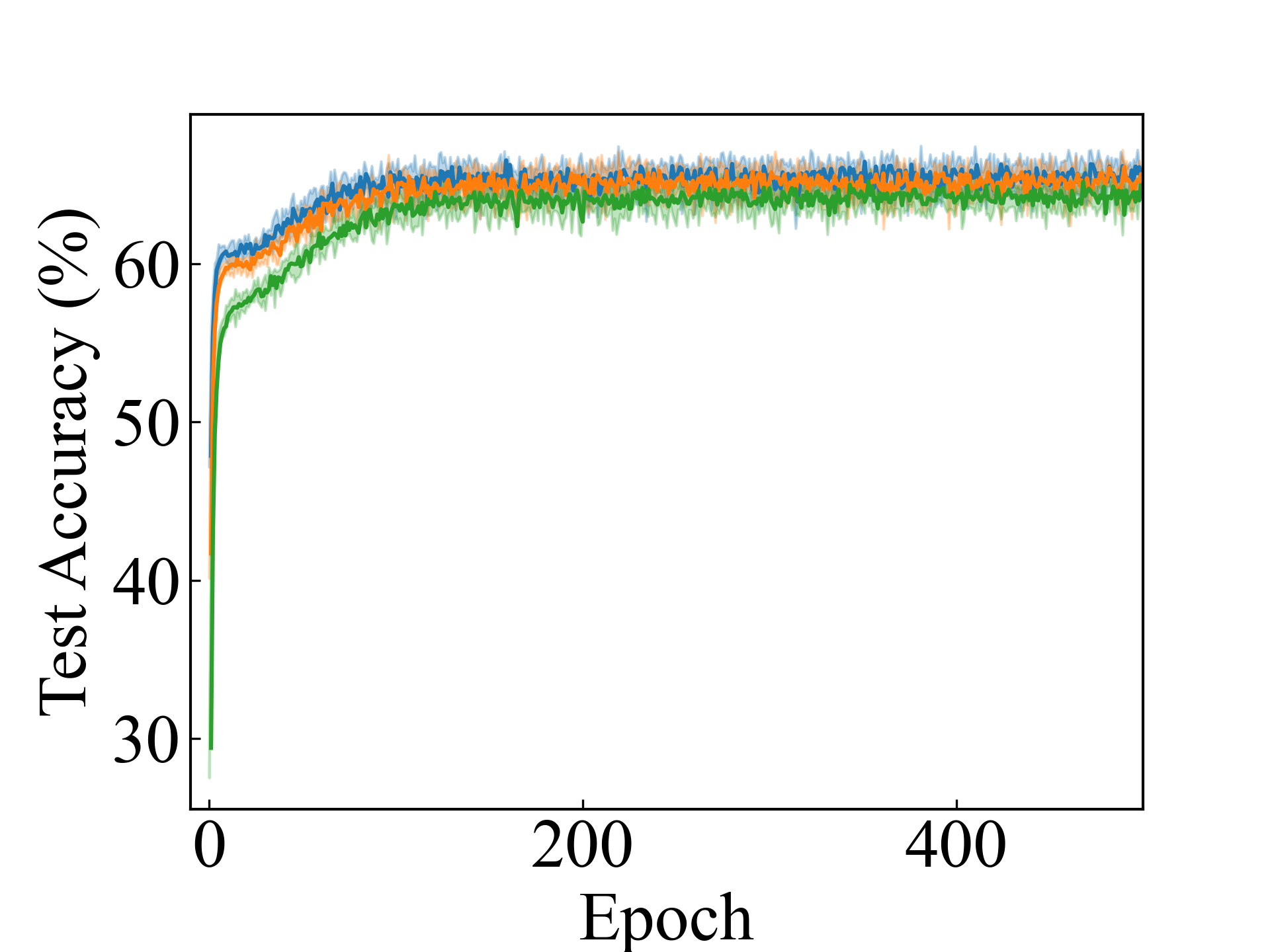}
      \centerline{\quad Linear, CC}
  \end{minipage}}
  \subfigure{
    \begin{minipage}[b]{0.24\columnwidth}
      \includegraphics[width=\columnwidth]{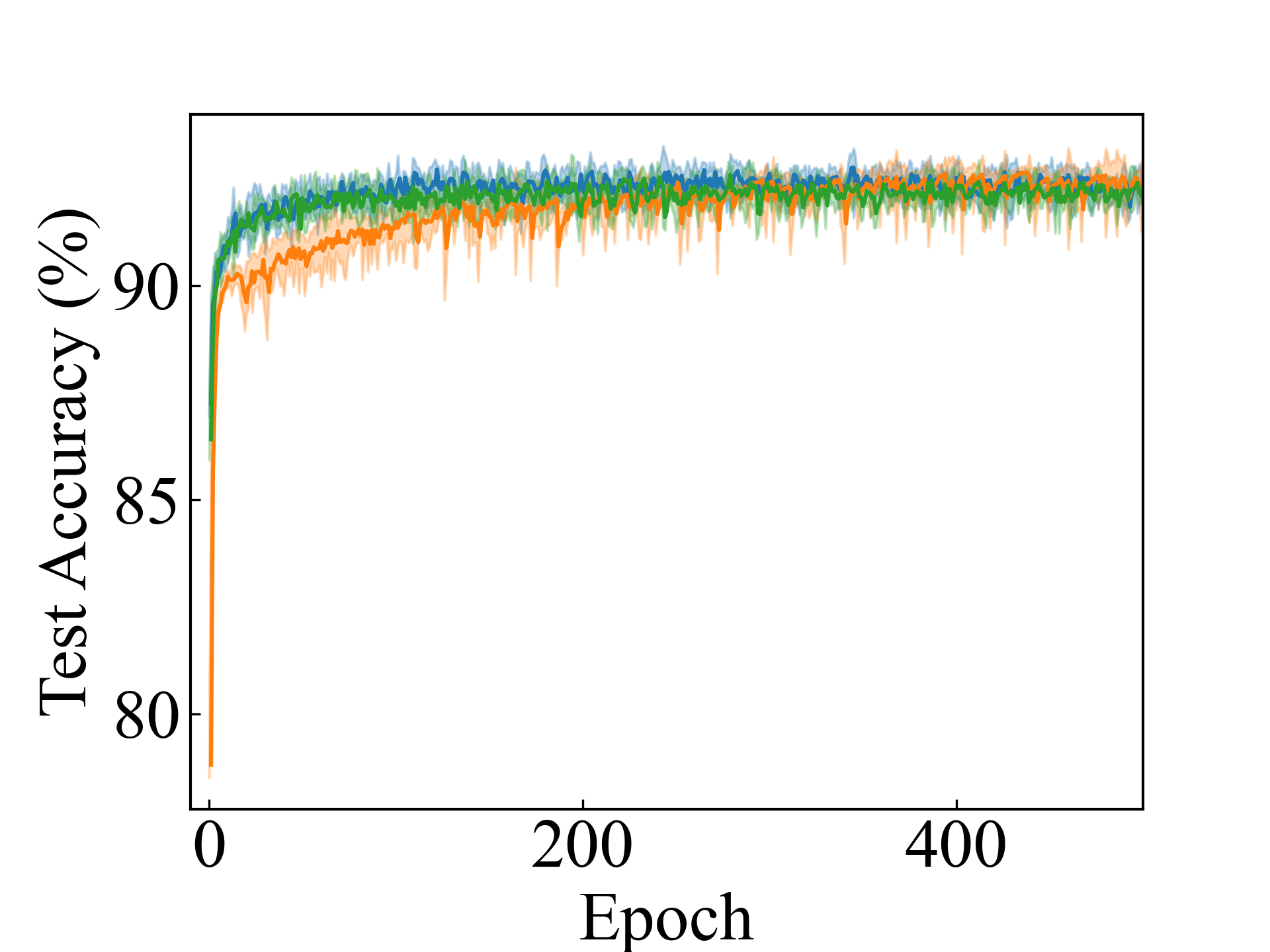}
      \centerline{\quad MLP, CC}
  \end{minipage}}
  \subfigure{
    \begin{minipage}[b]{0.24\columnwidth}
      \centering CIFAR-10
      \includegraphics[width=\columnwidth]{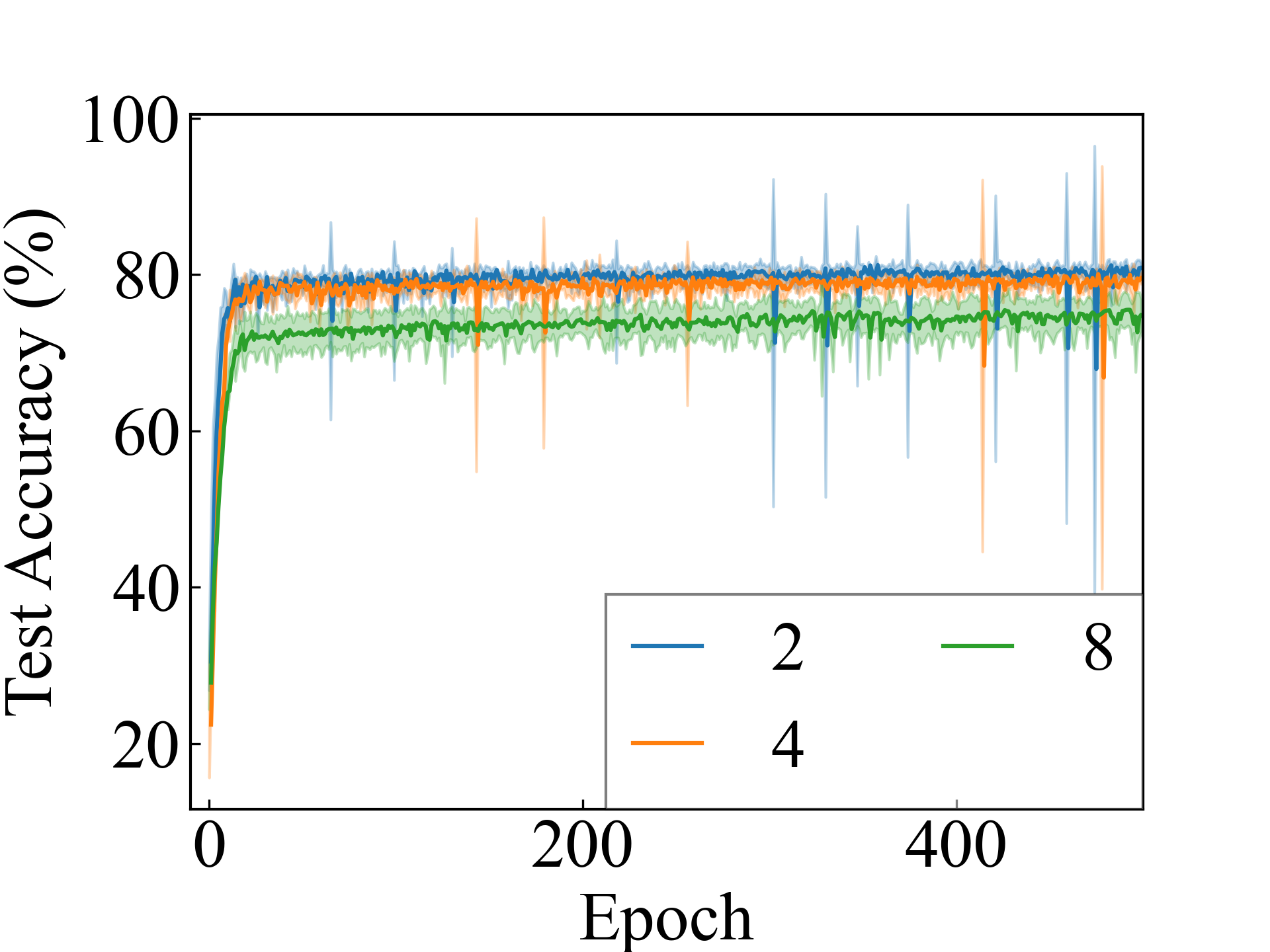}
      \centerline{\quad ResNet, RC}
  \end{minipage}}
  \subfigure{
    \begin{minipage}[b]{0.24\columnwidth}
      \includegraphics[width=\columnwidth]{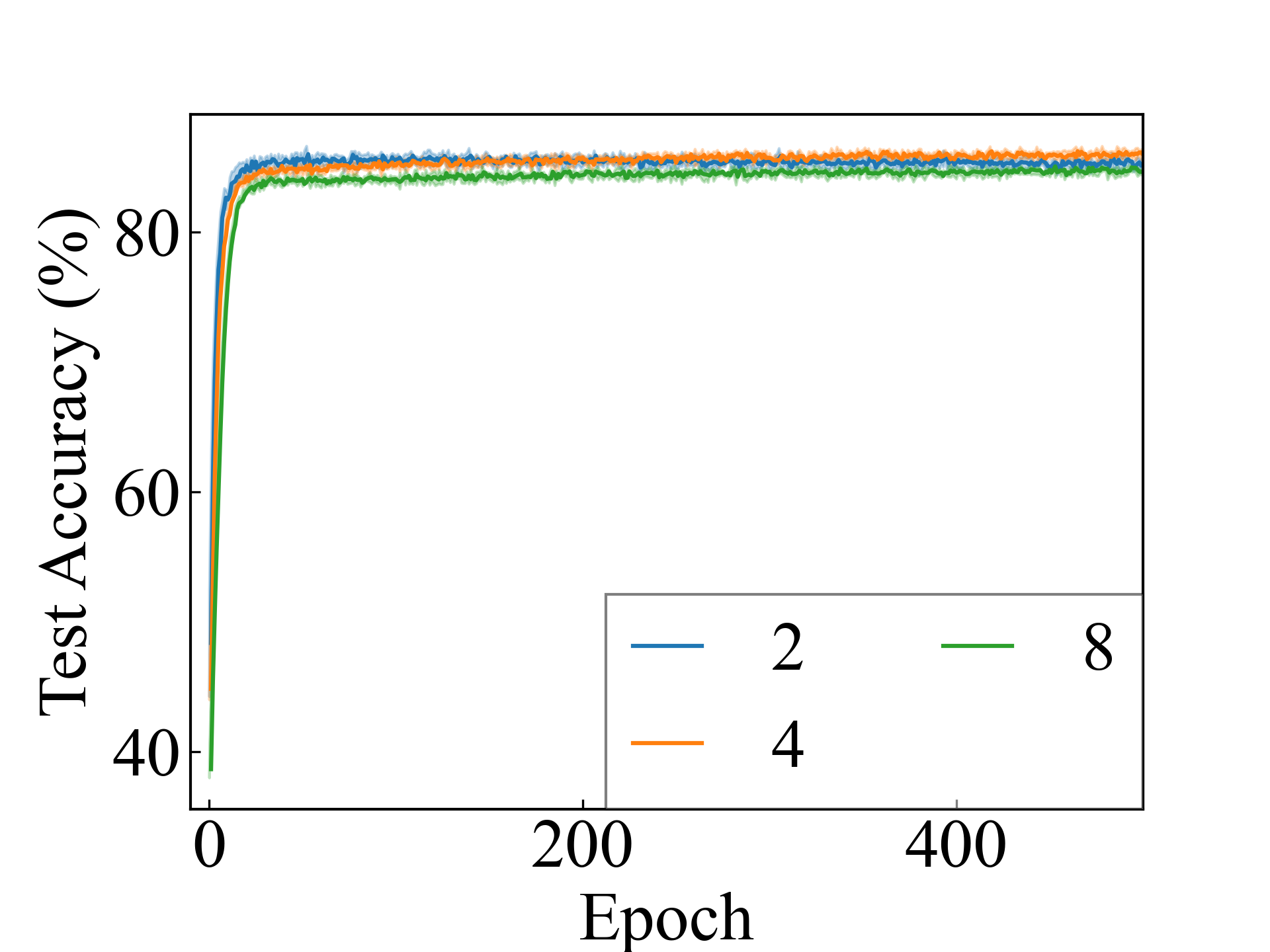}
      \centerline{\quad ConvNet, RC}
  \end{minipage}}
  \subfigure{
    \begin{minipage}[b]{0.24\columnwidth}
      \includegraphics[width=\columnwidth]{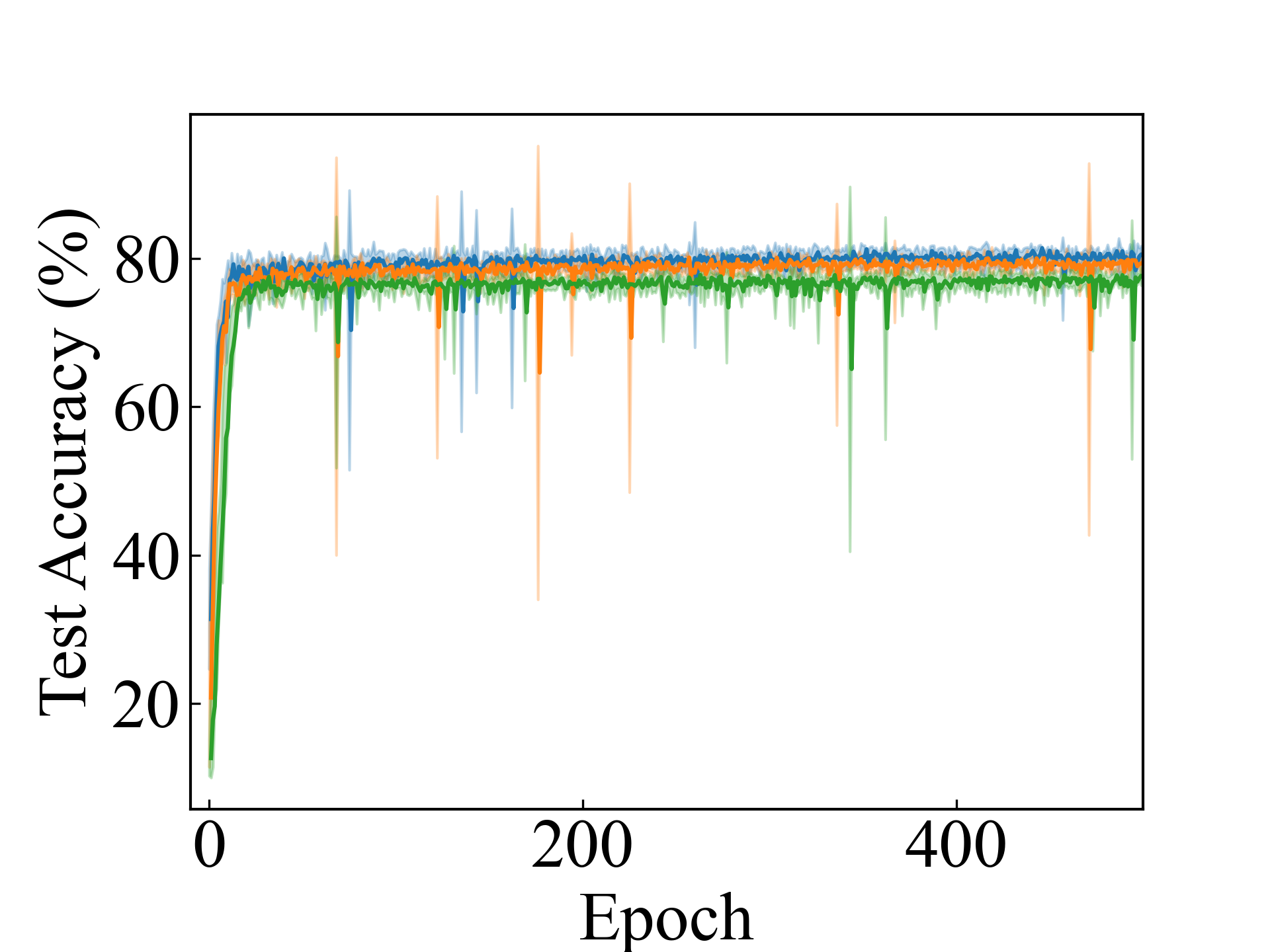}
      \centerline{\quad ResNet, CC}
  \end{minipage}}
  \subfigure{
    \begin{minipage}[b]{0.24\columnwidth}
      \includegraphics[width=\columnwidth]{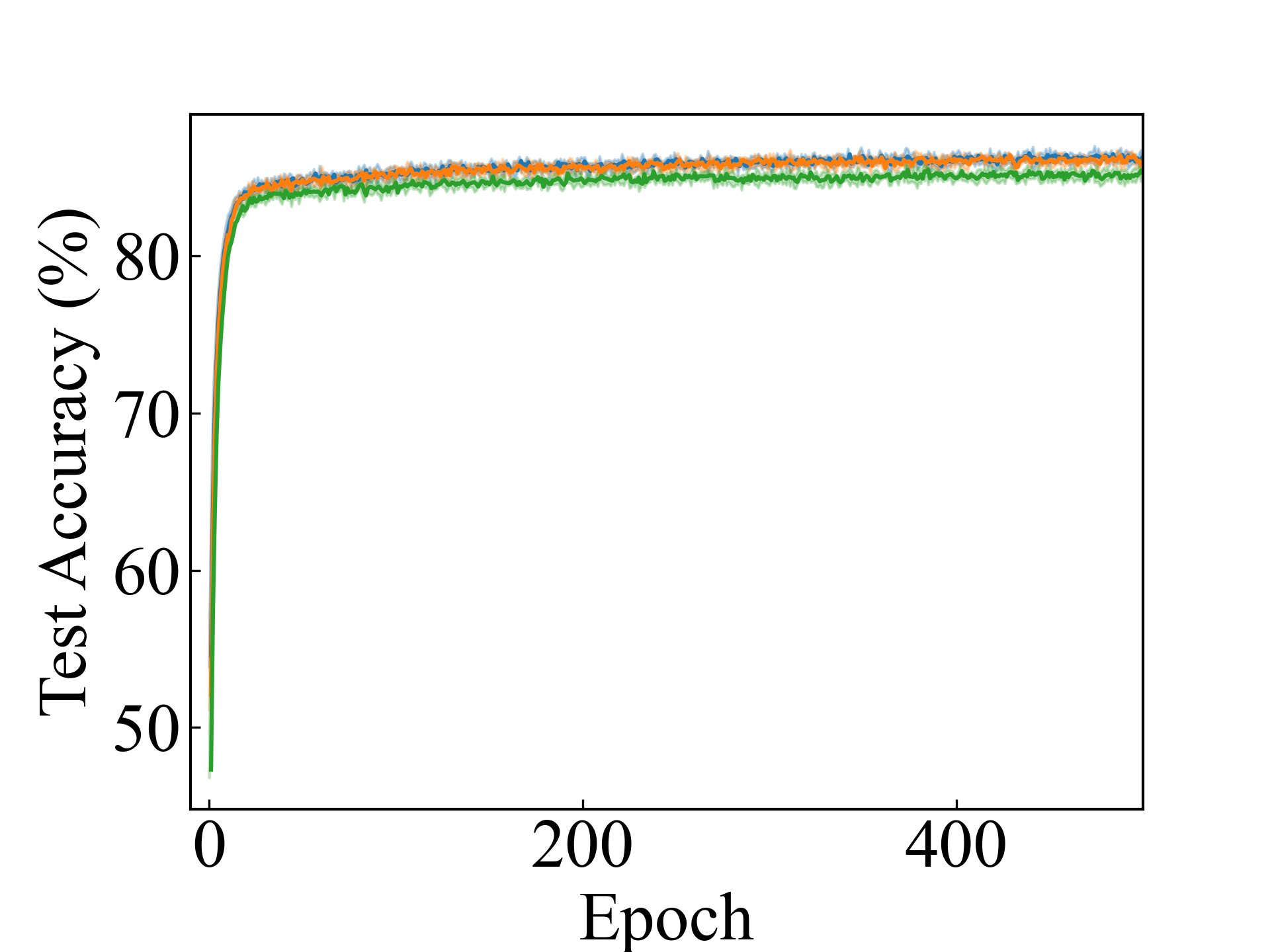}
      \centerline{\quad ConvNet, CC}
  \end{minipage}}
  \caption{
  Test accuracy for various settings with proposed methods (RC, CC).
  }
  \label{fig:test-accuracy-curve1}
\end{figure*}

\begin{figure*}[!t]
  \subfigure{
    \begin{minipage}[b]{0.24\columnwidth}
      \centering MNIST
      \includegraphics[width=\columnwidth]{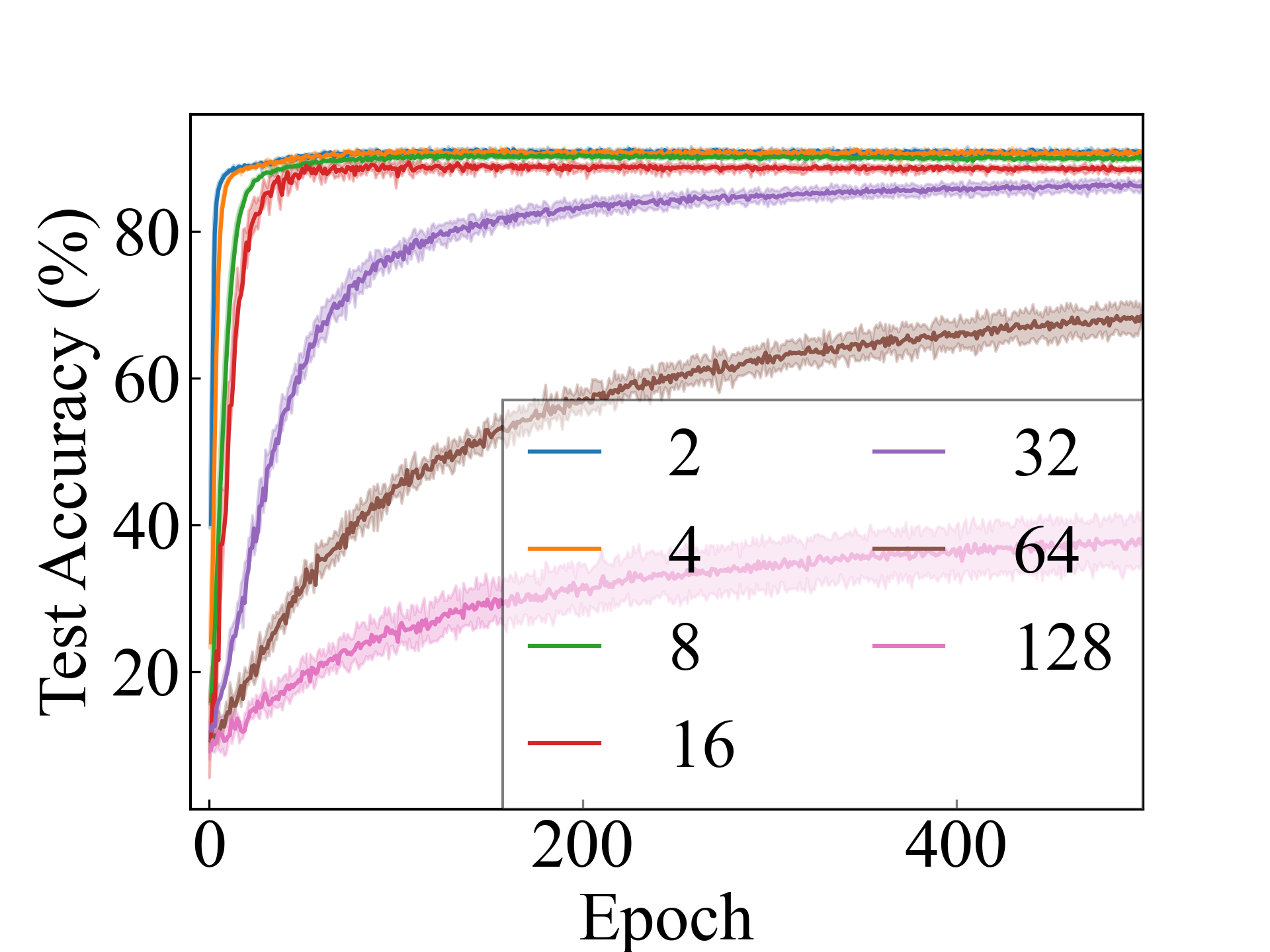}
      \centerline{\quad Linear, RC\_Approx}
  \end{minipage}}
  \subfigure{
    \begin{minipage}[b]{0.24\columnwidth}
      \includegraphics[width=\columnwidth]{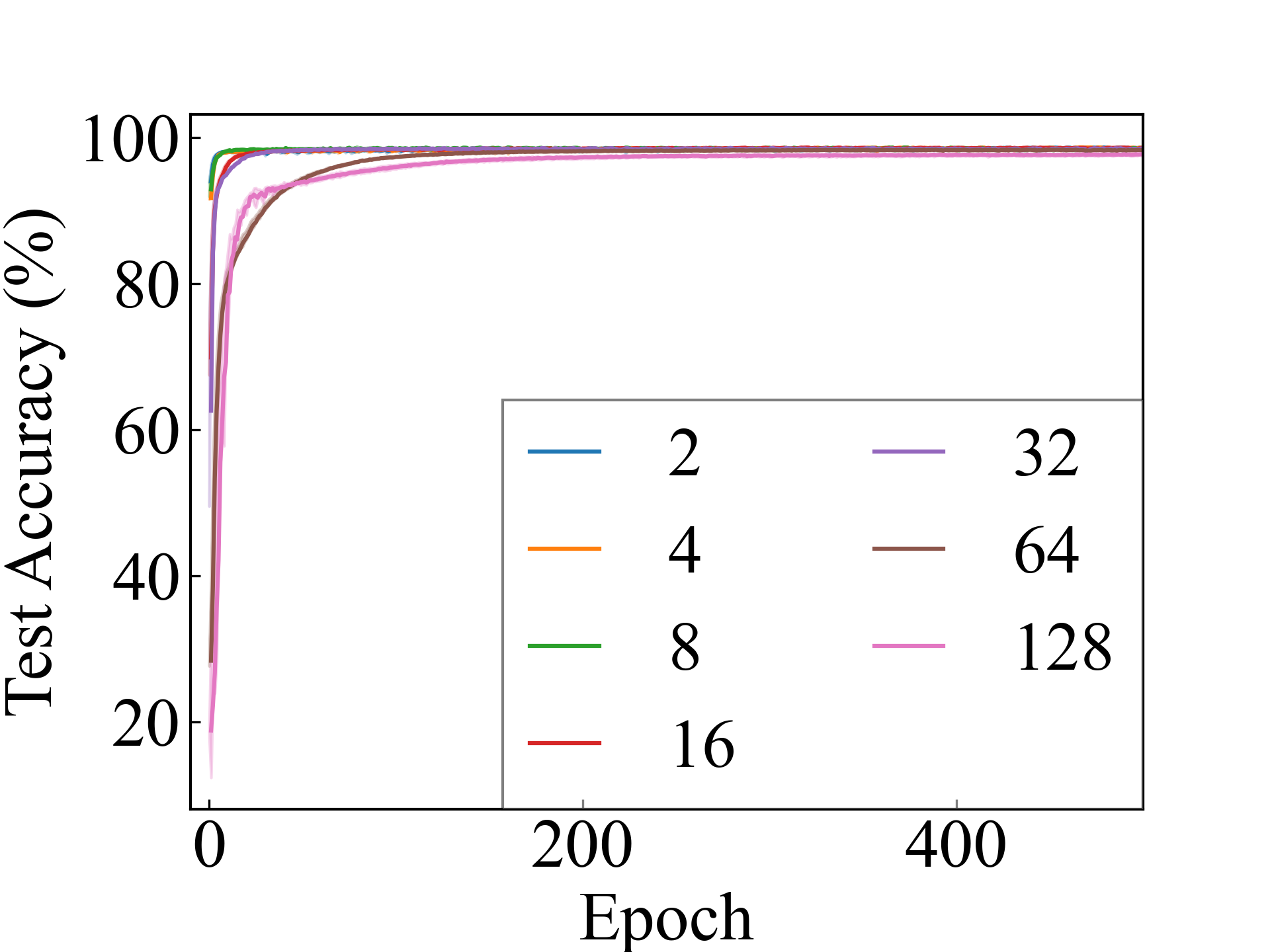}
      \centerline{\quad MLP, RC\_Approx}
  \end{minipage}}
  \subfigure{
    \begin{minipage}[b]{0.24\columnwidth}
      \includegraphics[width=\columnwidth]{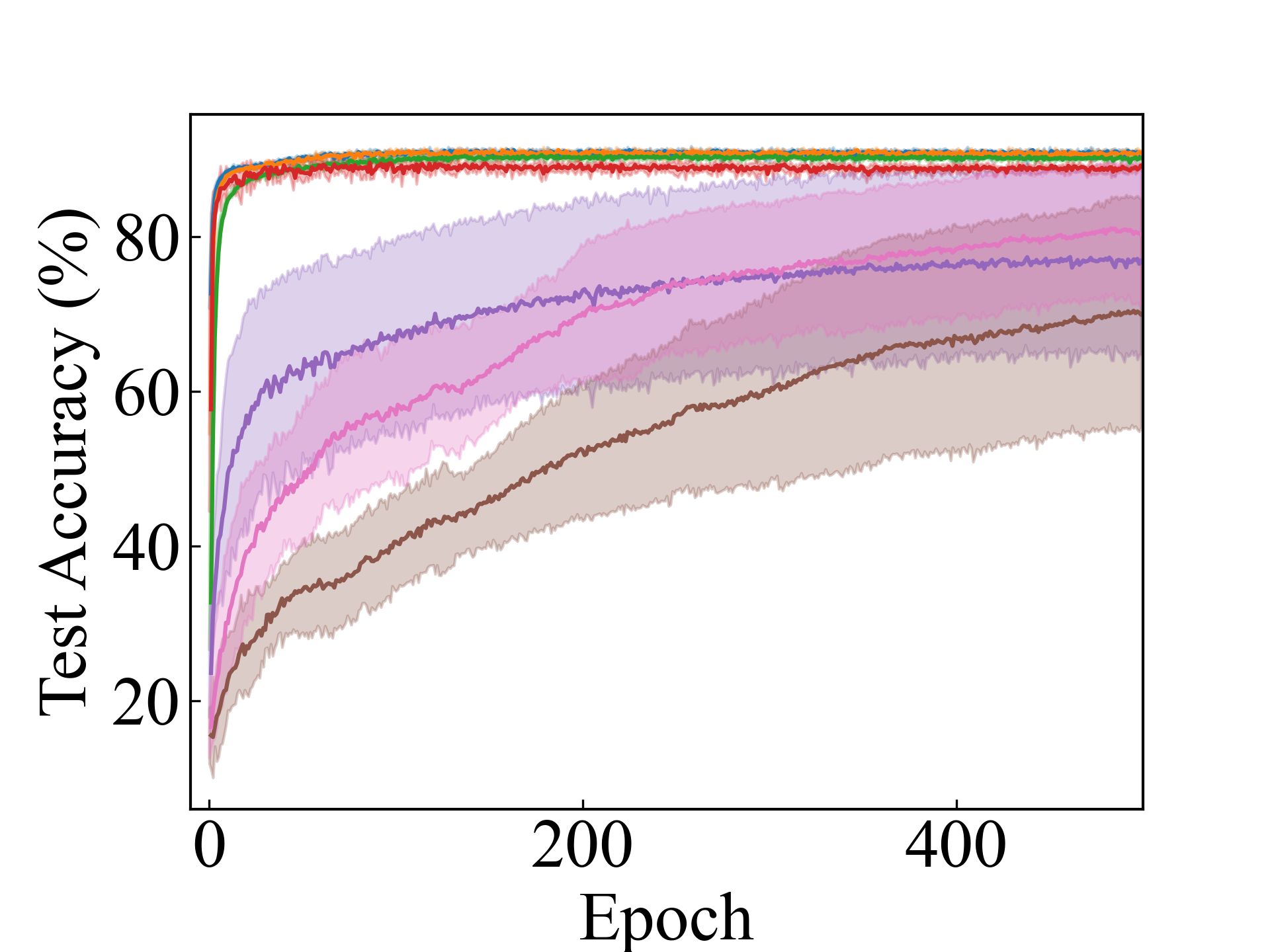}
      \centerline{\quad Linear, CC\_Approx}
  \end{minipage}}
  \subfigure{
    \begin{minipage}[b]{0.24\columnwidth}
      \includegraphics[width=\columnwidth]{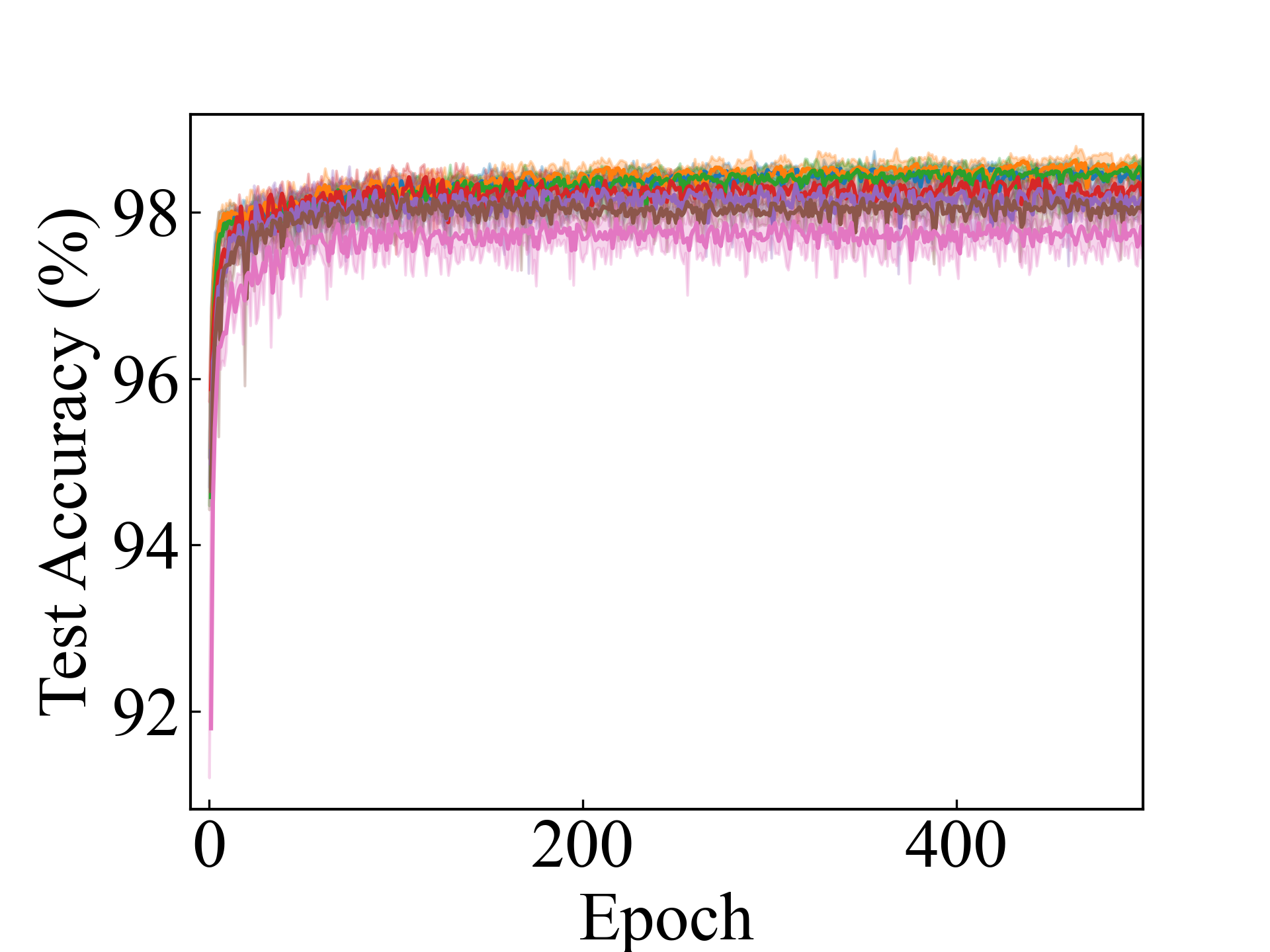}
      \centerline{\quad MLP, CC\_Approx}
  \end{minipage}}
  \subfigure{
    \begin{minipage}[b]{0.24\columnwidth}
      \centering F-MNIST
      \includegraphics[width=\columnwidth]{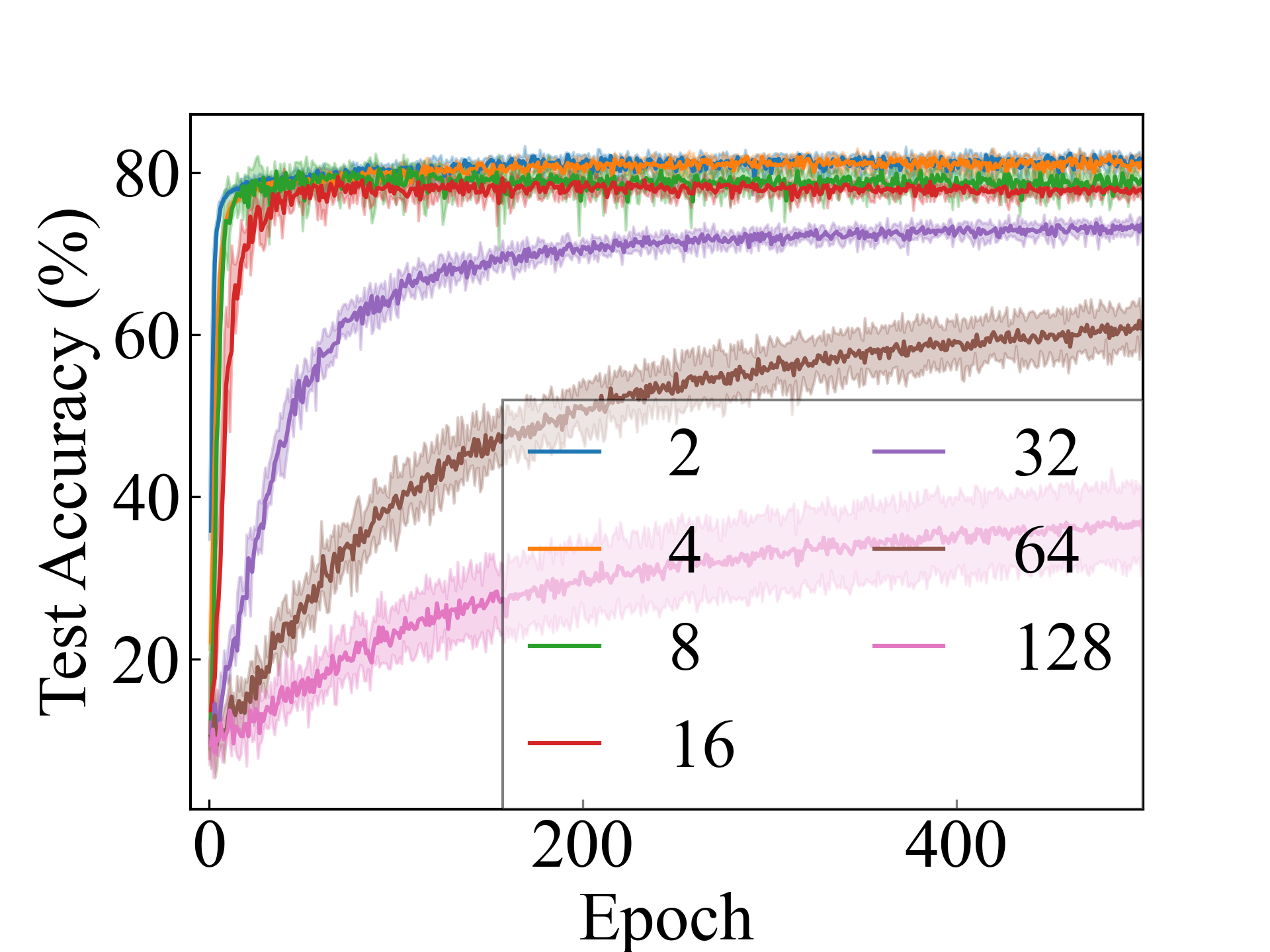}
      \centerline{\quad Linear, RC\_Approx}
  \end{minipage}}
  \subfigure{
    \begin{minipage}[b]{0.24\columnwidth}
      \includegraphics[width=\columnwidth]{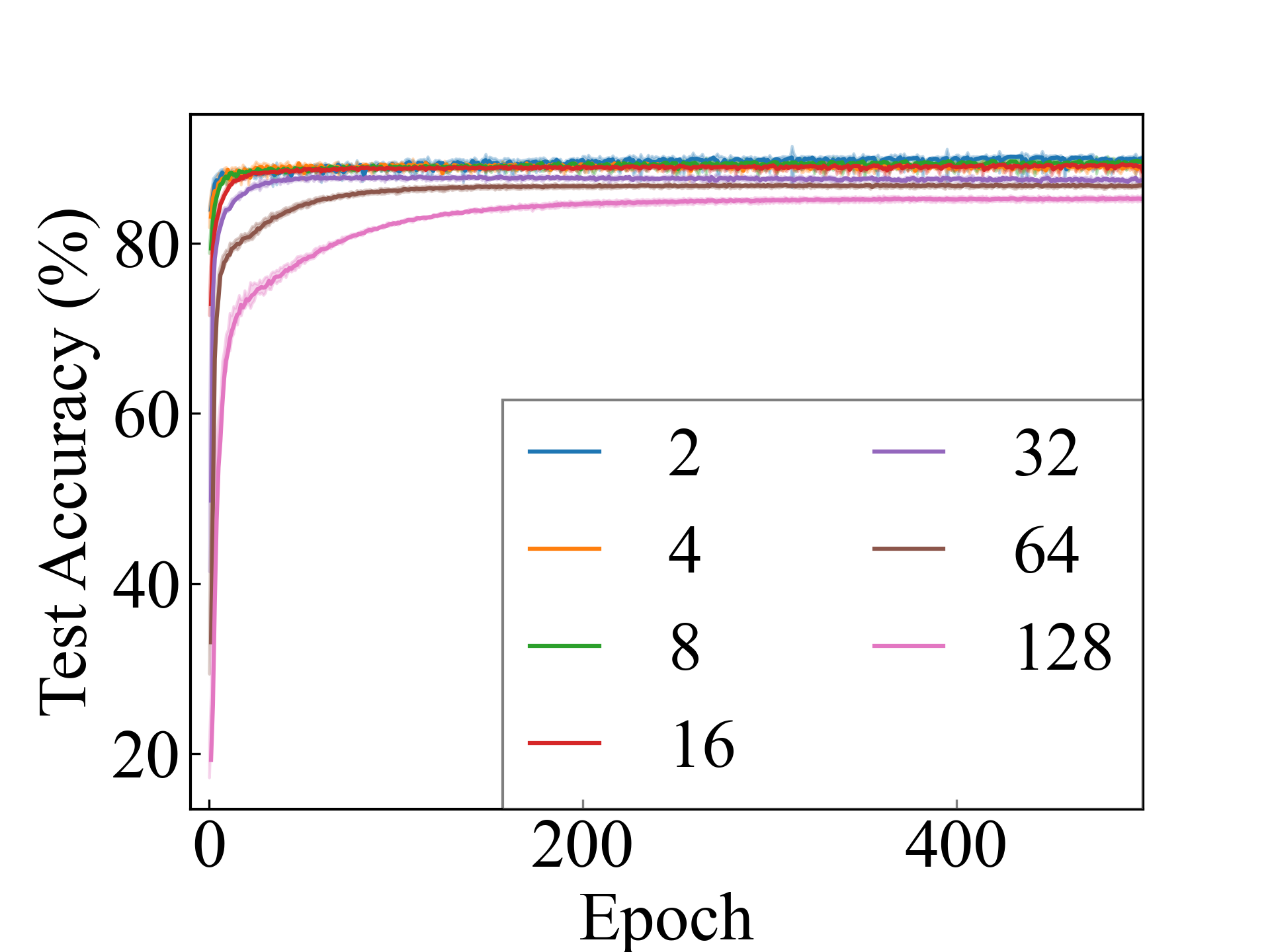}
      \centerline{\quad MLP, RC\_Approx}
  \end{minipage}}
  \subfigure{
    \begin{minipage}[b]{0.24\columnwidth}
      \includegraphics[width=\columnwidth]{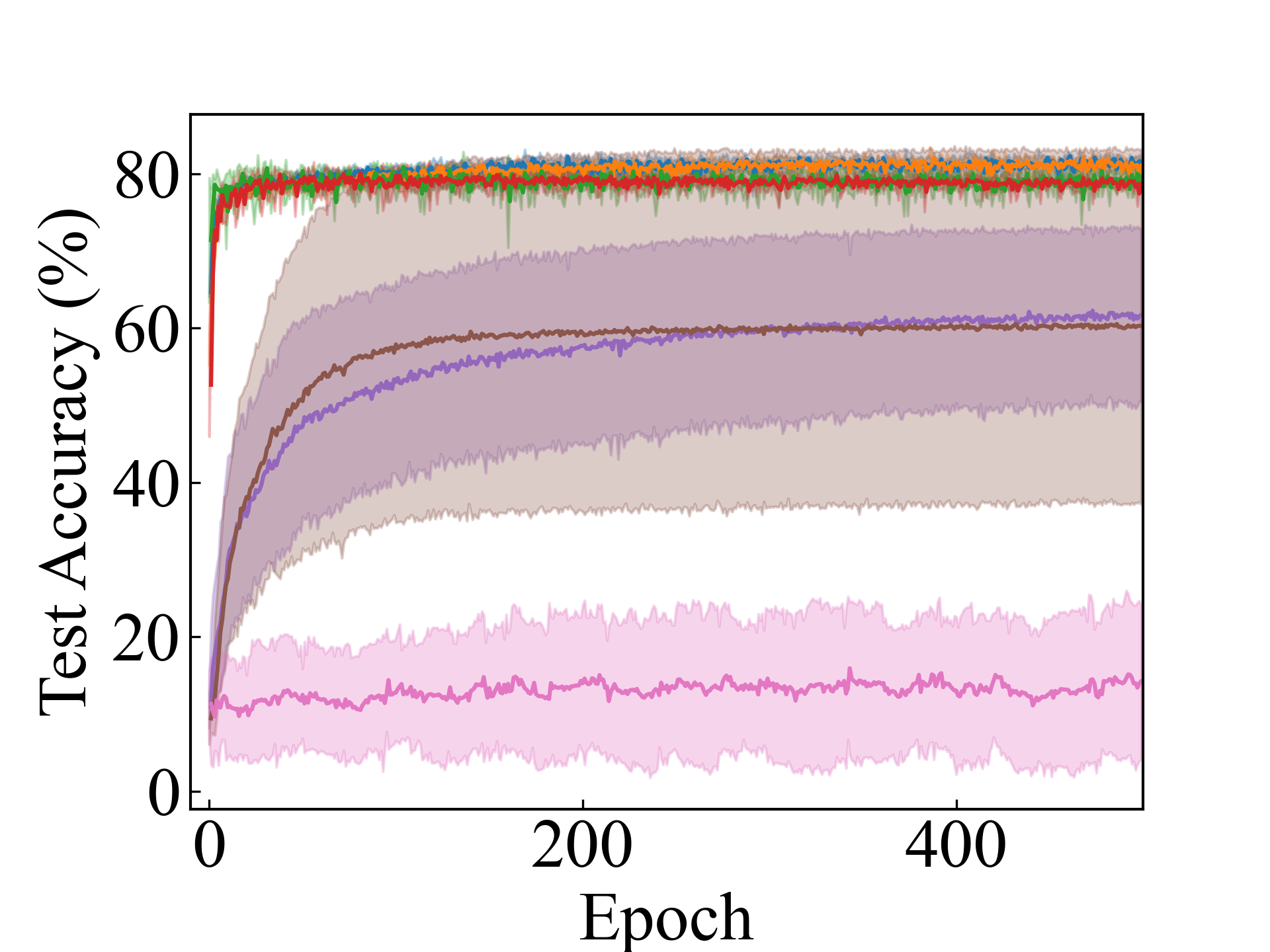}
      \centerline{\quad Linear, CC\_Approx}
  \end{minipage}}
  \subfigure{
    \begin{minipage}[b]{0.24\columnwidth}
      \includegraphics[width=\columnwidth]{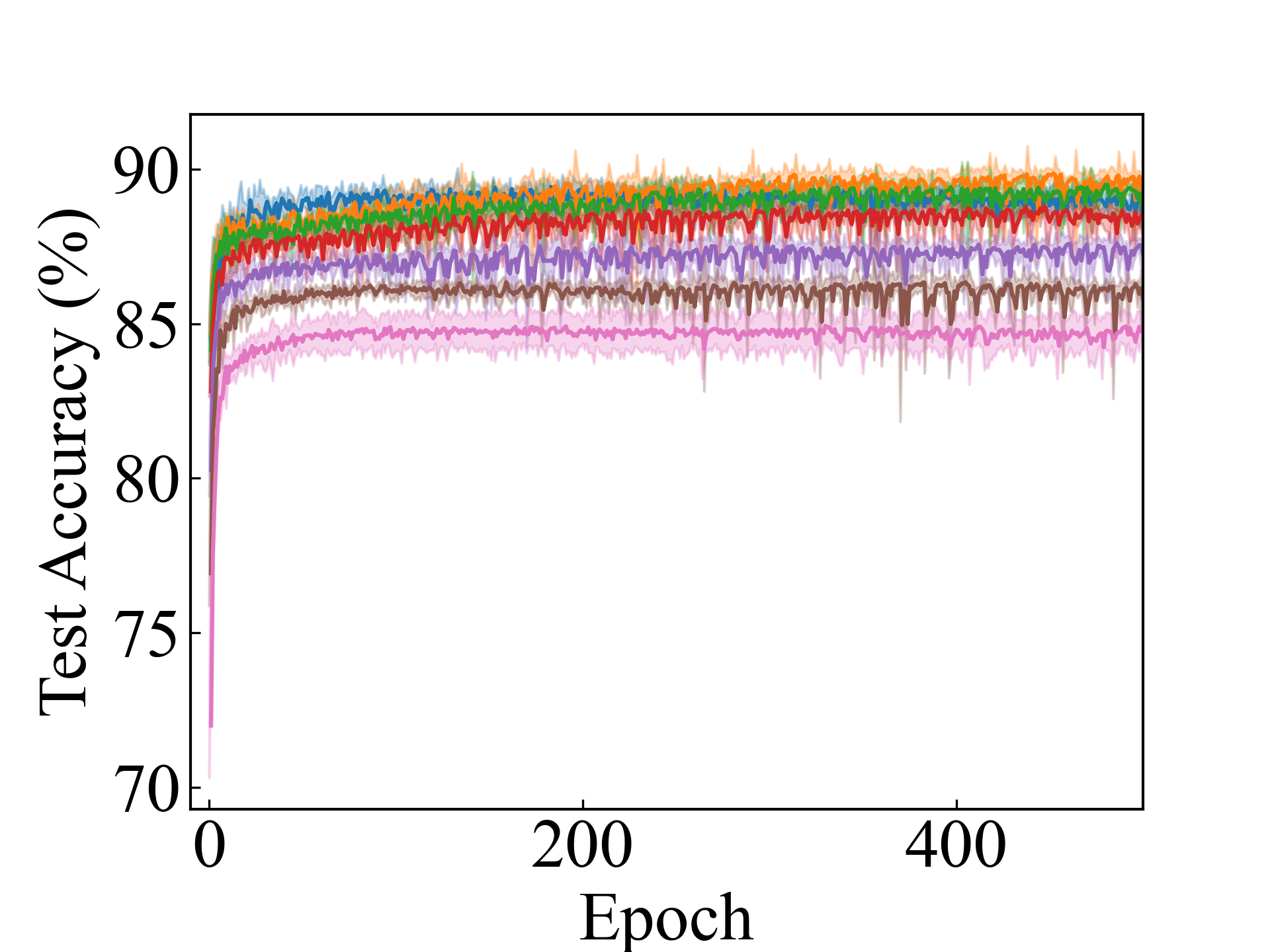}
      \centerline{\quad MLP, CC\_Approx}
  \end{minipage}}
  \subfigure{
    \begin{minipage}[b]{0.24\columnwidth}
      \centering K-MNIST
      \includegraphics[width=\columnwidth]{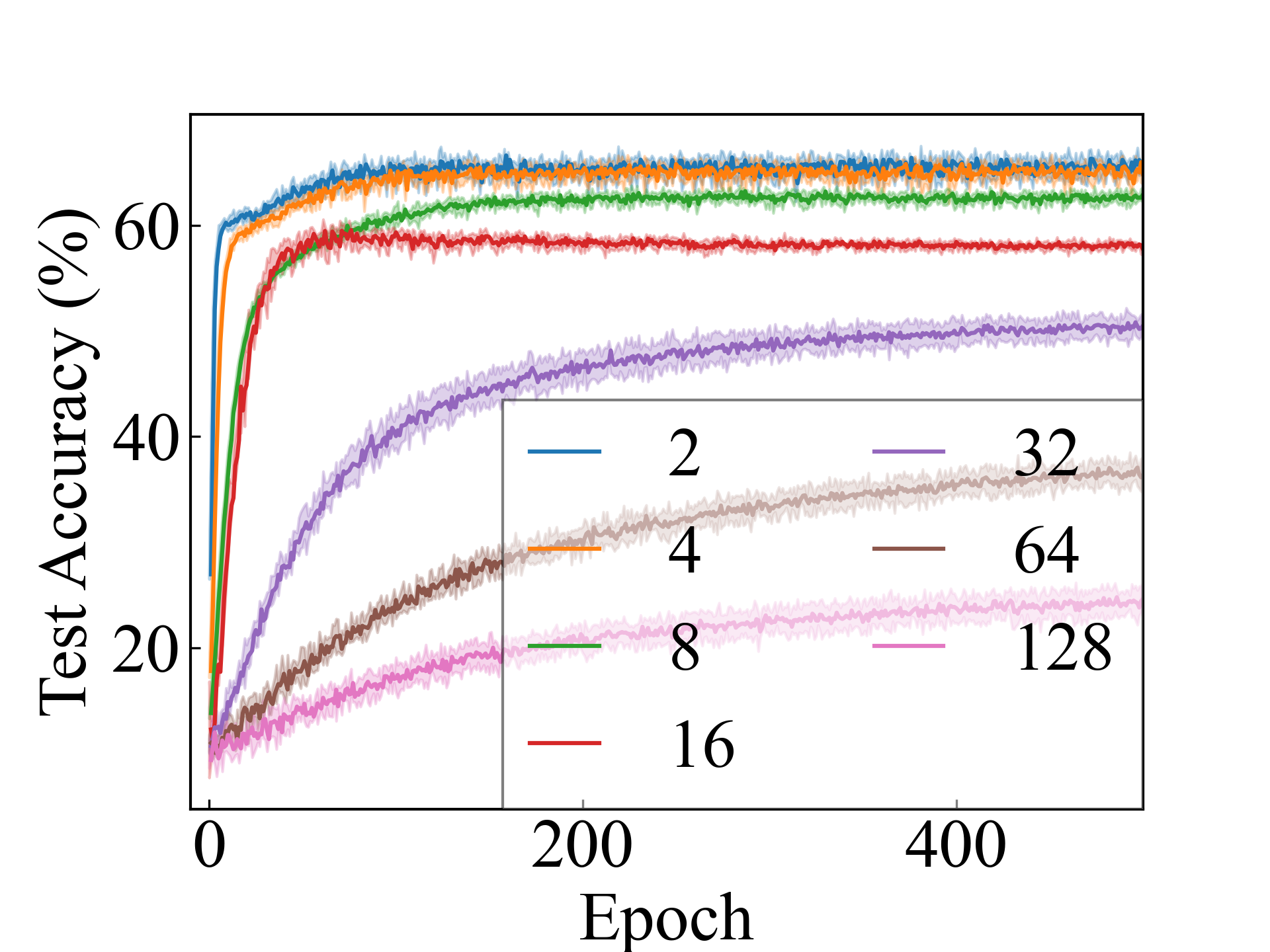}
      \centerline{\quad Linear, RC\_Approx}
  \end{minipage}}
  \subfigure{
    \begin{minipage}[b]{0.24\columnwidth}
      \includegraphics[width=\columnwidth]{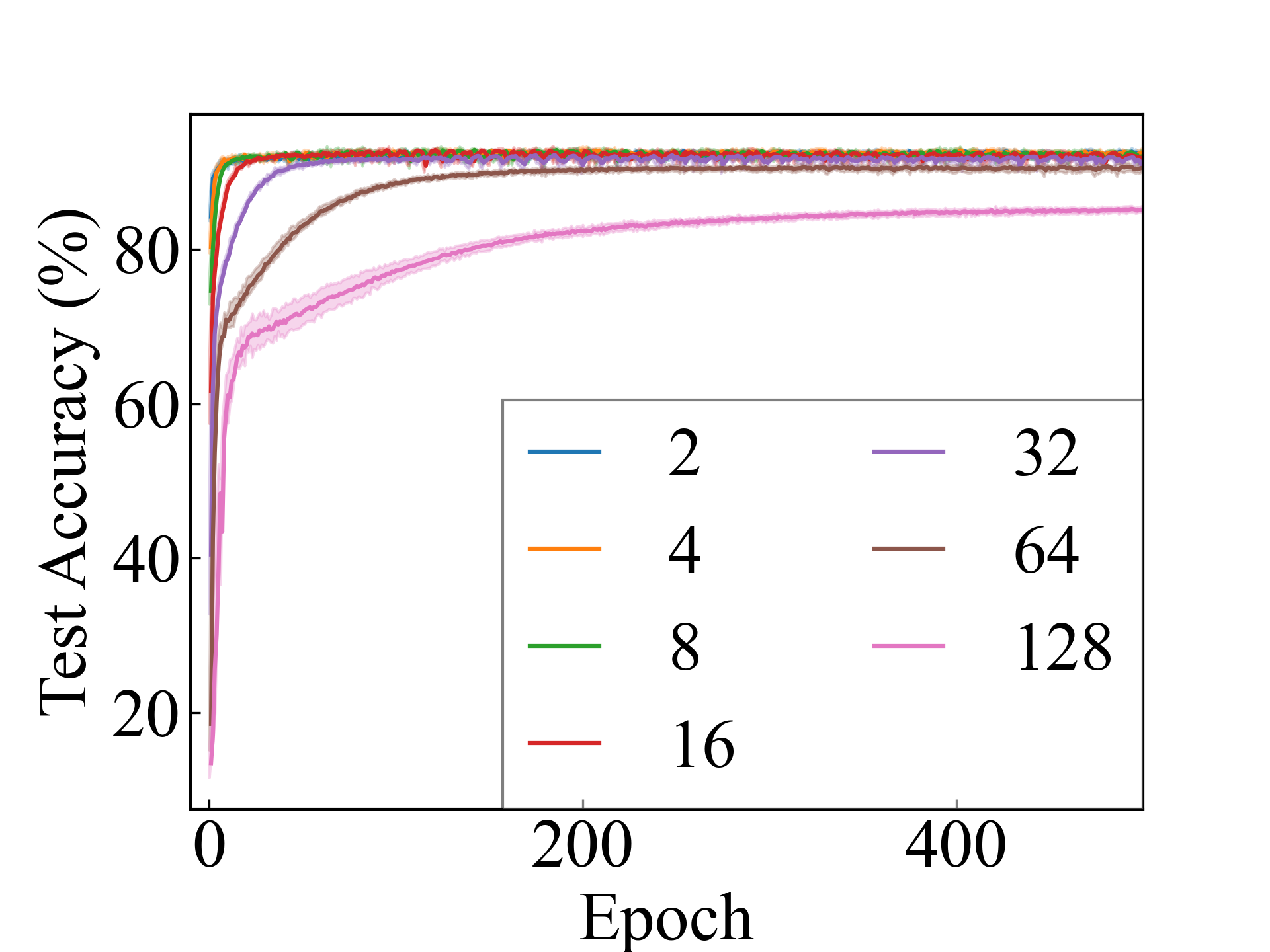}
      \centerline{\quad MLP, RC\_Approx}
  \end{minipage}}
  \subfigure{
    \begin{minipage}[b]{0.24\columnwidth}
      \includegraphics[width=\columnwidth]{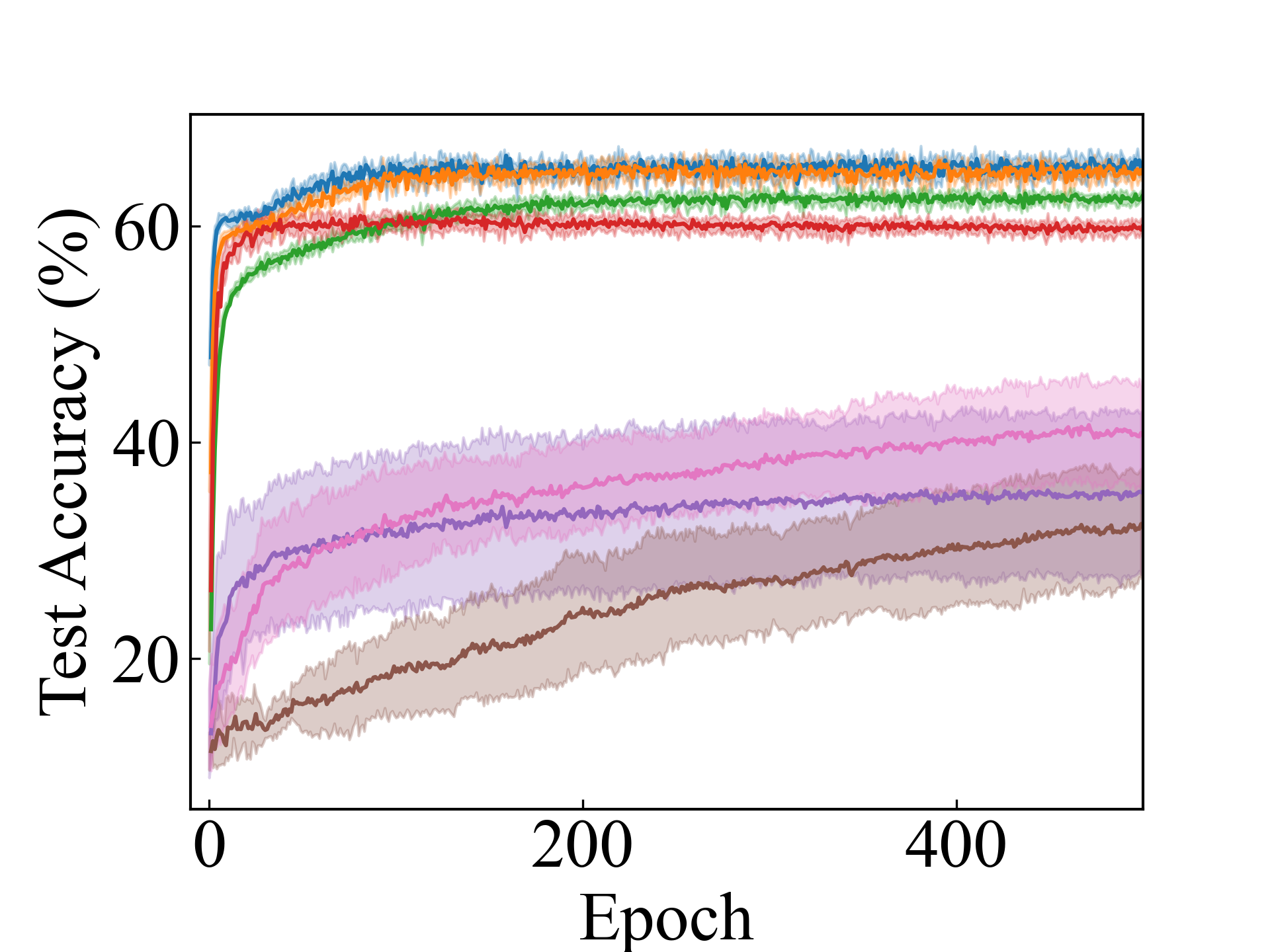}
      \centerline{\quad Linear, CC\_Approx}
  \end{minipage}}
  \subfigure{
    \begin{minipage}[b]{0.24\columnwidth}
      \includegraphics[width=\columnwidth]{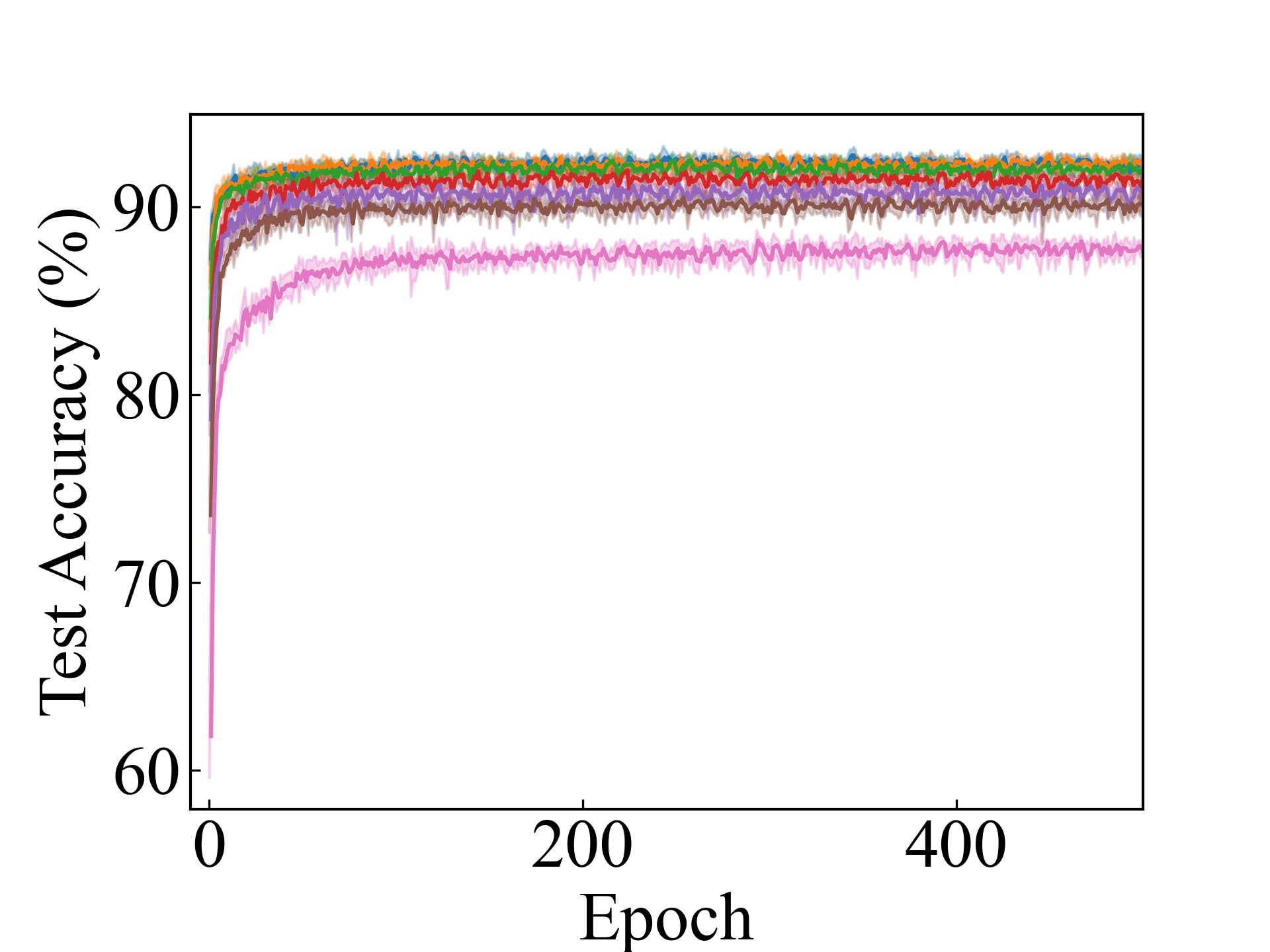}
      \centerline{\quad MLP, CC\_Approx}
  \end{minipage}}
  \subfigure{
    \begin{minipage}[b]{0.24\columnwidth}
      \centering CIFAR-10
      \includegraphics[width=\columnwidth]{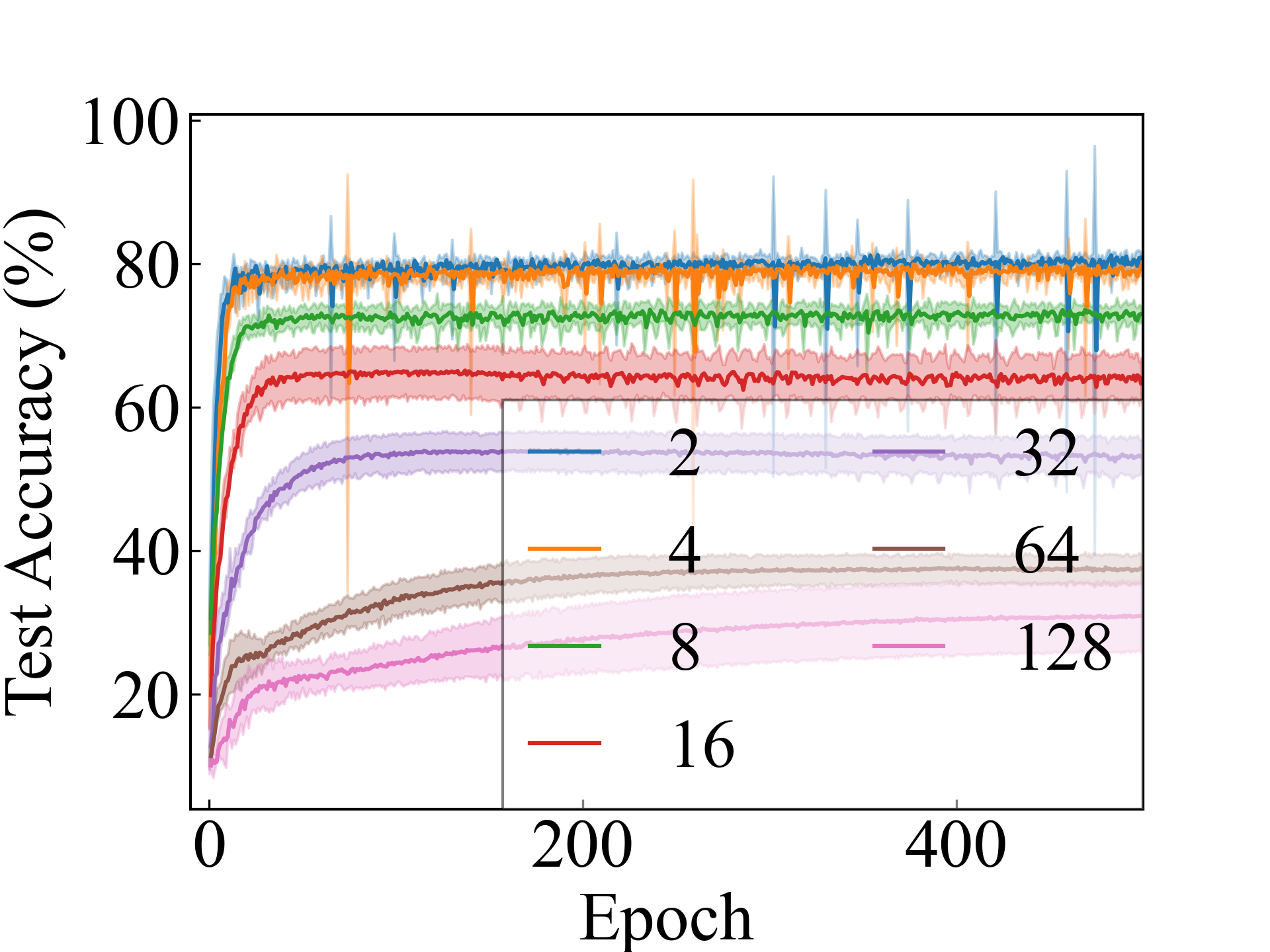}
      \centerline{\quad ResNet, RC\_Approx}
  \end{minipage}}
  \subfigure{
    \begin{minipage}[b]{0.24\columnwidth}
      \includegraphics[width=\columnwidth]{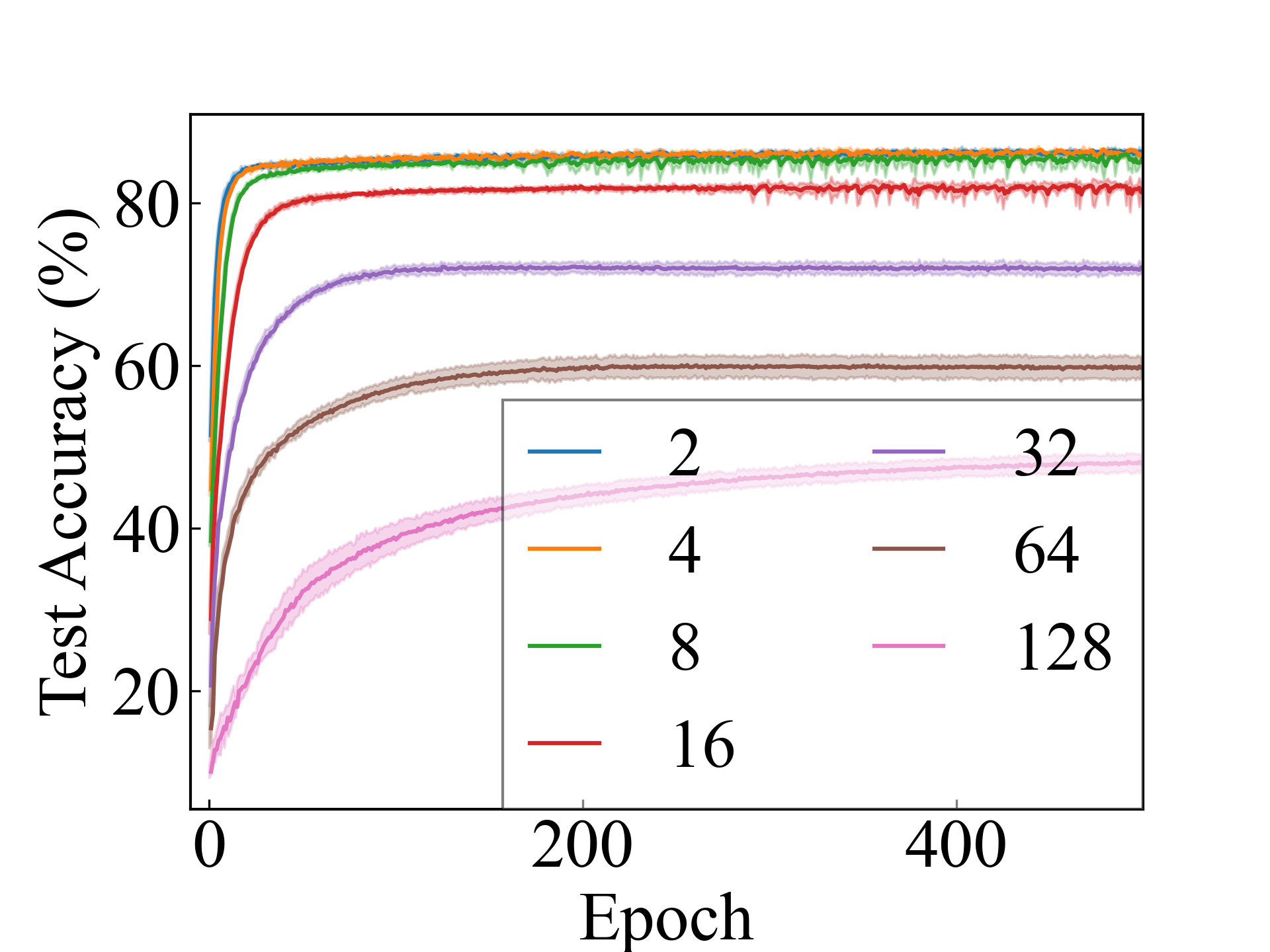}
      \centerline{\quad ConvNet, RC\_Approx}
  \end{minipage}}
  \subfigure{
    \begin{minipage}[b]{0.24\columnwidth}
      \includegraphics[width=\columnwidth]{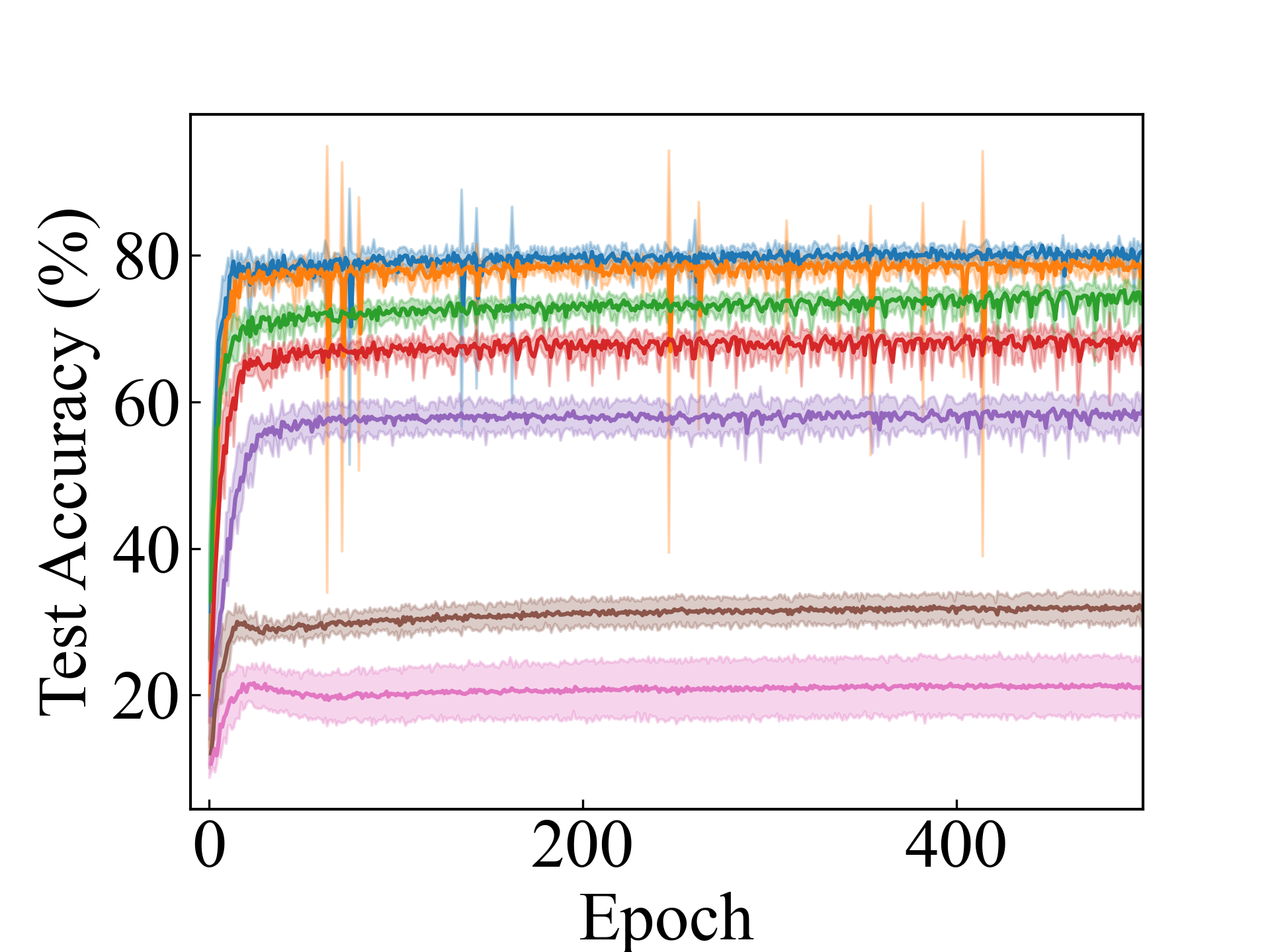}
      \centerline{\quad ResNet, CC\_Approx}
  \end{minipage}}
  \subfigure{
    \begin{minipage}[b]{0.24\columnwidth}
      \includegraphics[width=\columnwidth]{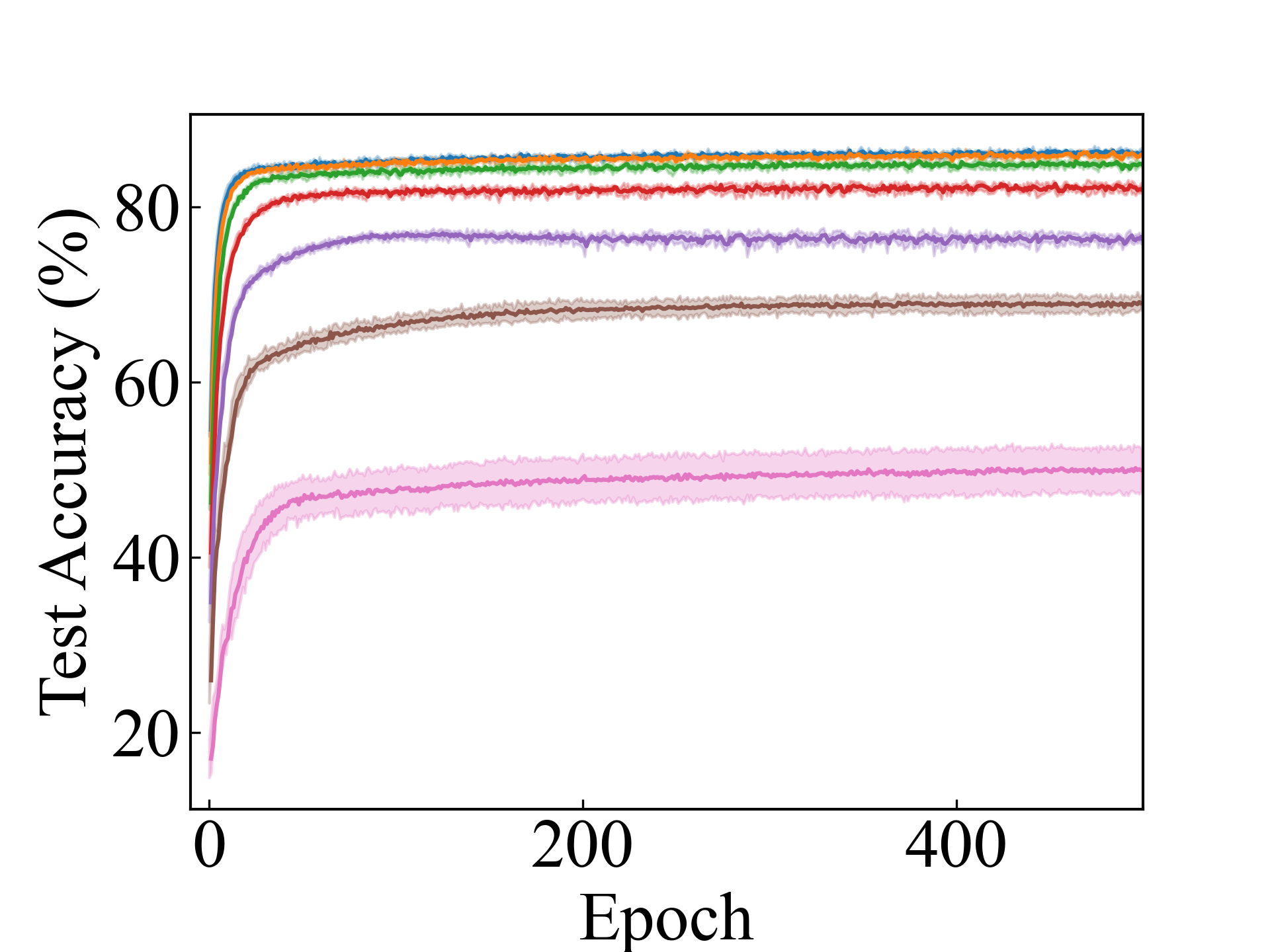}
      \centerline{\quad ConvNet, CC\_Approx}
  \end{minipage}}
  \caption{
  Test accuracy for various settings with proposed methods (RC\_Approx and CC\_Approx).
  }
  \label{fig:test-accuracy-curve2}
\end{figure*}